\documentclass[opre,nonblindrev]{myinforms}
\OneAndAHalfSpacedXI %

\usepackage{endnotes}
\let\footnote=\endnote
\let\enotesize=\normalsize
\def\notesname{Endnotes}%
\def\makeenmark{$^{\theenmark}$}
\def\enoteformat{\rightskip0pt\leftskip0pt\parindent=1.75em
	\leavevmode\llap{\theenmark.\enskip}}

\usepackage{natbib}
\bibpunct[, ]{(}{)}{,}{a}{}{,}%
\def\bibfont{\small}%
\def\enoteformat{\rightskip0pt\leftskip0pt\parindent=1.75em
  \leavevmode\llap{\theenmark.\enskip}}
\TheoremsNumberedThrough     %
\ECRepeatTheorems

\EquationsNumberedThrough    %

\usepackage[linesnumbered,ruled,boxed]{algorithm2e}  
\usepackage{setspace}
\SetKwInput{KwInit}{Initialization}{}{}
\SetKwInput{KwInput}{Input}                %
\SetKwInput{KwOutput}{Output} 			   %
\usepackage{comment}
\usepackage{bbm}
\usepackage{mathrsfs}
\usepackage{enumerate}
\usepackage{tabularx}
\usepackage{booktabs}
\usepackage{color}
\usepackage[dvipsnames]{xcolor}  
\usepackage{amsmath}
\usepackage{mathtools}
\usepackage{thmtools} 
\usepackage{thm-restate}
\usepackage{float}
\usepackage{longtable}
\usepackage{subfigure}
\usepackage{hyperref}
\usepackage{amssymb}
\usepackage[normalem]{ulem}
\hypersetup{ 
	colorlinks,
	linkcolor={red!50!black},
	citecolor={blue!50!black},
	urlcolor={blue!80!black}
}
\usepackage{cleveref}
\AtEndEnvironment{proof}{\hfill $\square$}
\crefname{assumption}{assumption}{assumptions}
\AtEndEnvironment{example}{\hfill $\square$}
\newenvironment{continuance}[1]{\par\vspace{1em}\noindent\textsc{#1, Cont'd}.}
  {\par}
\AtEndEnvironment{continuance}{\hfill $\square$}

\usepackage{marginnote}
\allowdisplaybreaks[1]
\newcommand{\mnote}[3]{}

\newcommand{\recv}{\mu} %
\newcommand{\re}{r} %

\newcommand{\UCB}{\mathrm{UCB}}
\newcommand{\LCB}{\mathrm{LCB}}
\newcommand{\event}{\mathcal{A}} %
\newcommand{\drecv}{\delta} %

\newcommand{\ie}{\emph{i.e.}}
\newcommand{\eg}{\emph{e.g.}}

\newcommand{\eb}{\mathbf{e}}

\newcommand{\kb}{\boldsymbol{k}}

\newcommand{\nbb}{\boldsymbol{n}}

\newcommand{\Acal}{\mathcal{A}}

\newcommand{\Mcal}{\mathcal{M}}

\newcommand{\radius}{\Delta}

\newcommand{\bvUCB}[1]{\boldsymbol{v}^{\UCB}_{#1}}

\newcommand{\vLCB}[1]{v^{\LCB}_{#1}}
\newcommand{\vUCB}[1]{v^{\UCB}_{#1}}

\newcommand{\profit}{\Pi}
\newcommand{\bv}{\boldsymbol{v}}
\newcommand{\bre}{\boldsymbol{r}}
\newcommand{\bo}{\boldsymbol{o}}
\newcommand{\bs}{\boldsymbol{s}}
\newcommand{\cvec}{\boldsymbol{c}} %
\newcommand{\const}{a}
\newcommand{\btht}{\boldsymbol{\theta}}

\newcommand{\setU}{\mathcal{U}}
\newcommand{\setexp}{\mathcal{E}}
\newcommand{\setS}{\mathcal{D}} %

\newcommand{\U}{\boldsymbol{u}}
\newcommand{\M}{M}
\newcommand{\R}{R}
\newcommand{\s}{\mathcal{S}}

\newcommand{\Eb}[1]{\mathbb{E}\left[#1\right]}
\newcommand{\Epi}[1]{\mathbb{E}_\pi\left[#1\right]}
\newcommand{\prob}[1]{\mathbb{P}\left(#1\right)}
\newcommand{\C}{\bar c} %
\newcommand{\Eover}[2]{\mathbb{E}_{#1}\left[#2\right]}

\newcommand{\mbr}[1]{\left[#1\right]}
\newcommand{\lbr}[1]{\left\{#1\right\}}
\newcommand{\floor}[1]{\left\lfloor#1\right\rfloor}

\newcommand{\bd}{\textbf}
\newcommand{\blue}{ }

\newcommand{\seq}{\boldsymbol{d}}

\newcommand{\argmaxx}{\mathop{\mathrm{argmax}}}

\newcommand{\inst}{I}
\newcommand{\alg}{\mathscr{A}}
\newcommand{\MNL}{\mathrm{MNL}}
\newcommand{\MNLI}{\mathrm{MNLI}}

\newcommand{\Reg}{\mathrm{Reg}}

\newcommand{\balpha}{\boldsymbol{\alpha}}

\newcommand{\tht}{\theta}
\newcommand{\1}[1]{\mathbbm{1}\left\{#1\right\}}
\newcommand{\norm}[2]{\left\lVert#1\right\rVert_{#2}}

\newcommand{\Abs}[1]{\left|#1\right|}

\newcommand{\lm}{\lambda}

\newcommand{\edit}[1]{#1}

\begin{document}
		
	\RUNAUTHOR{Liang et al.}
	
	\RUNTITLE{Online Joint Assortment-Inventory Optimization}
	
	\TITLE{Online Joint Assortment-Inventory Optimization under MNL Choices}
	
	\newcommand*\samethanks[1][\value{footnote}]{\footnotemark[#1]}
	\ARTICLEAUTHORS{%
		\AUTHOR{Yong Liang\thanks{Authors are listed in alphabetical order.}}
 		\AFF{Research Center for Contemporary Management, Key Research Institute of Humanities and Social Sciences at Universities, School of Economics and Management, Tsinghua University, 100084 Beijing, China, \EMAIL{liangyong@sem.tsinghua.edu.cn}}
        \AUTHOR{Xiaojie Mao\samethanks}
 		\AFF{Department of Management Science and Engineering, School of Economics and Management, Tsinghua University, 100084 Beijing, China, \EMAIL{maoxj@sem.tsinghua.edu.cn}}
        \AUTHOR{Shiyuan Wang\samethanks}
 		\AFF{Department of Management Science and Engineering, School of Economics and Management, Tsinghua University, 100084 Beijing, China, \EMAIL{wangshiy20@mails.tsinghua.edu.cn}}
} %

\ABSTRACT{%
    We study an online joint assortment-inventory optimization problem, in which we assume that the choice behavior of each customer follows the Multinomial Logit  (MNL) choice model, and the attraction parameters are unknown \textit{a priori}. The retailer makes periodic assortment and inventory decisions to dynamically learn from the customer choice observations about the attraction parameters while maximizing the expected total profit over time. In this paper, we propose a novel algorithm that can effectively balance exploration and exploitation in the online decision-making of assortment and inventory. Our algorithm builds on a new estimator for the MNL attraction parameters, an innovative approach to incentivize exploration by adaptively tuning certain known and unknown parameters, and an optimization oracle to static single-cycle assortment-inventory planning problems with given parameters. We establish a regret upper bound for our algorithm and a lower bound for the online joint assortment-inventory optimization problem, suggesting that our algorithm achieves nearly optimal regret rate, provided that the static optimization oracle is exact.  Then we incorporate more practical approximate static optimization oracles into our algorithm, and bound from above the impact of static optimization errors on the regret of our algorithm. We perform numerical studies to demonstrate the effectiveness of our proposed algorithm. At last, we extend our study by incorporating inventory carryover and the learning of customer arrival distribution.
}%

\KEYWORDS{assortment planning, inventory management, multi-armed bandit, multinomial logit, online optimization, upper confidence bound} 
\HISTORY{}

\maketitle

\section{Introduction}
In the realm of modern retail, the success of retailers hinges on their ability to optimize their product assortment and inventory levels. The assortment optimization problem involves selecting a subset of products from a pool of similar alternatives to offer to customers. Finding the optimal assortment has a significant impact on retailers' profits, and thus it is prioritized by consultants, software providers, and retailers alike \citep{kok2008assortment}. Nonetheless, merely selecting the ``optimal'' assortment falls short of maximizing profits. In most offline and many online retail practices, managing inventory comes at a significant cost, and inventory capacity is often limited. Therefore, optimizing inventory decisions is just as vital to profit maximization as optimizing product assortment. 
Consequently, in order to maximize profit, the retailer must jointly optimize the assortment and inventory decisions that produce the optimal sequence of assortments. Despite its significance, the joint optimization problem of assortment and inventory is challenging, and the literature on this topic appears to be limited.

To optimize assortment and inventory decisions, retailers must accurately model customers' choice preferences. While the existing literature typically assumes fully-specified models with known parameters, in practice, the parameters are usually unknown and need to be inferred from sales data. If a sufficiently large and high-quality dataset already exists, retailers can estimate model parameters and use them to optimize decisions. However, acquiring such data is often a challenge, particularly when retailers expand into new territories where data must be acquired from scratch to learn local customers' preferences. %
Moreover, even when historical data are available, the data may not accurately reflect current customers' preferences when new products are introduced or existing products are removed. %
Consequently, the ``estimate-then-optimize'' approach may be impractical. Instead, retailers effectively face an online decision-making setting where they need to learn customers' preferences \edit{on the fly} while making decisions. 

In this paper, we focus on the online joint optimization  of  assortment and inventory decisions. Specifically, we assume that customers' choice behaviors follow the widely-used Multinomial Logit (MNL) choice model, and the attraction parameters are unknown to the retailer \textit{a priori}. The retailer needs to determine a sequence of periodic assortment and inventory decisions to optimize the expected total profit over a long planning horizon, while learning the choice model parameters on the fly. The planning horizon is discretized into periods that correspond to inventory cycles. At the beginning of each inventory cycle, the retailer determines both the assortment and the inventory levels for the assorted products. Customers then arrive sequentially, each making a purchase decision for at most one product from the assortment set available upon arrival. At the end of each inventory cycle, unsold products are salvaged. \edit{We later relax this assumption to allow for inventory carryover.} %
For this online joint assortment and inventory optimization problem, the key to designing an effective algorithm is to strike a balance between \emph{exploration} and \emph{exploitation}. That is, we need to balance the exploration of seemingly suboptimal products to collect more informative data and the exploitation of products that appear attractive according to existing data.

The focal problem studied in this paper presents significant challenges, arising from complicated product substitution behaviors due to stochastic stockout events. %
\edit{As noted previously, stockout events happen constantly when the inventory for certain products is depleted due to customer demand.}
Depending on different realizations of customers' choices, the number of possible sequences of product stockout events is overwhelmingly large, as is the number of possible ways the assortment can evolve. Since the assortment directly affects each customer's choices, the overall choice probability for a particular product across an inventory cycle can evolve in an intractable way. Consequently, even with known attraction parameters, efficiently calculating the expected demand for a product remains an open question \citep{aouad2022stability}. The problem becomes even more complicated when the attraction parameters are not known and must be estimated from data, particularly since that parameter estimation is itself a difficult task due to the complexities of the underlying choice distributions.  
Moreover, the stochastic stockout events also pose serious challenges to the design of exploration-exploitation algorithms. In order to balance exploration and exploitation, the retailer needs to make decisions not only according to the estimates of the expected profit of different assortment and inventory decisions, but also the uncertainty in the profit estimates. However, due to the stockout events, the expected profit and the attraction parameters generally display intricate relationships, so even when the uncertainty in the attraction parameter estimates can be characterized, propagating the uncertainty to the profit estimates is difficult. In particular, we will illustrate how this cannot be achieved by directly extending some existing upper confidence bound methods.

The main contributions of this paper are as follows. First, we propose an exploration-exploitation algorithm that addresses the aforementioned challenges. To our best knowledge, this is the first algorithm for the online joint assortment and inventory optimization problem under the MNL choice model with unknown attraction parameters. Specifically, the proposed algorithm offers two key contributions: a consistent and efficient estimator for the unknown attraction parameters and a novel exploration-exploitation algorithm that deliberately tunes products' unit profits to achieve sufficient exploration, even though their true values are known.
\edit{Our proposed estimation method, which we will discuss in detail later, has advantages over maximum likelihood estimation in multiple dimensions. Moreover, our proposal to tune known parameters appears as an unconventional exploratory strategy, which could be useful for other complex online decision-making problems.}\mnote{blue}{AEQ1\\R1Q1}{-20mm}
Second, we evaluate the performance of the decisions yielded by the proposed algorithm by comparing them with the optimal decisions obtained when the attraction parameters of all products are known \textit{a priori}. In particular, we adopt the widely-used relative profit gap, also known as the \textit{regret}, as a performance metric. We show that our algorithm achieves nearly optimal regret rate by establishing and matching a non-asymptotic upper bound and a worst-case lower bound on the regret. Third, while the above results rely on an oracle that can exactly solve the static joint assortment and inventory optimization problem, we extend our algorithm by incorporating approximate optimization oracles. Then, we demonstrate that the resulting regret bound automatically adapts to the errors of the approximate oracle. {Fourth, we study an extension where unused inventory is carried over to the next inventory cycle. We devise an algorithm that achieves the same regret convergence rate. Lastly, we also examine an extension where the demand arrival distribution is unknown and also needs to be learned. } \mnote{blue}{AEQ2\\AEQ3\\R2Q1\\R2Q2}{-20mm}

The remainder of this paper is organized as follows. \Cref{sec-literature} reviews the related literature. \Cref{sec-formulation} formulates the problem. \Cref{sec-algorithm} discusses the main challenges of the online optimization problem and presents our algorithm which consists of both a novel estimation approach and a novel exploration-exploitation algorithm. \Cref{sec-results} derives upper and lower regret bounds. \Cref{sec-oracles} incorporates into the proposed algorithm approximation oracles for the static joint assortment and inventory optimization problem. \Cref{sec-simulation} performs numerical experiments to demonstrate the performance of the proposed algorithm. \Cref{sec:Extensions} studies two extensions. At last, \Cref{sec-conclusion} concludes this paper.

\section{Literature Review}
\label{sec-literature}
We review three lines of closely related research: assortment optimization, joint assortment and inventory optimization, and the multi-armed bandit (MAB) problem.

\textbf{MNL choice model and assortment optimization.} In the literature, there exist a variety of choice models describing the choice preferences of customers. The multinomial logit (MNL) model is arguably one of the most popular models due to its convenience and its superior performance in practice \citep{feldman2022customer}. In particular, under an MNL choice model with known parameters, the (static) assortment optimization problem, where the retailer selects the assortment to maximize the profit from the next customer, can be solved efficiently \citep{ryzin1999relationship,talluri2004revenue}. 
Several studies further extend the MNL assortment optimization problem by incorporating practice-driven constraints, such as the cardinality constraint \citep[][]{rusmevichientong2010dynamic} and the totally unimodular constraints \citep[][]{sumida2021revenue}. Some literature builds on the MNL model to optimize assortment under complex channel structures \citep{dzyabura2018offline} or to jointly optimize assortment with other decisions such as pricing \citep[][]{gao2021assortment}. In addition, more complex choice models extending from the MNL model, including the mixture-of-logits \citep[][]{rusmevichientong2014assortment} and the nested logit models \citep[][]{Li2015dlevel}, have been proposed, leading to more complicated assortment optimization problems. It is worth noting that besides parametric choice models such as the MNL model, there are also non-parametric approaches to model customer choices \citep[e.g.,][]{farias2013nonparametric}. %
Our study builds on the MNL choice model, but we study the joint optimization of assortment and inventory.

A current line of research focuses on online assortment optimization, where customer preferences are unknown \textit{a priori} and need to be estimated from data while making the assortment decisions. Early studies often rely on strong assumptions that are not needed in our paper. For example, \cite{caro2007dynamic} assume that the demands of different products are independent, while the MNL model considered in our paper captures demand substitutions among products. Other studies, such as those by \cite{rusmevichientong2010dynamic} and \cite{saure2013optimal}, consider the online assortment optimization under the MNL model but require prior knowledge of ``separability'', which refers to a constant gap between the expected revenue of the optimal assortment and those of other alternative assortments. However, the knowledge of separability is often not available in practice.

Our study is closely related to recent works by \cite{agrawal2017thompson, agrawal2019mnl} and \cite{chen2021optimal}, which address online assortment optimization  under the MNL choice model with cardinality constraints on the assortment set. \cite{agrawal2017thompson} and \cite{ agrawal2019mnl} formulate the assortment optimization problem as a multi-armed bandit (MAB) problem  and extend the classic Thompson sampling and the upper confidence bound algorithms, respectively. \cite{chen2021optimal} derive a trisection search algorithm. All three algorithms achieve worst-case regret upper bounds of the same order. %
\cite{chen2018note} further establish a %
regret lower bound that matches the upper bound up to logarithmic factors when the cardinality constraint is sufficiently tight. %
It is noteworthy that  the online assortment optimization problem can be viewed as a special case of our problem with only one customer arrival in each inventory cycle. 
In particular, when specialized to this setting, the regret upper bound derived for our algorithm agrees with those of the three aforementioned algorithms.

\textbf{Joint assortment and inventory planning.} 
While the aforementioned literature focus on assortment management, in practice inventory management is also important. In particular, when the inventory is limited, stockout events can occur frequently and change the assortments available to customers. 
The stochastic stockout events lead to dynamic demand substitutions and significantly impact total profits \citep{smith2000management, mahajan2001stocking}. Thus, optimizing profit requires a joint optimization of assortment and inventory. However, even the static joint assortment and inventory optimization problem can be notoriously difficult to solve \citep{goyal2016near}, where the expected profit function generally lacks desirable properties such as quasi-concavity \citep{mahajan2001stocking} or submodularity \citep{aouad2019approximation}. {Consequently, various heuristics have been developed \citep[see, e.g.][]{mahajan2001stocking, honhon2010assortment, honhon2013fixed,farahatMultiproductNewsvendorProblem2018}.}\mnote{blue}{R1Minor1}{0mm} Alternatively, some recent literature alleviates the challenge by imposing assumptions that effectively restrict the possible sequences of stockout events \citep[see, e.g.][]{goyal2016near, segev2019assortment}.

A recent stream of literature investigates approaches to approximately solve the static joint assortment and inventory optimization problem under the MNL choice model with known parameters.  {Notably, \cite{aouad2018greedy} propose a greedy-like algorithm with a theoretical guarantee. \cite{aouad2022stability} develop a polynomial-time approximation scheme and establish a ``stability'' property of expected demand with respect to the MNL attraction parameters. This stability property is also used in our analyses. \mnote{blue}{R1Minor1}{0mm}Recently, \cite{sun2024unifiedalgorithmicframeworkdynamic} introduced approximation algorithms with superior performance. While the above research considers a capacity constraint on the total product inventory, some other studies do not impose hard constraints and instead  account for ordering costs. In this context, \cite{liang2021assortment} design a heuristic algorithm based on fluid approximation and demonstrate its asymptotic optimality when the expected number of arriving customers approaches infinity. Extending the choice model to the Markov chain choice model for the unconstrained problem, \cite{mouchtakiJointAssortmentInventory} develop an sample average approximation-based algorithm that is near-optimal relative to an LP upper bound. \cite{zhangLeveragingDegreeDynamic} further refine the regret guarantee.} While our focus is on solving the \textit{online} joint assortment and inventory optimization problem, these aforementioned approximation algorithms and heuristics can be directly used as optimization oracles in our algorithm. Details will be discussed in \Cref{sec-oracles}.

In contrast, the literature on the online joint assortment and inventory optimization problem is scarce, although there exists abundant literature on online assortment optimization problems (as reviewed above) and on online inventory optimization \citep[e.g.][]{chen2020dynamic, gao2022efficient} separately. 
 In the online joint assortment and inventory optimization problem, we find that the estimation of choice model parameters and the balance of exploration and exploitation are both very challenging due to stochastic stock-out events (see \Cref{subsec-challenge-algsteps} for an overview).
 To the best of our knowledge, our study is the first to investigate this problem under the MNL model.

\textbf{Multi-armed bandit problem.} Another closely related line of research is on the multi-armed bandit (MAB) problem \citep[][]{lai1985asymptotically, auer2002finite}, a classic problem that exemplifies the exploration–exploitation tradeoff. There has been very rich literature on MAB, and we refer to \cite{bubeck2012regret}, \cite{MAL-068}, \cite{shipra2019recent} and \cite{lattimore2020bandit} for more comprehensive reviews and recent advances. MAB has been widely adopted to model the online assortment optimization problems under a variety of choice models, including the MNL model \citep{rusmevichientong2010dynamic, saure2013optimal, agrawal2017thompson, agrawal2019mnl, chen2018note, chen2021optimal}, the contextual MNL model \citep[][]{Oh2019thompson, chenxi2020dynamic}, and the nested logit model \citep[][]{chen2021dynamic}. Additionally, MAB has also been utilized to model online inventory management problems \citep[][]{gao2022efficient, cheung2022inventory}. The most famous exploration–exploitation algorithms for the MAB problem include the upper confidence bound (UCB) algorithm \citep[][]{rajeev1995applied, auer2002finite} and the Thompson sampling (TS) algorithm \citep[][]{william1933likelihood, agrawal2012further, agrawal2017near}. In this paper, we propose an exploration–exploitation algorithm for online joint assortment and inventory optimization based on a non-trivial extension of the UCB algorithm.

A notable extension of the MAB problem is the combinatorial multi-armed bandit (CMAB) problem that involves NP-hard combinatorial optimization. For example,  \cite{chen2013combinatorial} assume the availability of an approximation oracle that approximately solves the combinatorial optimization, and define the regret as the expected profit of the online solutions against that of the clairvoyant approximate solutions. Our online joint assortment and inventory optimization problem is also combinatorial, and thus we also consider using approximation oracles in our exploration-exploitation algorithm. We analyze the corresponding regret that incorporates both the regret from profit estimation errors and that from optimization errors of approximation oracles. The crucial difference between our work and that of \cite{chen2013combinatorial} is that they assume the outcome of one arm is independent of that of another, while we have no such restrictive assumption.

\section{Problem Formulation}\label{sec-formulation}

\begin{figure}[htbp]
	\begin{center}
		\includegraphics[height=1.3in]{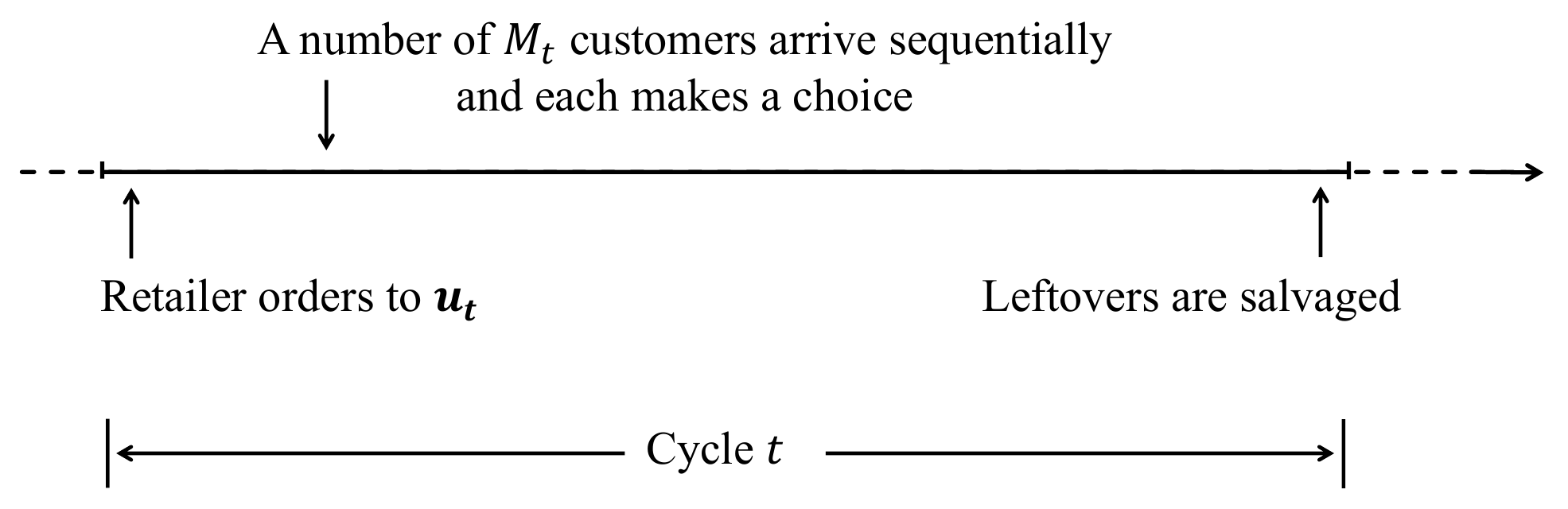}
		\caption{Sequence of events in cycle $t$: At the beginning of inventory {cycle} $t$, the retailer makes inventory ordering decision $\U_t$. Then, $M_t$ customers arrive sequentially, and each customer either purchases one product from the available assortment or leaves for the outside option. At the end of cycle $t$, inventory leftovers are salvaged.} \label{fig-cycles}
	\end{center}
\end{figure}

We consider a retailer making periodic joint assortment-inventory decisions for a set of substitutable products, denoted by $[N] \coloneqq \{1,2,\ldots,N \}$. 
The attraction of each product is unknown to the retailer \textit{a priori}. The objective of the retailer is to maximize the expected cumulative  profit over $T$ inventory cycles, where $T$ is \emph{not} known to the retailer from the very beginning.  \Cref{fig-cycles} illustrates a sequence of events occurring in an arbitrary inventory cycle $t \in [T]$. 

First, at the beginning of each cycle $t$, the retailer determines the inventory order-up-to levels for the $N$ products, denoted by $\U_t \coloneqq (u_{i,t})_{i \in [N]}$, where $u_{i,t}$ represents the order-up-to level of product $i\in [N]$. Note that $\U_t$ incorporates both the inventory  and assortment  decisions, as a zero inventory level $u_{i,t} =0$ indicates that product $i$ is not assorted. This implies that products with non-zero inventory must be assorted. We use $S(\U_t)$ to denote the set of products assorted by the inventory decision $\U_t$, namely, $\{i \mid i \in [N], \ u_{i,t} >0 \}$. We assume zero lead-time, and thus inventory orders are received instantaneously. 

Next, customers arrive sequentially and make purchases. Let $M_t$ denote the number of customer arrivals during cycle $t$. The number $M_t$ can be either a constant or a random variable whose distribution is known to the retailer and independent and identically distributed (i.i.d) over time $t$. 
\edit{In \Cref{sec:mt}, we consider $M_t$ with an unknown distribution and propose to learn it distribution on the fly while making decisions}.\mnote{blue}{AEQ2\\R2Q1}{-10mm}
Upon the arrival of customer $m \in [M_t]$, the assortment set available to the customer, denoted by $\s_{t, m}$, consists of products that were initially assorted at the beginning of cycle $t$ and have not been depleted yet. 
The customer $m$ either purchases one unit of product from $\s_{t, m}$ or chooses to not purchase at all (which we refer to as the ``outside'' option).  
We capture customers' choice behaviors by the multinomial logit (MNL) model. Specifically, the MNL model involves a vector of attraction parameters $\bv \coloneqq(v_i)_{i \in [N]}$ corresponding to the $N$ products, and an additional attraction parameter $v_0$ corresponding to the outside option (\ie, no purchase) indexed as $0$. 
Let $\seq_t = (d_{t, m})_{m \in [M_t]}$ denote the vector of choices made by customers arriving in cycle $t$, where $d_{t, m}$ represents the choice of customer $m$ in cycle $t$.
  Following the MNL model, the probability that customer $m$ in cycle $t$ purchases product $i$ when offered assortment $\s$ is given by:
\begin{equation}
	\label{equ-define-MNL-choice-probability}
	p_{i}(\s) = \mathbb{P}(d_{t, m} = i | \s_{t, m} = \s) = 
	\begin{cases}\frac{v_i}{v_0+\sum_{j \in \s} v_{j}}, & \text { if } i \in \s \cup\{0\} , \\ 0, & \text { otherwise. }\end{cases} 
\end{equation}
Without loss of generality, we normalize $v_0$ to $1$. \mnote{blue}{R1Q2}{0mm}\edit{We  assume that the other $v_i$'s belong to the interval $[v_{\min}, v_{\max}]$ for two positive constants $v_{\min}$ and $v_{\max}$. Here, $v_{\max}$ is assumed known to the retailer, which is not a restrictive assumption, as typically the outside option is the most frequent choice so $v_{\max} = 1$ is plausible in many settings. In stark contrast, $v_{\min}$ is not required to be known as it could be hard to specify in practice. In our setting, all customers' choices $d_{t, m}$ in all cycles, including the no-purchase choice, can be fully observed by the retailer. This setting is  standard in the data-driven assortment optimization literature \citep[e.g.,][]{agrawal2019mnl,Oh_Iyengar_2021}. }

It is noteworthy that although the random choice $d_{t, m}$ is \emph{conditionally} independent of $d_{t, 1}, d_{t, 2}, \ldots, d_{t, m-1}$ given the event $\{\s_{t, m} \subseteq \s\}$, they are \emph{unconditionally} dependent. 
Indeed, the probability distribution of each customer's choice depends on the available assortment, and the assortment constantly changes throughout an inventory cycle as previous customers' choices deplete the inventory of certain products. 
Due to these stockout events,  customers can display complicated substitution behaviors and the  choices of customers in the same inventory cycle also have complicated dependence. 
Consequently, the joint distribution of customers' choices is generally intractable.   
As we will discuss in \Cref{subsec-clairvoyant-problem,subsec-challenge-algsteps}, this poses serious challenges to our design of algorithms.

At the end of each cycle, we assume that all unsold products are salvaged at given values. This means that the leftovers of each cycle do not constrain the ordering decisions of the next cycle.  If the attraction parameters $\bv$ \emph{were} known, we could decompose the joint assortment-inventory optimization problem over $T$ cycles into $T$ disjoint single-cycle static optimization problems and solve them separately. In our problem, the attraction parameters $\bv$ \emph{are} actually unknown, so we need to estimate them and gradually refine our estimates as we collect more data over the inventory cycles. Accordingly, the decision-making in each cycle needs to be based on data accumulated in previous cycles, so the decision problems in different cycles are connected due to learning attraction parameters. This is intimately related to the exploration and exploitation trade-off in online assortment-inventory decision-making, a key challenge that we tackle in this paper (see \Cref{subsec-naive-extension}). 
We remark that the assumption that inventory leftovers are salvaged at the end of each inventory cycle is common in the literature \citep[see, for example,][]{bensoussan2007multiperiod,ding2002censored,lu2005censored}. 
In practice, this assumption suits particularly well for perishable products, such as newspapers, dairy products, and fresh fruits, and has evolved into a standard operation for many retailers. For example, Bianlifeng, a Chinese brand that operates a chain of more than 2,500 convenience stores, salvages unsold prepared foods and meal kits on a daily basis. 

\mnote{blue}{AEQ3\\R2Q2}{0mm}\edit{Finally, while the assumption of leftover salvage simplifies the problem, the setting in which inventory leftovers can be carried over to subsequent inventory cycles is also important. Therefore, in \Cref{sec:carry over}, we extend our model to take into account inventory carryovers, with which the decision in each cycle could affect the subsequent cycles, making the online optimization problem more complex. We devise an algorithm to address this complexity and show that it can still attain a near-optimal regret rate.}

\subsection{Static Assortment-Inventory Optimization Problem} \label{subsec-clairvoyant-problem}
Next, we describe the one-cycle optimization problem for the retailer, to which we refer as the \textit{static assortment-inventory optimization problem}. Note that in this problem there still exists dynamic demand substitution triggered by stockout events, and we adopt the notion of ``static'' to highlight that in this problem the joint assortment-inventory decision is to be determined once to maximize the one-cycle expected profit. Dropping the subscript indexing cycle, we consider several natural cardinality and capacity constraints for the retailer's decision $\U$. In particular, we assume that at most $K$ different products can be assorted in any inventory cycle. In addition, we consider two types of capacity constraints. The first one requires that the total inventory that can be ordered in one cycle to be no more than $\bar{c}$, and the second restricts the maximum inventory of each product $i$ to $c_i$. In practice, the first type of capacity constraint may correspond to the limited transportation capacity or warehouse space, while the second one may correspond to the maximum shelf space for each separate product. %
Then, the set of feasible inventory decisions, $\setU$, is given by  %
\begin{equation*}
	\setU \coloneqq \left\{\U\mid 0 \leq u_i \leq c_i,\ \forall i\in[N],\ \sum_{i = 1}^{N}u_i \le \C\ \text{and } |S(\U)|\le K\right\} \ .
\end{equation*}
\edit{Note that the constraint set $\, \setU$ is quite general. It can effectively drop any of the hard constraints by setting the corresponding upper bound large enough.}

The objective of the static assortment-inventory optimization problem is to maximize the one-cycle expected profit, consisting of the expected revenue and the expected salvage values, netting the ordering costs. For clarity of exposition, we make the following simplification assumption. 
\begin{assumption} \label{assum-order-salvage}
	The salvage values of all products are equal to their respective ordering costs. 
\end{assumption} 
The assumption implies that the leftovers at the end of each cycle do not contribute to any profit or cost, so the expected profit in the cycle equals the total net profit of only products sold in the cycle. 
 {We make this assumption merely for the exposition simplicity. In Appendix \Cref{sec-relax-assumption}, we relax this assumption and show that our proposed algorithm still works once it is slightly modified.} \mnote{blue}{R1minor3}{-5mm}

To formulate the expected profit in a single cycle, we let $X^{\U,\bv}_{i}$ denote the number of times that product $i$ is purchased in the cycle, when the initial inventory level is $\U$ and the attraction parameter value is  $\bv$. Obviously, for any product $i \in [N]$, the number of total purchases $X^{\U,\bv}_{i}$ cannot exceed the inventory level $u_i$.
It is noteworthy that $X^{\U,\bv}_{i}$ is a random variable because customers' choices are random and their arrival process may also be stochastic. 
In addition, we let $\bre \coloneqq (r_i)_{i \in [N]}$ be a known vector of unit profits of the $N$ products, where $r_i$ represents product $i$'s per-unit selling price netting its per-unit ordering price.
  Without loss of generality, we assume that $r_{\max} \coloneqq \max_{i \in {[N]}}\{r_i\} = 1$. Next, given the vector of attraction parameters $\bv$ and the vector of unit profits $\bre$, for an initial inventory level of $\U$, the \textit{one-cycle expected profit} is
\begin{equation}
	\label{equ-onecycle-revenue}
	R(\U; \bv, \bre) \coloneqq \Eb{\sum_{i \in [N]} r_i X_i^{\U,\bv}},
\end{equation}
where the expectation is taken over the distribution of random variables $X^{\U,\bv}_{i}$ for all $i \in [N]$. Finally, let $\U^*$ denote an optimal solution to the static assortment-inventory optimization problem, that is, 
\begin{equation}	\label{equ-define-ustar}
	\U^* \in \arg\max_{\U\in \setU} \R(\U; \bv, \bre) \ .
\end{equation} 
In other words, $\U^*$ maximizes the expected one-cycle profit given parameters $\bv$ and $\bre$.
The optimal solution $\U^*$ is referred to as a \textit{clairvoyant solution}, and $\R(\U^*; \bv, \bre)$ is known as the \textit{clairvoyant optimal one-cycle expected profit}, both defined in terms of the true attraction parameter $\bv$. 
However, in the context of online optimization, the attraction parameter $\bv$ is not known to the retailer, so she cannot act according to clairvoyant solution $\U^*$ in \eqref{equ-define-ustar}.

Before closing this subsection, we remark that even when the $\bv$ value is given, the problem of evaluating the expected profit in \eqref{equ-onecycle-revenue} for any given initial inventory decision $\U$ is generally intractable \citep{aouad2018greedy}, and the optimization problem in \eqref{equ-define-ustar} is even harder \citep{aouad2018greedy,aouad2022stability}. 
This is because the distribution of $X_i^{\U,\bv}$ is determined by the joint distribution of customers' choices, while the latter is generally intractable due to complicated substitution effects and customer choice's dependence induced by stochastic stockout events (see discussions below 
\eqref{equ-define-MNL-choice-probability}).
 In the remainder of this paper, we will first assume the existence of a black box oracle that can exactly solve problem \eqref{equ-define-ustar} with any given parameters $\bv$ and $\bre$, in order to focus on the exploration-exploitation trade-off. 
Given the exact optimization oracle, we propose an algorithm that can effectively balance exploration and exploitation, and analyze the performance of this algorithm.
 Later in \Cref{sec-oracles}, we further incorporate into our algorithm approximation oracles that can solve problem \eqref{equ-define-ustar} up to  certain error and analyze the corresponding performance guarantees.

\subsection{Online Joint Assortment-Inventory Optimization Problem}
In the online problem setting, the attraction parameter $\bv$ is unknown \textit{a priori}. The retailer needs to dynamically learn the attraction parameters
from the realized customers' choices in each inventory cycle  and adjust the inventory decisions accordingly, aiming to maximize the expected total profit over $T$ inventory cycles. Specifically, the retailer needs to design a policy $\pi \coloneqq  (\pi_1. \pi_2, \cdots, \pi_T)$ that generates the ordering decisions $(\U_1^{\pi}, \U_2^{\pi}, \cdots, \U_T^{\pi})$ according to historical observations and potentially some additional sources of  randomization encoded by a random variable $\hbar$.
The policy $\pi$ is a vector of measurable mappings such that
\begin{equation*}
	\U_t^{\pi} = \begin{cases}
		\pi_1(\hbar), &\text{ for }t = 1,\\
		\pi_t(\U_{t - 1}^{\pi}, \seq_{t - 1}, \U_{t - 2}^{\pi}, \seq_{t - 2}, \cdots, \U_{1}^{\pi}, \seq_1, \hbar), & \text{ for }t = 2, 3, \cdots, T\ .
	\end{cases}
\end{equation*} 
Particularly, when prescribing the inventory decision $\U_t^\pi$ for cycle $t$, the policy can only use historical choice observations and inventory decisions prior to cycle $t$, but not any future information. 
Moreover, the policy has to comply with the inventory constraints so $\U_t^{\pi} \in \setU$ for all $t \in [T]$.
This formally defines an \textit{admissible policy}.
Additionally, let $\mathbb{P}_\pi$ and $\mathbb{E}_\pi{}$ denote the probability distribution and expectation value over the random decision path of policy $\pi$, respectively. 

The expected total profit when applying policy $\pi$ can be written as 
\begin{equation}	\label{equ-define-opt-problem}
	\Epi{\sum_{t = 1} ^ {T}R(\U_t^{\pi}; \bv, \bre)},
\end{equation}
where $\{\U_t^{\pi}\}_{t \in [T]}$ is the sequence of history-dependent inventory decisions generated by policy $\pi$. For simplicity, we drop the superscript ${\pi}$ from $\U_t^{\pi}$ when  it is clear from the context. The objective of the online joint assortment-inventory optimization problem is to identify a policy that maximizes the expected total profit in \eqref{equ-define-opt-problem}. 
This is equivalent to minimizing the \textit{regret}.
Specifically, the regret of a policy $\pi$, denoted by $\Reg(\pi ; \bv, \bre, T) $, measures the    expected total profit loss of policy $\pi$ relative to the optimal policy based on known attraction parameters, that is, 
\begin{align}
	\Reg(\pi ; \bv, \bre, T) \coloneqq & \ \sum_{t = 1}^T \max_{\U_t \in \setU}  R(\U_t; \bv, \bre)  - \Epi{\sum_{t = 1}^T R(\U_t^\pi ; \bv, \bre) } \nonumber \\
	= & 
	\Epi{\sum_{t = 1}^T \left(R(\U^*; \bv, \bre) - R(\U_t^\pi; \bv, \bre)\right)
	} ,   \label{equ-def-regret}
\end{align}
where $\U^*$ is the clairvoyant solution to problem \eqref{equ-define-ustar}. However, it is generally intractable to find an optimal policy that exactly minimizes the regret. Instead, a more realistic goal is to find a policy whose regret grows with $T$ at a slow sub-linear rate \citep[\eg, ][]{lattimore2020bandit,bubeck2012regret}. In the next section, we will propose an algorithm that can provably achieve this goal provided that an exact oracle to static optimization problems of the form in \eqref{equ-define-ustar} is available.

\section{The Online Optimization Algorithm} \label{sec-algorithm}
Since the attraction parameters are unknown, an online optimization algorithm %
needs to fulfill two contradictory tasks, namely, \textit{exploration} and \textit{exploitation}. Exploration refers to employing inventory decisions in order to collect data towards more accurate estimation of unknown parameters, even if the decision appears suboptimal given the latest estimation results. Exploitation refers to making profit-maximizing decisions according to the  parameter estimates obtained from the latest data. 
Both exploration and exploitation can lead to suboptimal decisions and thus result in regret.

In order to achieve a small regret, we propose an exploration-exploitation algorithm based on a non-trivial extension of the legacy upper confidence bound (UCB) algorithm for multi-armed bandit (MAB) problems \citep{auer2002finite}. The UCB algorithm builds on the renowned principle of ``optimism in the face of uncertainty'', that is, it makes sequential decisions based on \emph{optimistic} estimates of the expected rewards/profits of decisions. This optimism incentivizes exploration, and by properly adjusting the level of optimism, we can effectively balance exploration and exploitation to achieve rate-optimal regret.

Although the principle of optimism is well known, implementing it in our joint assortment-inventory optimization problem turns out to be challenging.  
\edit{The stochastic product stockout events pose serious challenges both for the estimation of unknown parameters and the balancing of exploration and exploitation.}
 In this section, we first explain these challenges in \Cref{subsec-challenge-algsteps}.
Then, we discuss the details of estimation in \Cref{subsec-estimator} and balancing exploration and exploitation in \Cref{subsec-naive-extension,subsec-adjust-revenue}. \edit{We further disucss the advantage of our proposed  approach over an MLE-based approach in \Cref{subsec:mle}.}

\subsection{Main Challenges}\label{subsec-challenge-algsteps}
The UCB algorithm builds on a sequence of estimates of the expected profits of decisions, where the expected profits depend on the unknown attraction parameters. Therefore, it is crucial to form increasingly accurate estimates of the attraction parameters. The first challenge arises from the estimation of attraction parameters $\bv$ in the MNL model. Because the inventory is limited, products may run out of stock over time. Whenever stockout events occur, the assortment set available to all subsequent customers and thus their choice probabilities change accordingly.  This means that a customer's demand can be censored if earlier customers already deplete certain products. In turn, the customer's choice may further change the assortment set and impact the choices of later customers. Worse yet, the assortment set depends on customers' random choices and is thus highly unpredictable; in principle, there can be exponentially many possible assortments that an arriving customer can face in a single inventory cycle. As a result, customers' choices display highly complicated substitution behaviors and intricate dependence, so their probability distributions are generally intractable. With such complicated choice observations, accurately estimating the attraction parameters is very challenging. For example, existing estimators in the online assortment optimization literature \citep[see][]{agrawal2019mnl} 
are no longer valid in presence of product stockout events. 
In \Cref{subsec-estimator}, we overcome the challenge by proposing novel consistent estimators for the reciprocals of attraction parameters (\ie, $1/\bv$). These estimators are  based on the observation that certain summary statistics of customers' choices have tractable distributions parameterized by the reciprocals of attraction parameters.

The second challenge is in balancing the exploration and exploitation. 
We adopt a UCB-based algorithm, where we need to form increasingly accurate optimistic estimates for the expected profits of different decisions to properly incentivize exploration.
If there were no inventory control and only assortment decision was needed, then the expected profit could be shown to be monotone in the attraction parameters. Thus optimistic profit estimates could be easily formed by evaluating the expected profits at upper confidence bounds on the attraction parameters $\bv$, as suggested by \cite{agrawal2019mnl}. 
However, \edit{this monotonicity is invalid for general joint assortment-inventory optimization problems, because of complicated dependence of customer choice distributions on the attraction parameters}.
We show in a counterexample  that a heuristic approach directly motivated from \cite{agrawal2019mnl} may not give sufficiently optimistic profit estimates  so it can  lead to  very high regret due to lack of exploration (see \Cref{example-counter-example} in \Cref{subsec-naive-extension}). 
In \Cref{subsec-adjust-revenue}, 
we propose a novel approach that further tunes up the products' unit profits $\bre$ to incentivize stronger exploration. 
Importantly, we adaptively tune the unit profits according to the estimation errors of attraction parameters, so that we can effectively balance the exploration and exploitation.

At last, as explained in \Cref{subsec-clairvoyant-problem}, it is also challenging to evaluate and optimize the static assortment-inventory problem \eqref{equ-define-ustar} with any given values of attraction parameters and unit profits. 
For simplicity, we put aside this additional challenge for now and assume the existence of an exact oracle to solve static assortment-inventory problems. 
This enables us to focus on the central challenges of estimation and exploration-exploitation in online decision-making. 
We will relax this assumption and incorporate more practical approximate static optimization oracles in \Cref{sec-oracles}.

\subsection{Estimation of MNL Model Parameters}\label{subsec-estimator}
\mnote{blue}{New Method\\AEQ1\\R1Q1a}{0mm}In this section, we discuss the estimation of attraction parameters $\bv$ in the MNL model. As explained above, this is a very challenging task, because the assortment set can dynamically change due to stochastic stockout events. As a result, the choice probability for each product can dynamically change in an unpredictable manner, and customers' choices can be censored by limited inventory. In this setting, existing estimators proposed in \cite{agrawal2019mnl} for online assortment optimization without inventory limit are not applicable. In this section, we propose a novel method to consistently estimate the attraction parameters based on certain summary statistics carefully constructed from the choice observations. 
\edit{This involves tracking the number of no-purchases between consecutive purchases of each product when the product is available. With these summary statistics, we can update the estimates of MNL parameters at the end of each cycle.}

\mnote{blue}{R1Q2}{0mm}\edit{We collect the summary statistics for each product separately. For each product $i \in [N]$, we only consider the subsequence of data consisting of both purchases of product $i$ and non-purchases, when product $i$ is assorted. Then we count the number of no-purchases $\mu_{i,k}$ between the $(k-1)$-th and the $k$-th purchase of product $i$ in the subsequence. The summary statistics $\mu_{i, k}$ for $k = 1, 2, \dots$ will be used to update the estimate of $v_i$, at the end of inventory cycles. 
Notably, this counting process for product $i$ runs across inventory cycles, irrespective of the counting for other products. If product $i$ is not assorted in some inventory cycle, its counting process pauses,  but the counting resumes once product $i$ is assorted again. If there is no purchase for product $i$ in a cycle, there will be no estimation update for $v_i$ at the end of that cycle.
The following example illustrates the process.}

\begin{example}
\edit{We illustrate the summary statistics in an example with two inventory cycles and two products.
    Suppose that the retailer orders up to $\U_1=(1,1)$ and $\U_2 = (1,2)$ in the first and second inventory cycles, respectively. Figure~\ref{fig:estimation process} depicts a  sample sequence of customer choices. 
    When product $1$ is first purchased, we count the number of no-purchases before this purchase and while product $1$ was assorted. According to the sample sequence, this number, denoted by $\mu_{1,1}$, is $1$. When product $1$ is purchased the second time, we count the number of no-purchases, denoted by $\mu_{1,2}$, between its first and second purchases and while it was assorted. According to our example, only one such no-purchase occurred at the beginning of cycle 2, and thus $\mu_{1,2} = 1$. 
    Similarly, when product $2$ is first purchased, we count the number of no-purchases that occurred while product $2$ was assorted but before it was first purchased. There are three such no-purchases, two in the first cycle and one in the second cycle. Consequently, $\mu_{2,1} = 3$. When product $2$ is purchased the second time, we count the number of no-purchases between its first and second purchases and while it was assorted. There was no such no-purchase, and thus $\mu_{2,2} = 0$. Finally, at the end of cycle 1, we update the parameter estimate for only product 1, and at the end of cycle 2 we update those for both products 1 and 2.}
\begin{figure}[htbp]
	\begin{center}
		\includegraphics[height=2.in]{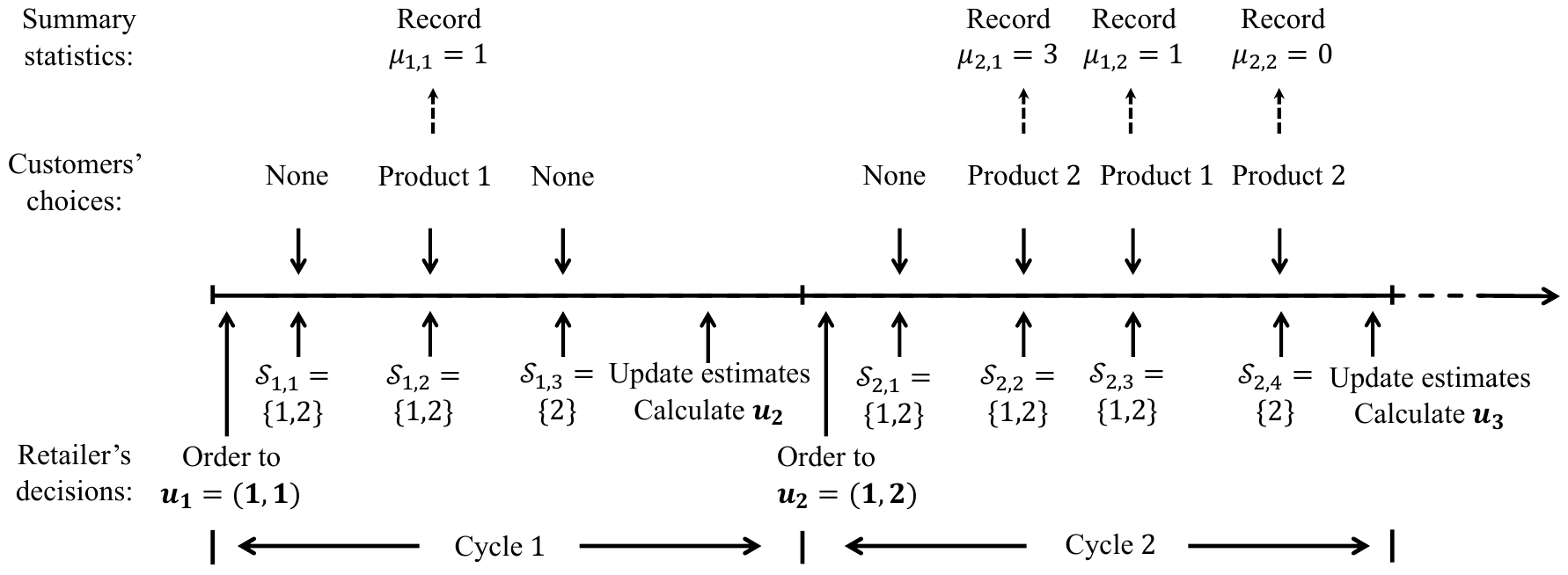}
		\caption{An example of the estimation procedure.} \label{fig:estimation process}
	\end{center}
\end{figure}
\end{example}

To motivate our estimator, we first establish the distributions of the summary statistics $\{\mu_{i,k}\}$.
\begin{restatable}{lemma}{estimatorrecvnew} 
\label{lemma:estimator-recv}
	\edit{For any product $i\in [N]$, $\mu_{i, k}$'s are independent and identically distributed geometric random variables with parameter $v_{i} / (1 + v_{i})$, whose distribution and expectation are given as follows}:
	\begin{equation}\label{eq: geometric mean}
	\begin{aligned}
	\prob{\mu_{i, k} = \mu} &= \left(\frac{1}{1 + v_{i}}\right)^{\mu} \frac{v_{i}}{1 + v_{i}}, \ \mu = 0, 1, 2, \dots, \\
	\Eb{\recv_{i, k}} &= \recv_{i} \coloneqq \frac{1}{v_i}. 
	\end{aligned}
    \end{equation}
    
\end{restatable}

In \Cref{lemma:estimator-recv}, we prove that the summary statistic $\mu_{i, k}$ for each product $i$ follows a geometric distribution, whose expectation equals the reciprocal of the corresponding attraction $v_i$. 
This expectation is well-defined within the range between $\recv_{\min} \coloneqq 1 / v_{\max}$ and $\recv_{\max} \coloneqq 1 / v_{\min}$. 
Intuitively, $\recv_{i, k}$ reflects the attraction of the no-purchase option relative to the assorted product $i$, so not surprisingly its expectation equals the ratio of the attraction $v_0 = 1$ over $v_i$. 
Importantly, this ratio is independent of the assortment set, so the distribution of the summary statistic $\recv_{i, k}$ remains fixed and simple even with a dynamically changing assortment set. 
By leveraging the summary statistics,   
we sidestep the complication brought by stochastic stockout events.

Motivated by \Cref{lemma:estimator-recv}, we propose to estimate the reciprocal attraction $\mu_i = 1/v_i$ of each product $i$ by averaging all  
summary statistics available for that product.
\edit{Formally, at the end of cycle $t-1$, we consider for each product $i$ the $\mu_{i,k}$'s collected by that time, with the total number of such summary statistics  denoted by $k_{i, t}$}. 
\edit{According to \Cref{eq: geometric mean}, each $\mu_{i, k}$ for $k \in [k_{i,t}]$ provides an unbiased observation for $\mu_i$.} Thus we average all of them to obtain the following estimator for $\mu_i$: 
\begin{equation}\label{eq: mu-est}
	\bar \recv_{i,t} \coloneqq \frac{1}{k_{i, t}}\sum_{k \in [k_{i, t}]} \recv_{i, k}\ . 
\end{equation} 
Accordingly, we can estimate the attraction parameter $v_i = 1/\mu_i$ by $\bar{v}_{i, t} \coloneqq 1/ \bar \recv_{i,t}$. 
\edit{We refer to our new estimator as \textit{Counting Estimator} (CE), as it is based on \emph{counting} summary statistics. Below we bound CE's estimation error.}
\begin{restatable}{lemma}{barmuconverge}
	\label{lemma:convergence-barmu}
	\edit{For product $i$  and any cycle $t$ such that $Q_{i, t} \coloneqq  \log (\sqrt{Nt\sum_{s=1}^{t-1}M_s} + 1)/k_{i, t} < 1/48$,}
	\begin{equation}
		\label{eq:concentration-bound-true-mu}
		\mathbb{P}\left(\mid  \bar \recv_{i, t} - \mu_i \mid \ge \max\{\sqrt{\mu_i}, \mu_i\} \sqrt{24Q_{i, t}} + 48Q_{i, t}\right) \le \frac{4}{Nt} \ .
	\end{equation}
\end{restatable}

\Cref{lemma:convergence-barmu} establishes that $\bar \recv_{i,t}$ is a \emph{consistent} estimator for the  parameter $\mu_{i} = 1/v_i$, as the estimation error of $\bar \recv_{i,t}$ vanishes to zero when $k_{i, t}$ grows indefinitely and \edit{$Q_{i, t}$ vanishes to $0$}.  
The error bound in \Cref{eq:concentration-bound-true-mu} quantifies the uncertainty of the estimator, which will be useful in designing our exploration-exploitation algorithm (see \Cref{subsec-naive-extension} \Cref{lemma: coverage-probability}).
\Cref{lemma:convergence-barmu} also ensures the consistency and error bound of the attraction parameter estimator $\bar{v}_{i, t}$. 
\edit{Note that the error bound in \Cref{lemma:convergence-barmu} relies on product $i$ being purchased sufficiently many times by the end of cycle $t-1$ (\ie, large enough $k_{i, t}$) so that $Q_{i, t} < 1/48$}. 
We refer to this condition as the \emph{adequate exploration} condition. 
However, the expected regret incurred when some products are not adequately explored is negligible, as detailed in the proof of \Cref{thm-main} below. 

\Cref{lemma:convergence-barmu} shows that our proposed CE is valid despite the ever-changing assortment set due to stockout events. This is possible because the distributions of our summary statistics $\mu_{i, k}$'s are free of the complication caused by stockout events. We note that \cite{agrawal2019mnl} estimate MNL parameters by some alternative summary statistics for an online assortment optimization problem. Their strategy has an epoch structure: they repeatedly offer the same assortment to customers in an epoch, and end the epoch as soon as a no-purchase occurs. Then the summary statistic for each product in an epoch is the number of purchases of that product within the epoch, and it is shown to be an unbiased observation for the corresponding attraction parameter. Crucially, this unbiasedness holds because there is no inventory limit in their setting and the same assortment is offered to each customer in an epoch. 
However, in our setting with limited inventory, the assortment set can change stochastically due to stockout events. In particular, customers' demands for the products can be censored by the limited inventory. If some products become out of stock before the first no-purchases occur, then their purchases cannot accurately reflect customers' true preferences. 
So the estimator in \cite{agrawal2019mnl} is not effective in our problem setting. %
In contrast, our summary statistics $\mu_{i, k}$'s count no-purchases when a product $i$ is available so they are never censored by the inventory.

\subsection{The Failures of Directly Extending the UCB Algorithm}
\label{subsec-naive-extension}

In this subsection, we outline the main idea of balancing exploration and exploitation in our problem by a direct extension of the UCB algorithm for MAB problems \citep{auer2002finite}. We will show that this extension, although intuitively simple, is in general computationally intractable. 
In particular, it cannot be remedied by an appealing heuristic approach stemming from the existing literature, since we show in a counterexample that this heuristic cannot incentivize sufficient exploration. 
This counterexample provides valuable insights that motivate our novel algorithm in the next subsection.

To extend the UCB algorithm to our problem, we make inventory decisions according to certain optimistic estimates for the expected profits of all possible decisions. 
That is, we construct high-confidence upper  bounds on the true expected profits $R(\U; \bv, \bre)$ for all feasible inventory levels $\U \in \setU$ from the data available in each cycle, and then choose the inventory level that achieves the highest upper confidence bound. 
The optimistic estimates given by the upper bounds incentivize us to explore inventory decisions whose expected profits are highly uncertain. Moreover, the optimistic estimates should gradually approach the true expected profits as more data are collected, so that we can also exploit the knowledge revealed by the observed data.

Before forming UCBs on the expected profits, we first note that we can form confidence bounds on the unknown attraction parameters using the proposed estimators $\bar \recv_{i,t}$'s  in \Cref{eq: mu-est}. 
Specifically, define the following confidence bounds on parameter $\mu_i$ based on data collected in before cycle $t$: 
\begin{align}\label{eq:define-recv-LCB}
\recv_{i, t}^{\LCB} \coloneqq \max\left\{\recv_{\min}, \ \bar \recv_{i,t} - \radius_{i, t}\right\}, ~~ \blue{\recv_{i, t}^{\UCB} \coloneqq \ \bar \recv_{i,t} +\radius_{i, t}} \ ,
\end{align}
where \edit{$\radius_{i, t}$ is the confidence radius given by $\radius_{i, t} = \max\{\sqrt{\bar \recv_{i,t}}, \bar \recv_{i,t}\}\sqrt{48Q_{i, t}} + 48Q_{i, t}$}. 
 {Obviously, when product $i$ is not sufficiently explored before cycle $t$ (\ie, $Q_{i, t} < 1/48$), we have $\recv_{i, t}^{\LCB}\coloneqq \recv_{\min} = 1/v_{\max}$. Note that the algorithm does not require $\recv_{\max} = 1/v_{\min}$ so we do not need to specify $v_{\min}$.} The reciprocals of the confidence bounds in \Cref{eq:define-recv-LCB} give confidence bounds for $v_i = 1/\mu_i$:
\begin{align}\label{equ-define-v-LCB}
v_{i,t}^{\LCB} \coloneqq \frac{1}{\recv_{i, t}^{\UCB}}\ , ~~ v_{i,t}^{\UCB} \coloneqq \frac{1}{\recv_{i, t}^{\LCB}}\ . 
\end{align}
Note that the confidence radius $\radius_{i, t}$ is nearly identical to the vanishing error bound in \Cref{lemma:convergence-barmu}, except that $\radius_{i, t}$ only involves observable quantities.  
The confidence radius also vanishes to zero as the data size $k_{i, t}$ grows indefinitely, and thus the above confidence bounds tightly concentrate around the true parameter values when a large amount of data are available. 
By slightly revising \Cref{lemma:convergence-barmu}, we can show that the confidence bounds proposed above have high coverage probabilities. 
\begin{restatable}{lemma}{coverageprob}
\label{lemma: coverage-probability}
\edit{For product $i$  and any cycle $t$ such that $Q_{i, t} < 1/48$},  we have 
\begin{align*}
\mathbb{P}\left(\recv_{i,t}^{\LCB} \le \mu_i \le \recv_{i,t}^{\UCB}\right) = \mathbb{P}\left(v_{i,t}^{\LCB} \le v_i \le v_{i,t}^{\UCB}\right) \ge 1 - \frac{6}{Nt}\ . 
\end{align*}
\end{restatable}

With the confidence bounds above, a direct extension of the UCB algorithm would pick the inventory level $\tilde{\U}_{t}$ for cycle $t$  as follows:
\begin{equation}\label{equ-policy-ucb-primal}
    \tilde{\U}_{t} \in \argmaxx_{\U \in \setU}\tilde{R}(\U),  ~~ \text{ where } \tilde{R}(\U) \coloneqq \max_{\tilde{v}_{i} \in [\vLCB{i, t} , \vUCB{i, t}] , \forall i \in [N]} R(\U; \tilde{\bv}, \bre)\ . 
\end{equation}
Apparently, \Cref{lemma: coverage-probability} implies that $\tilde{R}(\U)$ as defined in \eqref{equ-policy-ucb-primal} upper bounds the true expected profit $R(\U; {\bv}, \bre)$ for any inventory level $\U \in \setU$ with high probability. Moreover, the upper bound $\tilde{R}(\U)$  approaches the corresponding true expected profit as the data size grows to infinity, since in the limit the confidence bounds $v_{i,t}^{\LCB}, v_{i,t}^{\UCB}$'s  all converge to the true values of the attraction parameters. 
So we can expect the decision $\tilde{\U}_{t}$ to converge to the clairvoyant optimal decision $\U^*$ in \eqref{equ-define-ustar} as $t$ grows. 
Thus, \eqref{equ-policy-ucb-primal} would give a reasonable UCB algorithm if we could implement it in practice. 

Unfortunately, the optimization problem in \eqref{equ-policy-ucb-primal} is generally intractable, except in extreme cases. This is because the expected profit $R(\U; \tilde{\bv}, \bre)$, as a function of the value $\tilde{\bv}$ of attraction parameters, is generally highly complex, so maximizing it over the values of the attraction parameters within the confidence bounds is very difficult. This is a particularly grave challenge noting that the evaluation and optimization of $R(\U; \tilde{\bv}, \bre)$ at even a single $\tilde{\bv}$ value is already difficult \citep{aouad2018greedy,aouad2022stability}. Because of this computational intractability, directly following \Cref{equ-policy-ucb-primal} is not practical for general online assortment-inventory decision-making. 

A special case where \eqref{equ-policy-ucb-primal} can be efficiently solved is when only a single customer arrives in each inventory cycle (\ie, $M_t = 1$). 
In this special case, the inventory limit is vacuous, and our online joint assortment-inventory optimization problem degenerates to the MNL-bandit problem in \cite{agrawal2019mnl}.
The profit function $R(\U; \tilde{\bv}, \bre)$ in this special setting is  increasing in $\tilde \bv$ (Lemma A.3 in \cite{agrawal2019mnl}), so  $\tilde u_{t} \in \argmax_{\U\in\setU}\tilde R(u) = \argmax_{\U\in\setU} R(\U; \bv_{t}^{\UCB}, \bre)$, where $\bv_{t}^{\UCB} = (v_{i, t}^{\UCB})_{i \in [N]}$ is the upper confidence bound on $\bv$. 
The latter optimization problem under $M_t = 1$ is also polynomial-time solvable. This constitutes a key step of the MNL-bandit algorithm in \cite{agrawal2019mnl}.

One may wonder if we can implement the same idea as a heuristic approach for more general settings. That is, even when $M_t$'s take general deterministic or stochastic values, we still implement 
\begin{align}\label{eq: heuristic}
\tilde \U_{t}^{\text{Heur}} \in \argmax_{\U\in\setU} R(\U; \bv_{t}^{\UCB}, \bre).
\end{align} 
This corresponds to a static assortment-inventory optimization problem \eqref{equ-define-ustar} with parameters $\bv^{\UCB}_{t}$ and $\bre$, for which we assume the existence of optimization oracles. 
However, this heuristic approach actually does not work, because the  expected profit function $R(\U; \tilde{\bv}, \bre)$ is in general not monotone in $\tilde{\bv}$, so $R(\U; \bv_{t}^{\UCB}, \bre)$ may not be a valid profit upper bound to incentivize enough exploration. 
Below we show the failure of this heuristic approach in a simple counterexample.

\begin{example}\label{example-counter-example}
	Consider a small-scale instance of the online optimization problem, where the number of products $N$ is 2, the product-wise capacity vector $\cvec$ is $(1, 1)$, the total capacity $\C$ is 2, and the numbers of customer arrivals $\M_1, \dots, M_T$ are all equal to 2. Suppose that the attraction parameters of the no-purchase option and product $1$ are $v_0 = v_1 = 1$, and the unit profit of product $1$ is $r_1 = 1$. The attraction parameter $v_2$ and the unit profit $r_2$ of product $2$ are left undetermined.

	There are four feasible inventory decisions: $\U^{(0) }= (0, 0)$, $\U^{(1) }= (1, 0)$, $\U^{(2)} = (1, 1)$, and $\U^{(3)} = (0, 1)$, where $\U^{(0)}$ is trivially sub-optimal and  $\U^{(3)}$ can be shown to be dominated by $\U^{(2)}$.  
	In \Cref{app-counter-example}, we show that there exist a range of values of $v_2$ and $r_2$ such that $R(\U^{(2)}; \bv, \bre) \ge R(\U^{(1)}; \bv, \bre)$, \ie, the inventory decision $\U^{(2)} = (1, 1)$ that assorts both products is optimal.
	However, if the initial upper confidence bound $\bvUCB{1}$ on the attraction parameters are high enough, as they would be due to the lack of data at the beginning, then we will have  $R(\U^{(2)}; \bvUCB{1}, \bre) < R(\U^{(1)}; \bvUCB{1}, \bre)$. This means  the heuristic approach in \eqref{eq: heuristic} starts with implementing the inventory decision $\U^{(1)}= (1, 0)$. Since this decision $\U^{(1)}$ does not assort product 2, the upper confidence bound of $v_2$ cannot be updated. Consequently, the algorithm will persistently select $\U^{(1)}$ throughout, resulting in linear total regret in the long run. In 
	 \Cref{app-counter-example}, we provide all calculation details and also show that the dependence of $R(\U; \bv, \bre)$ on $\bv$ is not monotonic. 
\end{example}

As shown in \Cref{example-counter-example}, the heuristic approach in \eqref{eq: heuristic} may not incentivize sufficient exploration and thus miss the optimal decision altogether, despite that it significantly eases the computational burden of the exact approach in \eqref{equ-policy-ucb-primal}. 
In the next subsection, we propose a novel algorithm that modifies the heuristic approach, so that we can effectively balance exploration and exploitation while still maintaining the same level  of computational costs.

\subsection{Proposed Exploration-Exploitation Algorithm} \label{subsec-adjust-revenue}
Recall that a direct extension of the UCB algorithm shall implement the optimistic inventory decision prescribed by \eqref{equ-policy-ucb-primal}. 
However, the optimization problem is intractable to solve.
The heuristic approach in \eqref{eq: heuristic}, although computationally easier, cannot incentivize sufficient exploration. 
This is because the heuristic objective function $R(\U; \bv_{t}^{\UCB}, \bre)$ is generally not an upper bound on the unknown true profit $R(\U; \bv, \bre)$ and thus fails to provide valid optimistic profit estimates. 
In this subsection, we propose a novel algorithm that further adjusts the heuristic objective function. 
The resulting new objective  can provide desirable optimistic profit estimates, thereby enabling us to effectively balance the exploration and exploitation.

In our proposed algorithm, we consider not only the upper confidence bound $\bv_{t}^{\UCB}$ on the \emph{unknown} attraction parameters $\bv$, but also an upper bound $\hat \bre_t$ on the \emph{known} unit profits $\bre$ in each cycle. Specifically, we define $\hat \bre_t$ as an $N$-dimensional vector with its $i^{th}$ element given by 
\begin{align}
	\label{equ-define-rhat}
	\blue{\hat r_{i, t}} \coloneqq &\ \begin{cases}
	    1, &\blue{\text{if }\ Q_{i, t} > 1/48},\\
     \min\{1, r_i + \delta_{i, t} \}, &\ \text{otherwise},
	\end{cases} \quad
 \text{ where } \delta_{i, t} \coloneqq \ \frac{\recv_{i, t}^{\UCB}}{\recv_{i, t}^{\LCB}}  - 1 . 
\end{align}

Note that $\delta_{i, t} \le \ \frac{2\Delta_{i, t}}{\recv^{\LCB}_{i, t}}$, and it binds when $\recv_{\max} \ge \bar \recv_{i,t} +\radius_{i, t}$ and $\recv_{\min} \le \bar \recv_{i,t} - \radius_{i, t}$. According to~\eqref{equ-define-rhat}, each $\hat r_{i, t}$ tunes up the corresponding unit profit $r_i$ by an amount $\delta_{i, t}$ determined by the relative magnitude of the confidence radius $\Delta_{i, t}$.  
By tuning up the unit profits, we can unambiguously bring in more optimism, since the expected profit $R(\U; \bv, \bre)$  increases linearly  with the unit profits $\bre$ (see \Cref{equ-onecycle-revenue}). 
Importantly, we tune up the profits judiciously according to the confidence radius $\Delta_{i, t}$, so that we encourage more optimism for under-explored products whose attraction parameters have higher uncertainty. 
Below we show that this additional profit tuning can indeed remedy the problem of the heuristic objective function in \eqref{eq: heuristic}. 

\begin{restatable}{lemma}{policyucb}\label{lemma-policyUCB}
    For any cycle $t$, if $\bv = (v_i)_{i \in [N]}$, $\bv^{\LCB}_{t} = (v_{i, t}^{\LCB})_{i \in [N]}$, $\bv^{\UCB}_{t} = (v_{i, t}^{\UCB})_{i \in [N]}$ satisfy $\vLCB{i, t} \le v_i \le \vUCB{i, t}$ for all $i \in [N]$, then $R(\U;\bv_{t}^{\UCB}, \hat \bre_{t})\ge R(\U; \bv, \bre)$ for all feasible inventory decisions $\U \in \setU$. 
\end{restatable}

\Cref{lemma-policyUCB} shows that the expected profit $R(\U;\bv_{t}^{\UCB}, \hat \bre_{t})$, when evaluated at \emph{both} the parameter upper bound $\bv^\UCB_{t}$ and the unit profit upper bound $\hat \bre_{t}$, provides an upper bound on the true expected profit. 
Moreover, this upper bound is expected to be increasingly tight as $t$ grows,  since both $\bv_{t}^{\UCB}$ and $\hat \bre_{t}$  converge to the true $\bv$ and $\bre$ respectively.
So $R(\U;\bv_{t}^{\UCB}, \hat \bre_{t})$ can provide valid optimistic estimates that meet all desiderata to constitute a valid UCB algorithm. 
Therefore, we propose to make inventory decisions by solving the following optimization problem for each cycle $t \in [T]$: 
\begin{equation}\label{equ-opt-epoch}
	\U_{t} \in \argmaxx_{\U \in \setU} \R(\U; \bv^{\UCB}_{t}, \hat \bre_{t}).
\end{equation} 

\mnote{blue}{R1Q1b}{0mm}\edit{Notably, \Cref{equ-opt-epoch} incorporate bounds on both the unknown attraction parameters and the known unit profits. However, these two bounds are fundamentally different. The attraction parameter bounds are derived from an analysis of errors in estimating the attraction parameters from choice data. In contrast, the unit profits are known and do not need estimation, so the corresponding bounds are not derived for estimation purposes. Instead, they serve as a tool to bring in more optimism, compensating for the insufficient exploration that could result from relying solely on the attraction parameter bounds.}

\edit{Furthermore, we remark that our proposal 
above, akin to the heuristic approach in \eqref{eq: heuristic},  also corresponds to a static assortment-inventory optimization problem \eqref{equ-define-ustar} for which we assume solution oracles exist. This means that our exploration-exploitation framework is versatile and can seamlessly integrate with any static optimization oracle (see discussion in \Cref{sec-oracles}). Such flexibility is a great advantage of our proposal. We will rigorously prove that this proposal can achieve a nearly optimal regret rate in the next section.
In this section, we only briefly validate the effectiveness of our algorithm in \Cref{example-counter-example}, the counterexample we previously used to demonstrate the failure of the heuristic approach.}

\begin{continuance}{\Cref{example-counter-example}}
In \Cref{example-counter-example}, the heuristic approach only uses the attraction parameters' upper confidence bounds in its objective function. 
This  approach may end up with only choosing the sub-optimal inventory decision $\U^{(1)} = (1, 0)$ and never exploring the second product.
Now with our novel algorithm, in the first cycle we additionally tune  $\bre$ up to $\hat \bre_1 = (r_{\max}, r_{\max})$. We can show that $R(\U^{(2)}; \bvUCB{1}, \hat\bre_1) > R(\U^{(1)}; \bvUCB{1}, \hat\bre_1)$ even if $\bvUCB{1}$ is very high, so our proposed algorithm implements the decision $\U^{(2)} = (1, 1)$ and  successfully explores both products in the first cycle.
After collecting the data in the first cycle, our algorithm updates the upper bounds on attraction parameters and unit profits, and continue to explore decisions with high uncertainty in all subsequent cycles. 
\end{continuance}

The effectiveness of our algorithm crucially rests on the conclusion  in \Cref{lemma-policyUCB}. We now outline the proof of \Cref{lemma-policyUCB} to provide a deeper understanding of our proposed algorithm. 
Define  $\tilde{\bv}_{t}(\cdot)$ and $\tilde{\bre}_{t}(\cdot)$ as maps from any vector $\balpha = (\alpha_1,\cdots, \alpha_N ) \in [0, 1]^N$ to $[v_{\min},v_{\max}]^N$ and $[0, r_{\max}]^N$ respectively, with their $i$-th components given by 
\begin{equation}\label{eq: path}
    \tilde{v}_{i, t}(\boldsymbol{\alpha}) \coloneqq (1 - \alpha_i)\cdot v_{i, t}^{\LCB} +\alpha_i \cdot v_{i, t}^{\UCB} \textrm{, and } \tilde{r}_{i, t}(\boldsymbol{\alpha}) \coloneqq \min\{1, r_i + \alpha_i \cdot\delta_{i, t}\}.
\end{equation}
We can consider the expected profit $R(\U;  \tilde{\bv}_{t}(\boldsymbol{\alpha}), \tilde{\bre}_{t}(\boldsymbol{\alpha}))$ parameterized by any vector $\balpha \in [0, 1]^N$. 
In particular, our proposed objective function $\R(\U; \bv^{\UCB}_{t}, \hat \bre_{t})$ in \eqref{equ-opt-epoch} 
is a special example given by the all-one vector $\balpha = (1, \dots, 1)$. 
In fact, it is the maximum of $R(\U;  \tilde{\bv}_{t}(\boldsymbol{\alpha}), \tilde{\bre}_{t}(\boldsymbol{\alpha}))$ over $\balpha \in [0, 1]^N$, since we can show that the  expected profit $R(\U;  \tilde{\bv}_{t}(\boldsymbol{\alpha}), \tilde{\bre}_{t}(\boldsymbol{\alpha}))$ is monotone in $\balpha$. 

\begin{restatable}{lemma}{monoinalpha}\label{lemma-mono-alpha}
    Let $\boldsymbol{\alpha}_1$ and $\boldsymbol{\alpha}_2$ be two vectors in $[0, 1]^N$ such that  $\boldsymbol{\alpha_1} \le \boldsymbol{\alpha_2}$. Then for any inventory decision $\U \in \setU$ and any cycle $t \in [T]$, we have $R(\U;  \tilde{\bv}_{t}(\boldsymbol{\alpha}_1), \tilde{\bre}_{t}(\boldsymbol{\alpha}_1))\le R(\U;  \tilde{\bv}_{t}(\balpha_2), \tilde{\bre}_{t}(\balpha_2))$.
\end{restatable}

\Cref{lemma-mono-alpha} asserts that, although $R(\U;\bv, \bre)$ is not necessarily monotone in $\bv$ with a fixed $\bre$, $R(\U;\tilde{\bv}(\boldsymbol{\alpha}), \tilde{\bre}(\boldsymbol{\alpha}))$ is monotone in ${\boldsymbol{\alpha}}$ as $\tilde{\bv}(\boldsymbol{\alpha})$ and $\tilde{\bre}(\boldsymbol{\alpha})$ vary simultaneously with ${\boldsymbol{\alpha}}$. \mnote{blue}{R1Q1b}{0mm}\edit{The main idea of the proof of \Cref{lemma-mono-alpha} is to leverage certain stability property of the expected demand in the attraction parameters %
to demonstrate that any possible decrease in $R(\U; \bv, \bre)$ due to an increase in $\bv$  can be offset by an appropriate increase in $\bre$ as described in \eqref{eq: path}}. Given \Cref{lemma-mono-alpha}, we can reformulate our proposal in \eqref{equ-opt-epoch} as follows: 
\begin{equation} \label{equ-maxmax}
    \U_{t} 	\in   \argmaxx_{\U \in \setU} \max_{\boldsymbol{\alpha}\in[0, 1]^N} R(\U;  \tilde{\bv}_{t}(\boldsymbol{\alpha}), \tilde{\bre}_{t}(\boldsymbol{\alpha})).
\end{equation}

It is now easy to explain why \Cref{lemma-policyUCB} is true. Given that $v_i \le \vUCB{i, t}$ for all $i \in [N]$,  there exists a vector $\boldsymbol{\tilde \alpha}_{t} \in [0, 1]^N$ such that $\tilde{\bv}_{t}(\boldsymbol{\tilde \alpha}_{t}) = \bv$. Then the conclusion of \Cref{lemma-policyUCB} follows from 
\begin{align*}
R(\U; \bv_{t}^{\UCB}, \hat\bre_{t}) \ge R(\U; \tilde{\bv}_{t}(\boldsymbol{\tilde \alpha}_{t}), \tilde{\bre}_{t}(\boldsymbol{\tilde \alpha}_{t})) = R(\U; \bv, \tilde{\bre}_{t}(\boldsymbol{\tilde \alpha}_{t})) \ge R(\U;\bv, \bre).
\end{align*}
Here the leftmost inequality holds because $\R(\U; \bv^{\UCB}_{t}, \hat \bre_{t}) = \max_{\boldsymbol{\alpha}\in[0, 1]^N} R(\U;  \tilde{\bv}_{t}(\boldsymbol{\alpha}), \tilde{\bre}_{t}(\boldsymbol{\alpha}))$. The rightmost inequality holds because $R(\U;\bv, \bre)$ linearly increases with $\bre$ and $\tilde{\bre}_{t}(\boldsymbol{\tilde \alpha}_{t}) \ge \bre$. 
In summary, \Cref{lemma-policyUCB} follows from the monotonicity of $R(\U;\tilde\bv, \tilde\bre)$ as $\tilde\bv$ and $\tilde\bre$ vary along the path specified in \Cref{eq: path}.

We remark that our proposal of tuning known parameters (\ie, the unit profits) appears an unconventional exploratory strategy, as existing online decision-making literature typically tunes values of unknown parameters \cite[e.g.,][]{lattimore2020bandit}. Our proposal presents a novel 
 way to incentivize exploration, which we could be useful in other online decision-making problems, especially when the dependence of the objective function on unknown parameters is complex but its dependence on known parameters is  simple. 

\subsubsection*{Summary of the proposed algorithm.}
We finally summarize our proposed exploration-exploitation algorithm for  online joint assortment-inventory optimization  in Algorithm~\ref{alg-main}. 
This algorithm puts together the estimators in \eqref{eq: mu-est}, the parameter confidence bounds in \eqref{eq:define-recv-LCB}, \eqref{equ-define-v-LCB}, and the decision rule in \eqref{equ-opt-epoch}. 
In Appendix \Cref{app-sec-alg-step}, we further provide a step-by-step explanation to Algorithm~\ref{alg-main}.

\begin{algorithm}[!ht]
	\label{alg-main}
    \small
	\DontPrintSemicolon
  \KwInit{$\vUCB{i,1} = v_{\max}$,\ $\hat\re_{i,1} = r_{\max}$,\ $n_i = 0$,\ $k_{i} = 0$,\ 
  $\forall i\in [N]$\; 
  $\bv_{1}^{\UCB} = (v_{1,1}^{\UCB}, \dots,  v_{N,1}^{\UCB})$,\ $\hat\bre_1 = (\hat r_{1, 1}, \dots, \hat r_{N, 1})$,\ 
  $\U_{1} \in \arg\max_{\U\in \setU}R(\U; \bvUCB{1}, \hat\bre_{1})$,\  $t = 1$}
	\While{$t < T$}
	{
		Order up to $\U_t$, and observe the choices  $\seq_t = (d_{t, m})_{m \in [M_t]}$ made by $M_t$ customers in  cycle $t$\;
		 \For{$m \gets1$ \KwTo $M_t$} {
        \label{alg-startWhile}
			\eIf{$d_{t, m} = 0$}
			{\For{$i\in\s_{t,m}$}{$n_i = n_i + 1$}
                } 
                {$i = d_{t,m}$,\  
                $k_i = k_i + 1$,\ 
                $\recv_{i, k_i} = n_i$,\ 
                $n_i = 0$\;
                }
		}{\label{alg-end-while}}
        $t = t + 1$ \;
        \For{$i\gets1$ \KwTo $N$}{
        \label{alg-all-prod}
        Update $k_{i,t} = k_i$\;
        Compute $\bar \recv_{i,t}$ according to \Cref{eq: mu-est} \;
        Compute $\recv_{i, t}^{\LCB}, \recv_{i, t}^{\UCB}, v_{i,t}^{\LCB}, v_{i,t}^{\UCB}$ per \Cref{eq:define-recv-LCB,equ-define-v-LCB}, respectively\;
        Compute $\hat \re_{i, t}$ according to \Cref{equ-define-rhat}\;}
        \label{alg-all-prod-2}
        Set $\bv_{t}^{\UCB} = (v_{1,t}^{\UCB}, \dots,  v_{N,t}^{\UCB})$ and $\hat\bre_t = (\hat r_{1, t}, \dots, \hat r_{N, t})$ \;
         Compute $\U_{t} \in \arg\max_{\U\in \setU}R(\U; \bvUCB{t}, \hat\bre_{t})$\;
         \label{alg-calculate-u}
	}		
	
	\caption{Exploration-Exploitation Algorithm for the Online Joint Assortment-Inventory Optimization Problem}
\end{algorithm}

\subsection{Maximum Likelihood Estimation} \label{subsec:mle}

\mnote{blue}{AEQ1\\R1Q1\\R1Q2}{0mm} 
\edit{Our proposed algorithm is based on the counting estimators (CE) and the confidence bounds in \Cref{subsec-estimator,subsec-naive-extension} respectively. In fact, we can replace these by leveraging maximum likelihood estimation (MLE).
MLE has been applied to estimate MNL parameters in online assortment optimization without inventory capacity constraints \citep[e.g.,][]{Oh_Iyengar_2021}, and relatedly, logistic regression coefficients in generalized linear bandits \citep{filippi2010parametric}. We can also adapt MLE to our setting. Here we outline the high-level idea and defer the details to \Cref{sec-MLE}. Specifically, we reparameterize our MNL choice model by $\btht^* = (\theta_1^*, \cdots, \theta_N^*)$ where $\theta_i^* = \log v_i$ for $i \in [N]$ and $\theta_0^* = 0$,
and rewrite the choice probability for $i \in \{0\}\cup [N]$ given an assortment set $\s$ as $p_i(\s \mid \btht^*)$. At the end of each cycle $t$, we can estimate $\btht^*$ from the historical data by maximizing the corresponding likelihood function. In \Cref{sec-MLE} \Cref{lemma:MLE_FOC}, we show that the resulting MLE estimator  $\hat\btht_t$ solves the following equation:
\begin{equation}
    \sum_{s=1}^t\sum_{m=1}^{M_{s}}\sum_{i\in \s_{s,m}}
    \left[ \1{d_{s,m}=i} - p_{i}(\s_{s,m}\mid \btht)\right] \eb_i = \boldsymbol{0}, \label{eq:MLE eq}
\end{equation}
where $\s_{s, m}$ is the assortment set displayed to the $m$-th  customer in the $s$-th cycle, $d_{s,m} \in \{0\}\cup [N]$ denotes this customer's choice, and $\eb_i \in \mathbb{R}^N$ is a vector with its $i$-th entry being $1$ and all other entries being $0$.
}

\edit{
	In \Cref{sec-MLE} \Cref{lemma:theta convergence}, we further bound the estimation error of the MLE estimator $\hat\btht_t$. For each product $i$ and cycle $t$, let $n_{i,t} = \sum_{s=1}^{t-1}\sum_{m=1}^{M_{s}}\1{i\in \s_{s,m}}$ be the total number of customers for whom product $i$ is available before cycle $t$. Then for any $t$ such that $\min_{i \in [N]}n_{i, t} \ge \alpha_t^2$, we have 
	\begin{align*}
    \prob{\sum_{i\in [N]}(\hat\theta_{i,t}-\theta_{i}^*)^2 \cdot n_{i,t} \ge \alpha_t^2} \le \frac{1}{t} \text{ for } \alpha_t = \frac{1}{2\kappa}\sqrt{\frac{N}{2}\log\left(1 + \frac{2K\sum_{s=1}^{t-1} M_s}{N}\right) + \log {t}} \ , 
\end{align*}
where $\kappa$ is a positive constant such that $\kappa \le \inf_{\norm{\btht-\btht^*}{}\le 1,i\in[N],\s}  p_{i}(\s \mid  \btht)p_{0}(\s \mid \btht)$. This error bound motivates the confidence interval $\hat\theta_{i, t} \pm \alpha_t/\sqrt{n_{i, t}}$ for each $\theta_i^*$, provided that all products are sufficiently explored so that $\min_{i \in [N]}n_{i, t} \ge \alpha_t^2$. When this condition fails because some products are under-explored, we can force to explore these products until the condition is restored to make valid confidence intervals. 
In \Cref{sec-MLE}, we further convert the confidence intervals for $\btht$ into confidence intervals for $\bv$ via  proper transformations. We then  set the adjusted unit profits $\hat\bre_t$ accordingly, and carry out the rest of online decision-making following  Algorithm~\ref{alg-main}. In \Cref{sec-MLE}, we provide a detailed description of the MLE estimators, the corresponding online decision-making algorithm, and the regret analysis. 
}

\edit{While MLE can be adapted to our setting, it differs from our proposed method in several important aspects. Firstly, MLE requires repeatedly solving nonlinear equations given in \Cref{eq:MLE eq}. This is much more computationally expensive than our proposed estimators based on counting and averaging, as we numerically illustrate in  \Cref{sec-MLE} \Cref{table:estimator running time}. To speed up the computation, \cite{Oh_Iyengar_2021} proposes to perform online gradient descent updates rather than fully solving the MLE equation repeatedly, but we find that this would result in significant accuracy loss and it is outperformed by our proposed estimator in  numerical experiments. Secondly, MLE requires tracking the entire trajectory of customers' choices and assortments to form the likelihood function (see also \Cref{eq:MLE eq}). In contrast, our proposed method only requires maintaining one running counter for each product, separately. Thirdly, the MLE confidence bounds rely on the adequate exploration condition $\min_{i\in[N]}n_{i, t} \ge \alpha_t^2$, meaning that the validity of the confidence bound for one product needs all other products to be also sufficiently explored. In contrast, our estimators and confidence bounds for different products are completely decoupled. We find that this decoupling property of our method enables us to show that the regret incurred when some product $i$ violates the adequate exploration condition $Q_{i, t} < 1/48$ is negligible, but for MLE it is challenging to control the regret when the condition $\min_{i\in[N]}n_{i, t} \ge \alpha_t^2$ is violated so we incorporate additional forced exploration to meet this condition. Fourthly, MLE confidence bounds need specify a constant $\kappa$ that lower bounds the minimum eigenvalue of the log-likelihood Hessian matrix. This constant $\kappa$ could be challenging to specify: it needs to be small enough but if it is too small then the confidence bounds would be excessively wide. \Cref{sec-MLE} \Cref{lemma:theta convergence} provides a theoretically valid choice $\kappa = \frac{v_{\min}}{e\left(1 + Kv_{\max}\right)^2}$, provided that $v_{\min}$ is known. In contrast, our proposed confidence bounds are much easier to calculate, without needing knowledge of $v_{\min}$. Our numerical experiments in \Cref{fig:mle_widthcompare} show that our proposed confidence bounds are much tighter and converge much faster than the MLE confidence bounds based on the theoretical choice of $\kappa$. }

\section{Regret Analysis}\label{sec-results}
In this section, we analyze the total regret of our proposed exploration-exploitation algorithm and establish its rate-optimality. This section is organized as follows. 
In \Cref{subsec-upperbound}, we present a non-asymptotic upper bound on the regret of Algorithm~\ref{alg-main}. In \Cref{subsec-lowerbound}, we derive a matching regret lower bound and discuss the optimality of Algorithm~\ref{alg-main}.

\subsection{Regret Upper Bound} \label{subsec-upperbound}
In this subsection, we derive an upper bound on the total regret of Algorithm~\ref{alg-main} and outline its proof. Let $\pi$ denote the policy given by Algorithm \ref{alg-main}, and \edit{let $\Reg(\pi ; T)$ denote the worst-case total regret of implementing policy $\pi$ across all $\bv\in [v_{\min}, v_{\max}]^{N}$ and $\bre \in [0,1]^{N}$. Our main result is the following upper bound on the worst-case total regret $\Reg(\pi ; T)$ of our proposed algorithm.} \mnote{blue}{AEQ2\\R1Q3\\R2Q1}{10mm}

\begin{restatable}{theorem}{main}\label{thm-main}
    For any instance of the online joint assortment-inventory optimization problem with $N$ products, attraction parameters $\bv=(v_i)_{i \in [N]}$, unit profits $\bre = (r_i)_{i \in [N]}$, maximum assortment cardinality $K$, and expected number of customers $M$ per cycle, the worst-case total regret of the policy $\pi$ generated by Algorithm~\ref{alg-main} over $T$ inventory cycles satisfies
	\begin{equation*}
		\Reg(\pi ; T) =O  \left(\sqrt{MNT\log{(NT)}}  \right) \ .
	\end{equation*}
\end{restatable}

According to \Cref{thm-main}, the order of the dependence of regret on $M$, $N$ and $T$ is $\sqrt{MNT\log(NT)}$. \Cref{thm-main} applies universally to any underlying arrival distribution $M_t$, provided that the expected value $\Eb{M_t}$ equals $M$. Next, we outline the proof of \Cref{thm-main}.  

\subsubsection*{Proof outline.} 
\edit{To derive the regret bound in \Cref{thm-main}, for any $\bv \in [v_{\min}, v_{\max}]^{N}$ and $\bre \in [0,1]^{N}$, 
we decompose the total regret $\Reg(\pi ;\bv ,\bre, T)$ as defined in \Cref{equ-def-regret} into two parts according to whether certain events happen, and then upper bound each of the two parts separately. We show that these upper bounds are independent on the instance of $\bv \in [v_{\min}, v_{\max}]^{N}$ and $\bre \in [0,1]^{N}$, and thus we obtain \textit{instance independent} worst-case regret upper bound.} We first define the events used in the regret decomposition.

\begin{definition}\label{def-event}
	For all $t \in [T]$ and $i\in[N]$, \edit{define the event $\Acal_{i,t}^1\coloneqq \lbr{Q_{i, t} < 1/48}$ (recall that $ Q_{i, t} \coloneqq  \log (\sqrt{Nt\sum_{s=1}^{t-1}M_s} + 1)/k_{i, t}$ and $Q_{i, t} < 1/48$ is the adequate exploration condition for product $i$), 
    and the event $\event_{i,t}^2$ as 
	\begin{equation*}
			\event_{i,t}^2 \coloneqq  \left.\bigg\{ \recv_{i, t}^{\LCB}\le \recv_{i} \le \recv_{i, t}^{\UCB}\text{, and }\  \recv_{i, t}^{\UCB} -  \recv_{i, t}^{\LCB} \le 12\sqrt{3} \max\{\sqrt{\recv_{i}}, \recv_{i}\}\sqrt{Q_{i, t} } + 192 Q_{i, t}   \right.\bigg\}.
	\end{equation*}
Define the event $\Acal_t \coloneqq \bigcap_{i\in[N]} \lbr{\Acal_{i,t}^{1c}\bigcup \Acal_{i,t}^2}$,
where $\event_{i,t}^{1c}$ is the complementary event of $\event_{i,t}^{1}$.} 
\end{definition}

In \Cref{def-event}, the event $\event_t$ corresponds to the event that \edit{either a product $i$ is not adequately explored, or when it is, its confidence interval is well-behaved. Specifically, a well-behaved confidence interval $[\recv_{i, t}^{\LCB}, \recv_{i, t}^{\UCB}]$ means that the true value of $\recv_i$ lies in the  interval and the width of that interval is bounded above by a vanishing term.}
Then, by the reciprocal relationship between $\recv_{i}$ and $v_{i}$, we also have that $v_i \in [v_{i, t}^{\LCB}, v_{i, t}^{\UCB}]$, and that $v_{i, t}^{\UCB} - v_{i, t}^{\LCB}$ is bounded above by a vanishing term. 
Note that the former is the sufficient condition in \Cref{lemma-policyUCB} for our proposed objective $R(\U; \bvUCB{t}, \hat \bre_{t})$ to provide valid optimistic profit estimates, and the latter is useful in showing that the optimistic estimates are increasingly accurate (see \Cref{lemma-convergence-rate} below). 
Thus the events $\event_t$ for $t \in [T]$ define a sequence of desirable events for our proposed algorithm. 
We now show that these desirable events hold with high probability. %

\begin{restatable}{lemma}{eventprob}	\label{lemma-event-prob}
    For any cycle $t \in [T]$, event $\event_t$ happens with a probability of at least $1 - 9/t$.
\end{restatable}

Then, for every $\bv \in [v_{\min}, v_{\max}]^{N}$ and $\bre \in [0,1]^{N}$, we decompose the total regret  as follows: 
    \begin{align}
		\Reg(\pi ;\bv ,\bre, T) = 
		&\ \Epi{\sum_{t = 1}^T \left( R(\U^*; \bv, \bre) - R(\U_t; \bv, \bre) \right)}  \label{equ-reg-part} \\
		= &\  \Epi {\sum_{t \in [T]}  
  \left(R(\U^*; \bv, \bre) - R(\U_t; \bv, \bre)\right) \mathbbm{1}(\event_{t}^c)}
  + \Epi {\sum_{t \in [T]}   \left(R(\U^*; \bv, \bre) - R(\U_t; \bv, \bre)\right) \mathbbm{1}(\event_{t})} \nonumber , 
\end{align}
where the second equality follows from decomposing the regret according to whether the corresponding desirable event in \Cref{def-event} happens. 
Next, we explain the two parts of \Cref{equ-reg-part} in more detail and sketch how to upper bound each of them.

\subsubsection*{(i) Upper bounding $ \Epi {\sum_{t \in [T]}    \left(R(\U^*; \bv, \bre) - R(\U_t; \bv, \bre)\right) \mathbbm{1}(\event_{t}^c)}$.} 
This term captures the expected total regret incurred in cycles $t \in [T]$, in which the desirable event $\event_{t}$ does not hold. 
First, the regret incurred in a single inventory cycle is upper bounded by $M$, the expected number of customer arrivals per cycle. That is, for any feasible decision $\U \in \setU$,  we have 
\begin{align}\label{eq: trivial-bound}
R(\U^*; \bv, \bre) - R(\U; \bv, \bre) \le R(\U^*; \bv, \bre) \le M,
\end{align}
where the first inequality trivially holds and the second inequality holds because one customer purchases at most one product and generates at most one unit of profit (recall that $r_{\max} = 1$) while on average $M$ customers arrive in each inventory cycle.
Thus we have
\begin{equation*}
    \Epi {\sum_{t \in [T]}    \left(R(\U^*; \bv, \bre) - R(\U_t; \bv, \bre)\right) \mathbbm{1}(\event_{t}^c)} \le M\cdot \Epi {\sum_{t \in [T]}    \mathbbm{1}(\event_{t}^c)}.
\end{equation*}
This result, in conjunction with \Cref{lemma-event-prob} that bounds the probability of $\event_t^c$ for $t \in [T]$ by $9/t$, leads to the conclusion that the second part of \Cref{equ-reg-part} is also at most logarithmic in $T$.
 
\subsubsection*{(ii) Upper bounding $\Epi {\sum_{t \in [T]}    \left(R(\U^*; \bv, \bre) - R(\U_t; \bv, \bre)\right) \mathbbm{1}(\event_{t})}$.}
	
This term calculates the expected total regret of all cycles $t \in [T]$ where the desirable event $\event_{t}$ holds. This is also the dominant term in the total expected regret. According to \Cref{lemma-event-prob}, the event $\event_{t}$ happens with high probability, thus in order to obtain a tight regret upper bound, we need to carefully bound the single-cycle regret $R(\U^*; \bv, \bre) - R(\U_t; \bv, \bre)$ given that $\event_{t}$ is true. 
We note that for all cycles $t \in [T]$ where the desirable event $\event_{t}$ holds, we have 
\begin{equation}\label{eq: ucb-chain}
    R(\U^*; \bv, \bre) - R(\U_t; \bv, \bre) \le  R(\U^*; \bvUCB{t}, \hat \bre_{t}) - R(\U_t; \bv, \bre) \le  R(\U_t; \bvUCB{t}, \hat \bre_{t}) - R(\U_t; \bv, \bre),
\end{equation}
where the first inequality holds because $R(\U^*; \bv, \bre) \le R(\U^*; \bvUCB{t}, \hat \bre_{t})$ according to \Cref{lemma-policyUCB}, and the second inequality holds because $\U_t$ is chosen to maximize $R(\U; \bvUCB{t}, \hat \bre_{t})$ over $\U \in \setU$. 
As a result, we can bound the single-cycle regret by the estimation error of the optimistic expected profit at the chosen decision $\U_{t}$. It then remains to bound this estimation error. Let $V\coloneqq v_{\max} / v_{\min}$. We have the following result.
\begin{restatable}{lemma}{convergencerate}\label{lemma-convergence-rate}
	For every $t \in [T]$, if the event $\event_{t}$ is true, then we have
 \begin{align*}
		R(\U_{t}; \bvUCB{t}, \hat\bre_{t}) - R(\U_{t}; \bv, \bre)\le &\  \sum_{i\in S(\U_t)} \left(2\delta_{i,t} + \delta_{i,t}^2\right)\Eb{X_{i,t}^{\U_t, \bv}}\1{Q_{i, t} < 1/48} \\
  &\ + \sum_{i\in S(\U_t)}V\Eb{X_{i,t}^{\U_t,\bv}}\1{Q_{i, t} \ge 1/48} .
	\end{align*}
\end{restatable}
\Cref{lemma-convergence-rate} asserts that the single-cycle regret $R(\U_t; \bvUCB{t}, \hat\bre_{t}) - R(\U_t; \bv, \bre)$ can be bounded by an increasingly tight upper bound when $\event_{t}$ is true. We can then sum up these upper bounds across all $t \in [T]$ to get an upper bound on the second part of \Cref{equ-reg-part}. 
To establish \Cref{lemma-convergence-rate}, we leverage some structural properties of the MNL choice model to 
show that the expected profit function $R(\U; \bv, \bre)$ is Lipschitz in the parameter vectors $\bv$ and $\bre$. 
This means that the profit estimation error $R(\U_{t}; \bvUCB{t}, \hat\bre_{t}) - R(\U_{t}; \bv, \bre)$ can be upper bounded by the size of the parameter estimation errors $\bvUCB{t} - \bv$ and $\hat\bre_{t} - \bre$. 
Then \Cref{lemma-convergence-rate} results from upper bounds on the parameter estimation errors implied by the desirable events $\event_{t}$'s.

Finally, by putting together the upper bounds described above for the two parts of \eqref{equ-reg-part}, we can reach the regret upper bound in \Cref{thm-main}. The detailed proof is provided in Appendix.

\subsection{Regret Lower Bound}\label{subsec-lowerbound}
In this section, we show that any admissible policy for the online joint assortment-inventory optimization problem must incur a worst-case regret of $\Omega(\sqrt{MNT/ K})$, which implies that our proposed Algorithm~\ref{alg-main} achieves nearly optimal regret rate.

\begin{restatable}{theorem}{lowerbound}\label{thm-lower-bound}  
For any admissible policy $\pi_{\MNLI}$, there exists an instance of online joint assortment-inventory
optimization problem with $N$ products, $T$ inventory cycles satisfying $T \ge N$, maximum assortment size $K$, and expected number of customers $M$ per cycle, such that the total regret of the policy $\pi_{\MNLI}$ on this instance satisfies 
	\begin{align*}
            \Reg(\pi_{\MNLI} ;  T) = 
            \begin{cases}
            \Omega(\sqrt{MNT}) & K \le N/4  \\
            \Omega(\sqrt{MNT/K}) & \text{otherwise}
            \end{cases}
            .
	\end{align*}
\end{restatable}

\Cref{thm-lower-bound} shows that our proposed Algorithm~\ref{alg-main} achieves a nearly optimal regret rate.
According to \Cref{thm-lower-bound}, when the maximum assortment size $K$ is small enough compared to the total number of products $N$ so that $K \le N/4$, no policy can achieve  regret rate better than $\sqrt{MNT}$ uniformly over all  instances. 
This regret rate is nearly achieved by our proposed Algorithm~\ref{alg-main}, only up to an additional $({1 + K\log K}/{M})$ factor and a logarithmic factor ${\sqrt{\log(NT)}}$  (see \Cref{thm-main}). In particular, the $({1 + K\log K}/{M})$ factor may be fairly close to $1$ if the expected number of single-cycle customer arrivals $M$ is much larger than the maximum assortment size $K$, as is common in  real-world applications. 
Thus our proposed algorithm achieves a nearly optimal regret rate when  $K \le N/4$ and $M \gg K$. 
In contrast, when  $K > N/4$, the regret upper bound in  \Cref{thm-main} involves an extra $\sqrt{K}$ factor compared to the lower bound in \Cref{thm-lower-bound}. 
We note that a similar phenomenon also appears in \cite{agrawal2019mnl}, as the regret upper bound for their online assortment optimization algorithm also involves an extra $\sqrt{K}$ factor relative to their regret lower bound. 
How to eliminate this extra factor, either by more refined regret analysis or better algorithms, is still an open question  even for the simpler online assortment optimization problem. 

Furthermore, we can compare our regret bounds with the regret bounds in \cite{agrawal2019mnl,chen2018note} for the MNL-bandit problem of online assortment optimization. 
The  MNL-bandit problem can be viewed as a special case of our problem with deterministic $M_1, \dots, M_T$ all  equal to $M = 1$. 
When specialized to this setting, our regret upper bound in \Cref{thm-main} has nearly the same rate with respect to $N, T$ as the dominating term in the regret upper bound for the MNL bandit algorithm in \cite{agrawal2019mnl}, despite worse dependence on $K$. 
Our lower bound in \Cref{thm-lower-bound} specialized to $M = 1$ recovers the lower bounds in \cite{agrawal2019mnl,chen2018note}, since the proof of \Cref{thm-lower-bound} directly builds on their lower bounds, as we will explain shortly. 
However, our novel algorithm and regret analysis can handle more general $M_1, \dots, M_T$ and various challenges caused by stochastic stockout events under limited inventory, thereby advancing the existing literature. 

Finally, we briefly sketch the proof for the lower bound in \Cref{thm-lower-bound}. The proof is based on constructing a reduction of the MNL-bandit problem for assortment optimization to our problem. 
In particular, it suffices to focus on problem instances with deterministic $M_1 = \dots = M_T = M$. %
For $M = 1$, our problem is equivalent to the MNL-bandit problem, so \Cref{thm-lower-bound} directly follows from \cite{chen2018note} when $K \le N/4$ and from \cite{agrawal2019mnl} when $K > N/4$. 
For $M > 1$, we show that for any MNL-bandit problem instance $I_{\text{MNL}}$ with $MT$ customer arrivals, we can construct an online assortment-inventory optimization problem instance $I_{\text{MNLI}}$ with $T$ inventory cycles and $M$ customer arrivals per cycle, such that any admissible policy $\pi_{\text{MNLI}}$ on instance $I_{\text{MNLI}}$ induces an admissible policy $\pi_{\text{MNL}}$ on instance $I_{\text{MNL}}$ with identical total regret. 
Then  we can again obtain \Cref{thm-lower-bound}  from 
the corresponding MNL-bandit lower bounds in \cite{chen2018note,agrawal2019mnl}.
This proof provides a convenient template to translate lower bounds for the MNL-bandit problem to lower bounds for the online joint assortment-inventory optimization problem. 
Therefore, whenever tighter lower bounds for the MNL-bandit problem are discovered in the future, they can be readily adapted to our problem in the same way. 
See Appendix \Cref{sec: reduction} for more details.

\section{Incorporating Approximate Static Optimization Oracles}\label{sec-oracles}
In previous sections, we assume the existence of an \emph{exact} oracle to solve static joint assortment-inventory optimization problems.
This simplified assumption allows us to ignore errors in solving static optimization problems, so that we can focus on the central challenge of exploration-exploitation trade-off in online decision making. In this section, we relax this assumption by incorporating \emph{approximate} static optimization oracles into our algorithm.

Specifically, recall that at the beginning of each cycle $t$,  our algorithm needs to solve 
\begin{equation*}
	\U_t \in \argmaxx_{\U \in \setU} \R(\U, \bvUCB{t}, \hat \bre_{t}).
\end{equation*}
Previously, we assumed that these optimization problems can be solved \emph{exactly} by an oracle. However, to the best of our knowledge, at present there is no such efficient exact oracle. Indeed, even efficiently evaluating the expected profit of given assortment and inventory decisions is notoriously difficult, let alone solving for the profit-maximizing decision \citep{aouad2022stability,aouad2018greedy}. 
Fortunately, although exact oracles do not exist, there exist a variety of approximation algorithms or heuristics that can serve as \emph{approximate} oracles. We refer to \cite{liang2021assortment,aouad2018greedy,aouad2022stability,sun2024unifiedalgorithmicframeworkdynamic} for several examples. \edit{In \Cref{sec:oracle-survey}, we provide brief descriptions of these approximate oracles and their respective computational complexities and theoretical guarantees.}

Based on the examples above, we formalize $\epsilon$-$\delta$ oracles to the static joint assortment-inventory optimization problem. The aforementioned approximation algorithms or heuristics are examples of $\epsilon$-$\delta$ oracles with appropriate error parameters $\epsilon$ and $\delta$. 

\begin{definition}\label{define-sigma-delta-oracle}
	An oracle to the static joint assortment-inventory optimization problem is called an $\epsilon$-$\delta$ oracle if, for any instance with any known value of parameters $\bv$ and $\bre$, the oracle can obtain a decision $\hat{\U} \in \setU$ such that $R(\hat \U; \bv, \bre) \ge (1 - \epsilon) \max_{\U \in \setU} R(\U; \bv, \bre)$ with a probability of at least $1 - \delta$.
\end{definition}
 
We can easily incorporate an $\epsilon$-$\delta$ oracle into our Algorithm~\ref{alg-main}. Specifically, we apply the  oracle to obtain an approximate solution $\hat \U_{t}$ for every cycle $t$, such that with a probability of at least $1- \delta$, 
\begin{equation*}
	 \R(\hat \U_t, \bvUCB{t}, \hat \bre_{t}) \ge (1-\epsilon)\max_{\U\in\setU}\R(\U, \bvUCB{t}, \hat \bre_{t}) = (1-\epsilon)\R(\U_{t}, \bvUCB{t}, \hat \bre_{t}).
\end{equation*}
Then we can extend \Cref{eq: ucb-chain}  and upper bound the single-cycle regret of the approximate solution $\hat \U_{t}$ in cycle $t$, in terms of both the profit estimation error and the optimization error.
\begin{restatable}{lemma}{ucboforacle}\label{lemma-ucb-oracle}
    For each cycle $t \notin \setexp$ such that the event $\event_{t}$ is true, the approximate solution $\hat\U_{t}$ obtained by an $\epsilon$-$\delta$ oracle satisfies the following inequality with a probability of at least $1-\delta$: 
        \begin{align}
        \label{equ:epsilon-delta-regret-cycle}
    R(\U^*; \bv, \bre) - R(\hat\U_t; \bv, \bre) \le   R(\hat\U_{t}; \bvUCB{t}, \hat \bre_{t}) - R(\hat\U_t; \bv, \bre) + \epsilon R(\U^*; \bv, \bre)\ .
    \end{align}
\end{restatable}

In the following theorem, we extend the analyses in \Cref{subsec-upperbound} to upper bound the regret of Algorithm~\ref{alg-main} based on an $\epsilon$-$\delta$ oracle.

\begin{restatable}{theorem}{regretworacle}\label{thm-regret-oracle}
Consider the policy $\pi_{\epsilon, \delta}$ generated by Algorithm~\ref{alg-main} based on an $\epsilon$-$\delta$ oracle. For any instance stated in \Cref{thm-main}, the total regret of policy $\pi_{\epsilon, \delta}$ over $T$ inventory cycles satisfies
	\begin{equation*}
		\Reg(\pi_{\epsilon, \delta}; \bv, \bre, T) = O\left( (\epsilon + \delta) R(\U^*;\bv, \bre)T +  \sqrt{MNT\log(\sqrt{NM}T)}\right)\ .
	\end{equation*}
\end{restatable}

Compared to \Cref{thm-main}, the regret bound in \Cref{thm-regret-oracle} includes an additional term accounting for the approximation errors of the $\epsilon$-$\delta$ oracle. \edit{This additional error term depends on the values of $\bv$ and $\bre$.} Note that the regret automatically adapts to the errors of the approximate static optimization oracle, even if we do not know the magnitude of the errors when implementing our algorithm. 
\edit{Importantly, \Cref{thm-regret-oracle} asserts that the regret caused by the exploration-exploitation trade-off is still sublinear in $T$ regardless of the instance.}

\section{Numerical Experiments}
\label{sec-simulation} \mnote{blue}{New sections\\AEQ4}{0mm}
\edit{
In this section, we evaluate our proposed algorithm via
three sets of numerical experiments: one in small scale for which we can exactly solve the static optimization problems and evaluate the regret relative to the clairvoyant decision, one in large scale with parameters calibrated from a real-world dataset, and one also in large scale with randomly generated parameters. For the latter two large-scale experiments, we use an approximate oracle to solve the static optimization problems and evaluate the cumulative profit instead of regret because the clairvoyant optimal solution is intractable.}

\edit{To highlight the effectiveness of our proposal to adaptively tune the unit profits $\bre$ and the importance of exploration, we compare the regret of our proposed exploration-exploitation algorithm with two benchmarks. %
The first benchmark is the heuristic approach described in \Cref{subsec-naive-extension}, which picks the inventory decision by maximizing $R(\U, \bvUCB{t}, \bre)$ over all feasible inventory levels $\U \in \setU$ for each cycle $t \in [T]$.  Since the heuristic approach only uses $\bvUCB{t}$ while fixing the values of unit profits $\bre$, we refer to it as the ``$\bv^\UCB$-only'' algorithm. The second benchmark is a purely greedy algorithm. Specifically, in every cycle $t$, the greedy algorithm exploits the latest attraction parameter estimate $\bar{\bv}_{t}$, and determines the inventory decision by maximizing $R(\U, \bar \bv_{t}, \bre)$ over $\U \in \setU$.} %

\subsection{Small Scale Experiments}
\edit{We first restrict our experiments to small scale problems, for which both profit evaluation and optimization can be implemented by exact brute-force enumeration. Accordingly, 
differences in performances can be fully attributed to the exploration-exploitation trade-off.}

\edit{Specifically, we test the two settings described in \Cref{table-numerical-settings}. In both settings, we simulate the number of customer arrivals in each cycle according to a Poisson distribution with parameter $M = 6$. We simulate each setting independently for $10$ times and report the average regret of each algorithm.}   

\begin{table}[h]
	\caption{Numerical experiment settings.}
    \small
	\label{table-numerical-settings}
	\begin{tabularx}{\textwidth}{p{0.12\textwidth} p{0.06\textwidth} p{0.06\textwidth} p{0.06\textwidth} X X}
		\toprule
		\bd{Index} & $N$ & $M$ & $\bar c$ & $\bv$& $\bre$ \\
		\midrule
		\small \bd{Setting 1}& 5 & 6 & 6 &  (0.9, 0.3, 0.3, 0.3, 0.2) &  (0.6, 1, 1, 1, 1) \\
		\small \bd{Setting 2}& 6 & 6 & 5 &  (0.9, 0.3, 0.3, 0.3, 0.2, 0.12) &  (0.6, 1, 1, 1, 1, 1) \\
		\bottomrule
	\end{tabularx}
\end{table}

\begin{figure}[htbp]
	\centering
	\subfigure[Setting 1.]
	{
		\begin{minipage}[b]{.48\linewidth}
			\centering
			\includegraphics[scale=0.4]{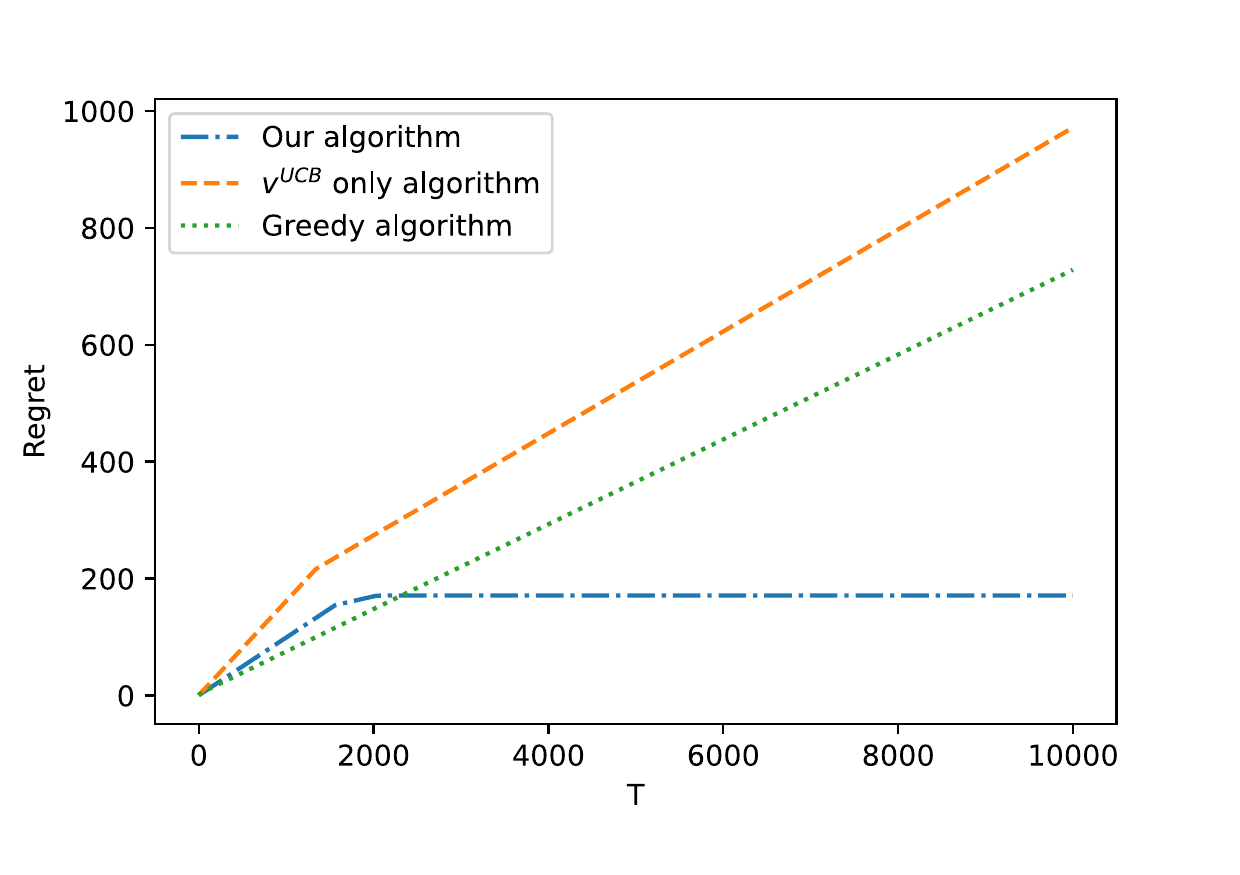}
		\end{minipage}
	}
	\subfigure[Setting 2.]
	{
		\begin{minipage}[b]{.48\linewidth}
			\centering
			\includegraphics[scale=0.4]{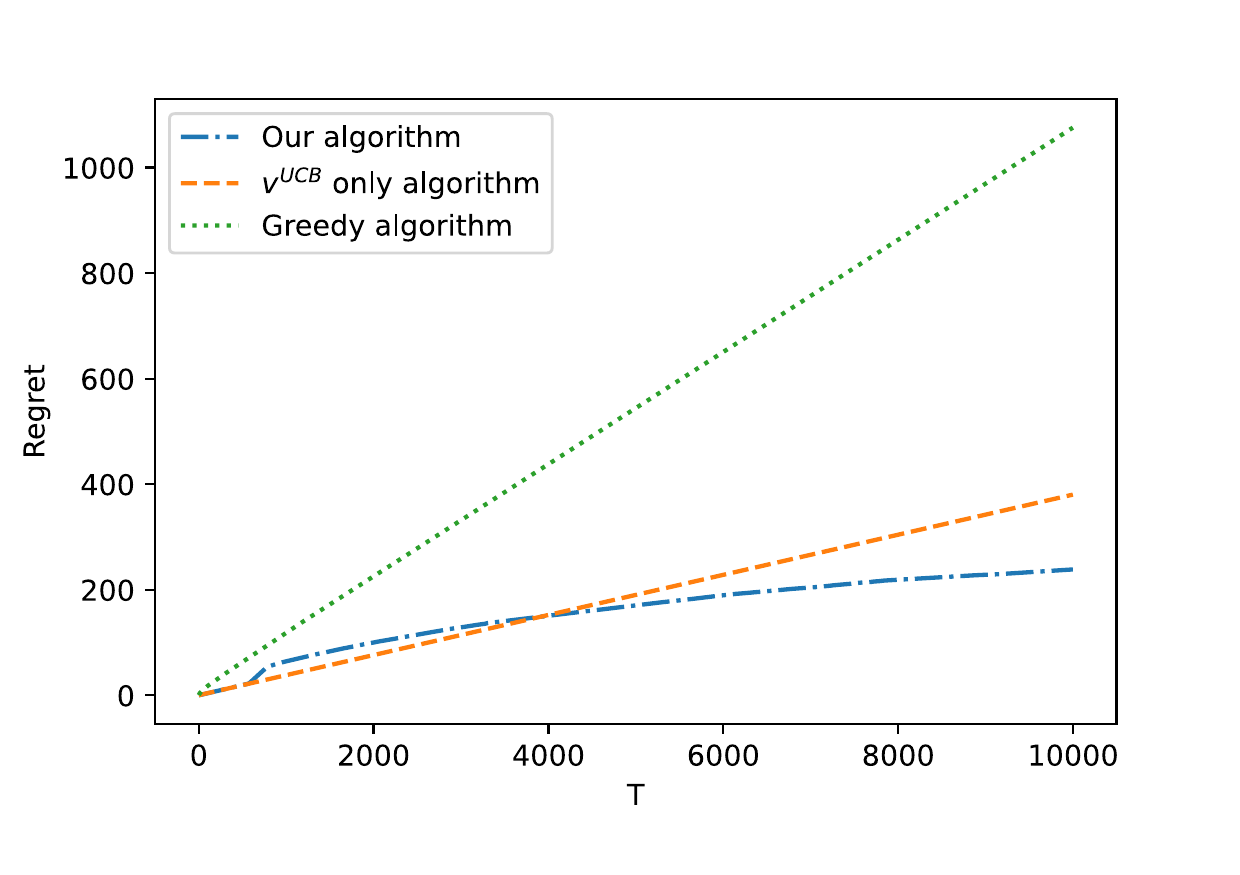}
		\end{minipage}
	}
	\caption{Average regrets of our algorithm and two benchmark algorithms over different time horizons $T$ under the two settings listed in \Cref{table-numerical-settings}. Results are based on $10$ independent replications of the experiment.}	\label{fig-samples}
\end{figure}

\edit{\Cref{fig-samples} shows the average regret of each algorithm up to $10,000$ inventory cycles in the two settings}. We observe that, while the regret of our proposed algorithm grows only at a sub-linear rate, the regrets of the two benchmarks may accumulate linearly with the inventory cycles. 
This is because the two benchmarks may fail to explore the decisions properly and thus get trapped in sub-optimal decisions within the time horizon of our experiments. In contrast, our proposed algorithm can sufficiently explore the decisions and gradually converge to the optimal decision.

We now inspect the two experiment settings more closely to better understand the behaviors of different algorithms.
It is noteworthy that in both experiment settings, the first product has the lowest profit $r_1 = 0.6$ but it is still assorted into the clairvoyant optimal inventory decision because of its high  attraction $v_1 = 0.9$. 
However, we find that this product is often left out by the ``$\bvUCB{}$-only'' algorithm. Indeed, the first product often does not appear profitable enough to the ``$\bvUCB{}$-only'' algorithm, since its unit profit is very low and its attraction parameter is not particularly advantageous when 
all products' attractions are optimistically estimated.
This explains why the ``$\bvUCB{}$-only'' algorithm tends to miss the optimal decision for a long time. 
Moreover, we find that in many simulation instances, the greedy algorithm fixates too early on a suboptimal decision due to random fluctuations of the parameter estimation errors.  
In contrast, our proposed algorithm  can sufficiently explore all  products and identify the true optimal decision.

\subsection{Large Scale Experiments with Real-world Data Calibrated Parameters}\label{sec: numerical-sushi}

\begin{figure}[htbp]
	\centering
	\subfigure[$N=20$]
	{
		\begin{minipage}[b]{.48\linewidth}
			\centering
			\includegraphics[scale=0.4]{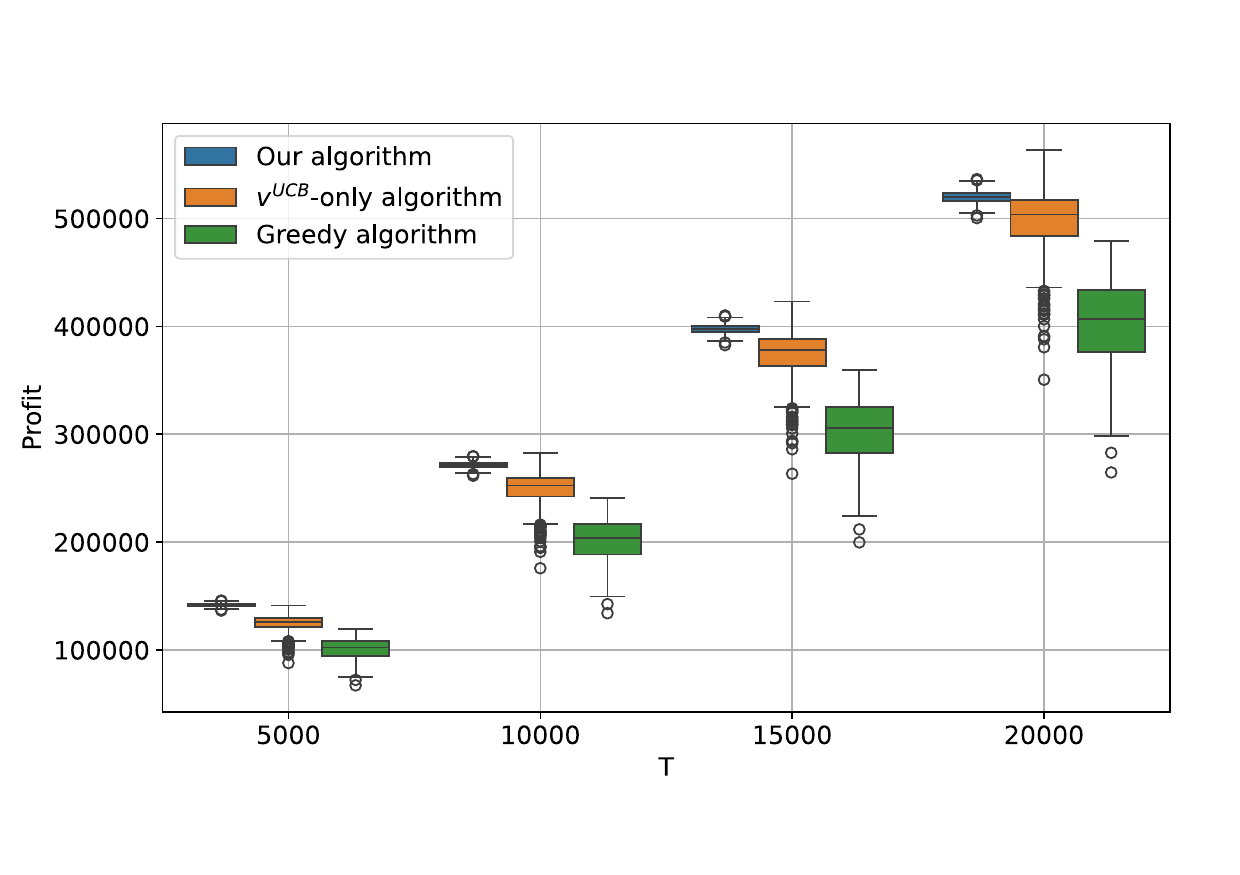}
		\end{minipage}
	}
	\subfigure[$N=30$]
	{
		\begin{minipage}[b]{.48\linewidth}
			\centering
			\includegraphics[scale=0.4]{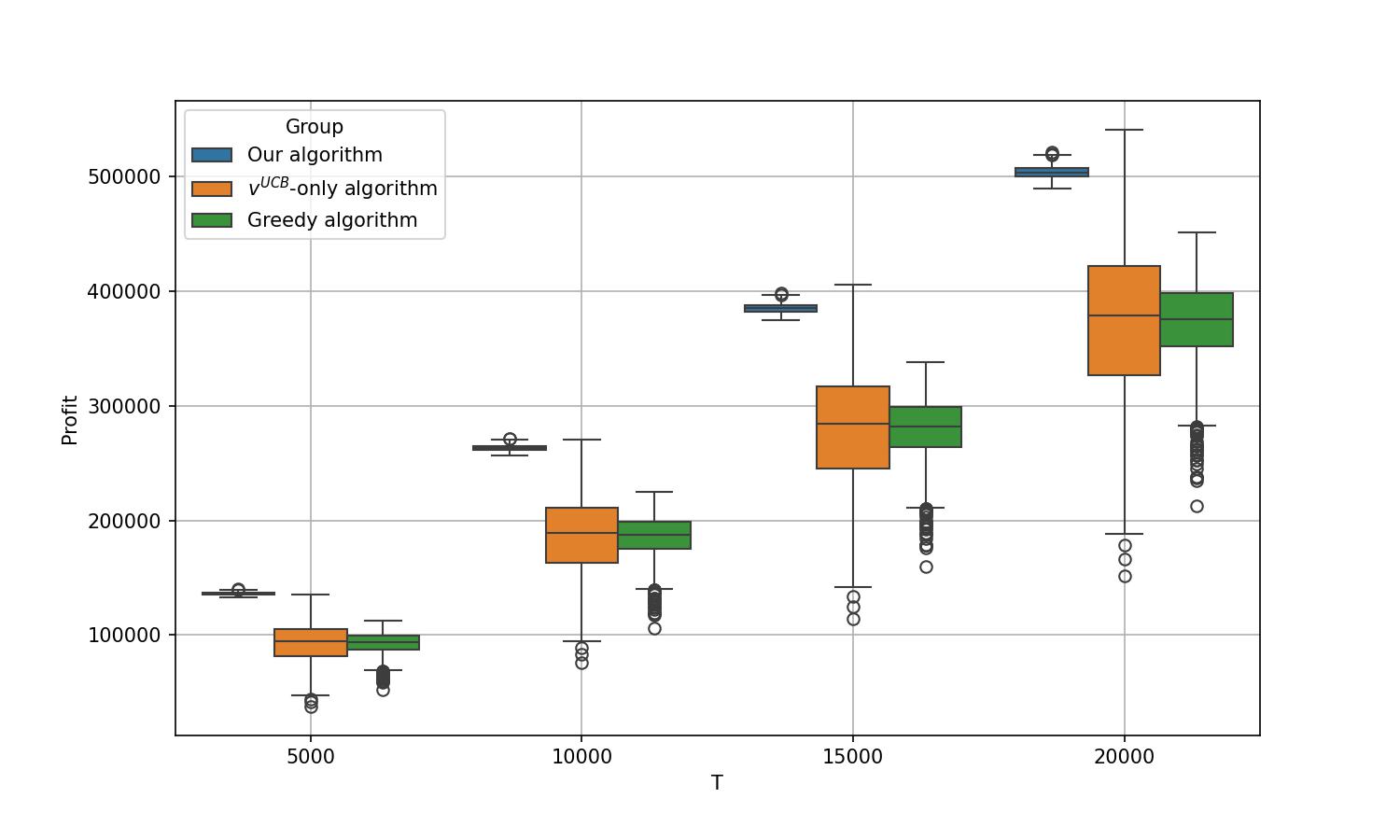}
		\end{minipage}
	}
        \subfigure[$N=40$]
	{
		\begin{minipage}[b]{.48\linewidth}
			\centering
			\includegraphics[scale=0.4]{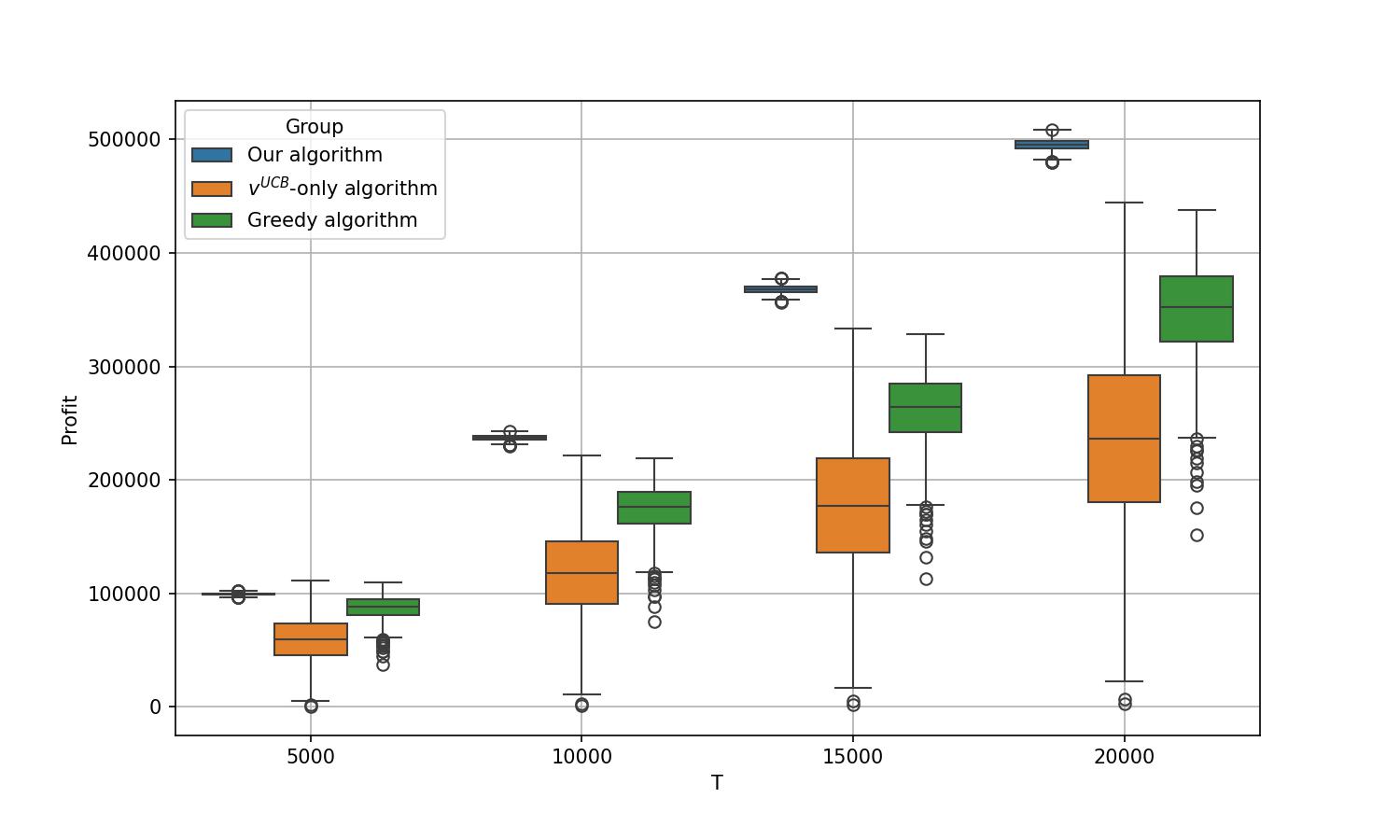}
		\end{minipage}
	}
        \subfigure[$N=50$]
	{
		\begin{minipage}[b]{.48\linewidth}
			\centering
			\includegraphics[scale=0.4]{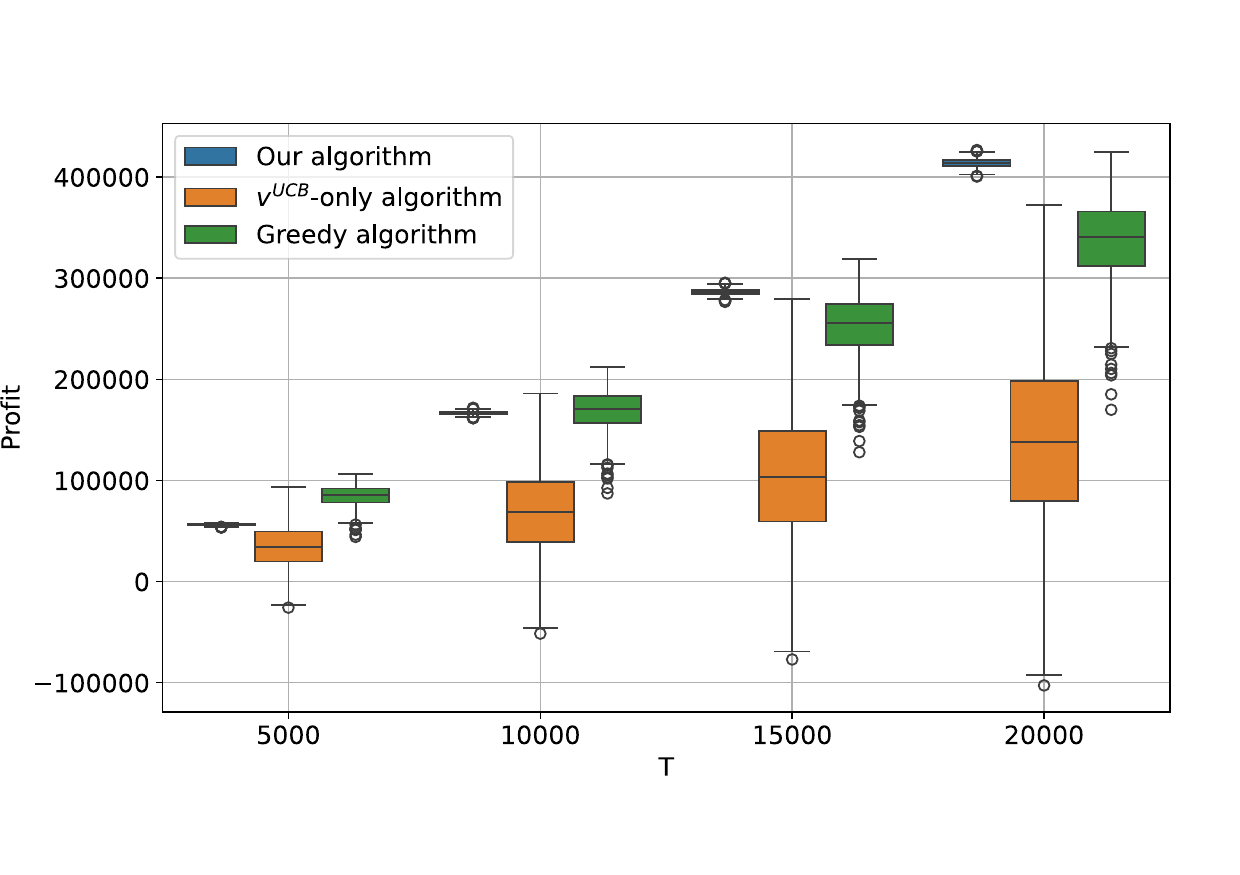}
		\end{minipage}
	}
	\caption{Boxplots of the cumulative profits at four time points, $T \in \{5000, 10000, 15000, 20000\}$. Results are based on $1000$ independent replications of each experiment. }	\label{fig:sushi calibrate}
\end{figure}

\mnote{blue}{New section\\AEQ4}{0mm}
\edit{In this subsection, we follow the widely-adopted approach in the online decision-making literature to conduct numerical  experiments with models calibrated from real-world data \cite[see, for example,][]{gah-yi2021,luo2024}. 
Specifically, we use the dataset in \cite{Toshihiro} that includes preference rankings from a survey of 5,000 individuals for 100 different types of sushi (\href{}{\url{https://preflib.simonrey.fr/static/data/sushi/}}). Using this dataset, we calibrate an MNL choice model  as follows.
We initialize the raw attraction parameter for each product based on the proportion of preference rankings where it is the top choice, and then sort the products in  descending order of  the raw attraction parameters. 
We focus on the top $N$ products and aggregate the remaining products into a single outside option, with their combined raw attraction parameters as the parameter $v_0$.  Finally, we normalize $v_0$ to $1$ and proportionally scale the attraction parameters of the top $N$ products. Obviously, a larger $N$ corresponds to greater  relative attraction of each product of interest  compared to the outside option.}

\edit{We run four experiments with $N = 20, 30, 40, 50$ respectively.
In each experiment, we fix the number of customer arrivals per cycle at  $M = 100$. 
Given the complexity of the single-cycle static optimization problems, which are too large to solve exactly, we employ approximate optimization oracles for our proposed algorithm and the two benchmarks. For simplicity, we do not impose inventory capacity or assortment cardinality constraint, and we use the computationally efficient linear programming approximate oracle in \cite{liang2021assortment}. 
Noting that the dataset lacks cost and price information, we set the ordering costs  for all sushi to $0.1$ and the salvage values to $0$ to  prevent excessive inventory ordering.  
Additionally,  we randomly assign prices to each sushi type from a uniform distribution ranging between $0.9$ and $1$. 
This setting with  unequal ordering costs and salvage values is addressed in \Cref{sec-relax-assumption}. Following the reformulations in that section, we apply our proposed algorithm and the benchmarks, assessing their cumulative profits   over a total time horizon of $T = 20,000$.
Each experiment is replicated $1,000$ times, with each replication based on a new set of randomly drawn sushi prices.}

\edit{
The boxplots in \Cref{fig:sushi calibrate} show the distributions of the cumulative profits for each algorithm across the $1000$ experiment replications at four time points $T \in \{5000, 10000, 15000, 20000\}$. It is evident that our proposed algorithm outperforms the two benchmarks  in terms of both median profit and the volatility of profits. The advantage of our algorithm is particularly pronounced when $T$ is large and the number of candidate products $N$ is large.}

\subsection{Large Scale Experiments with Randomized Attraction Parameters}\label{sec: numerical-randomized}

\begin{figure}[htbp]
	\centering
	\subfigure[Setting $1$]
	{
		\begin{minipage}[b]{.48\linewidth}
			\centering
			\includegraphics[scale=0.4]{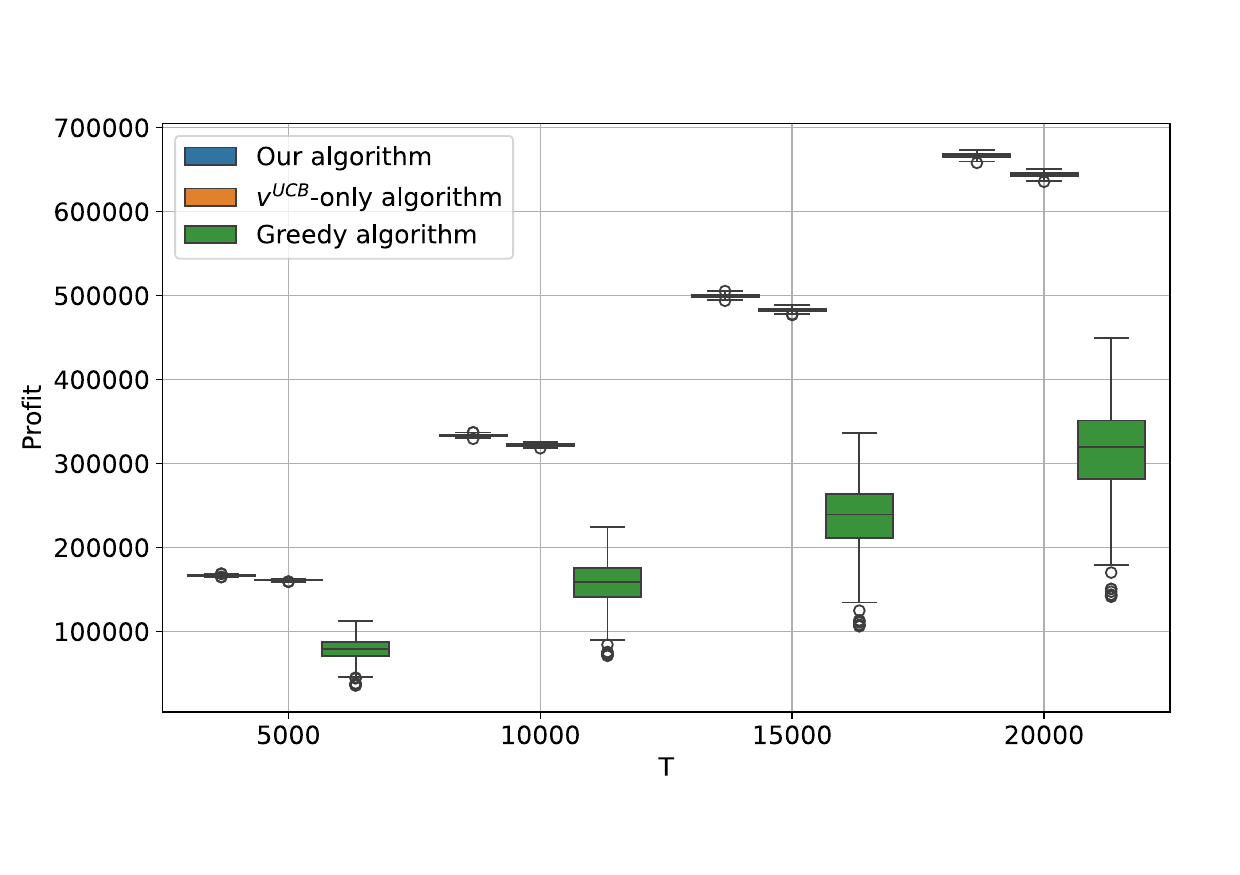}
		\end{minipage}
	}
	\subfigure[Setting $2$]
	{
		\begin{minipage}[b]{.48\linewidth}
			\centering
			\includegraphics[scale=0.4]{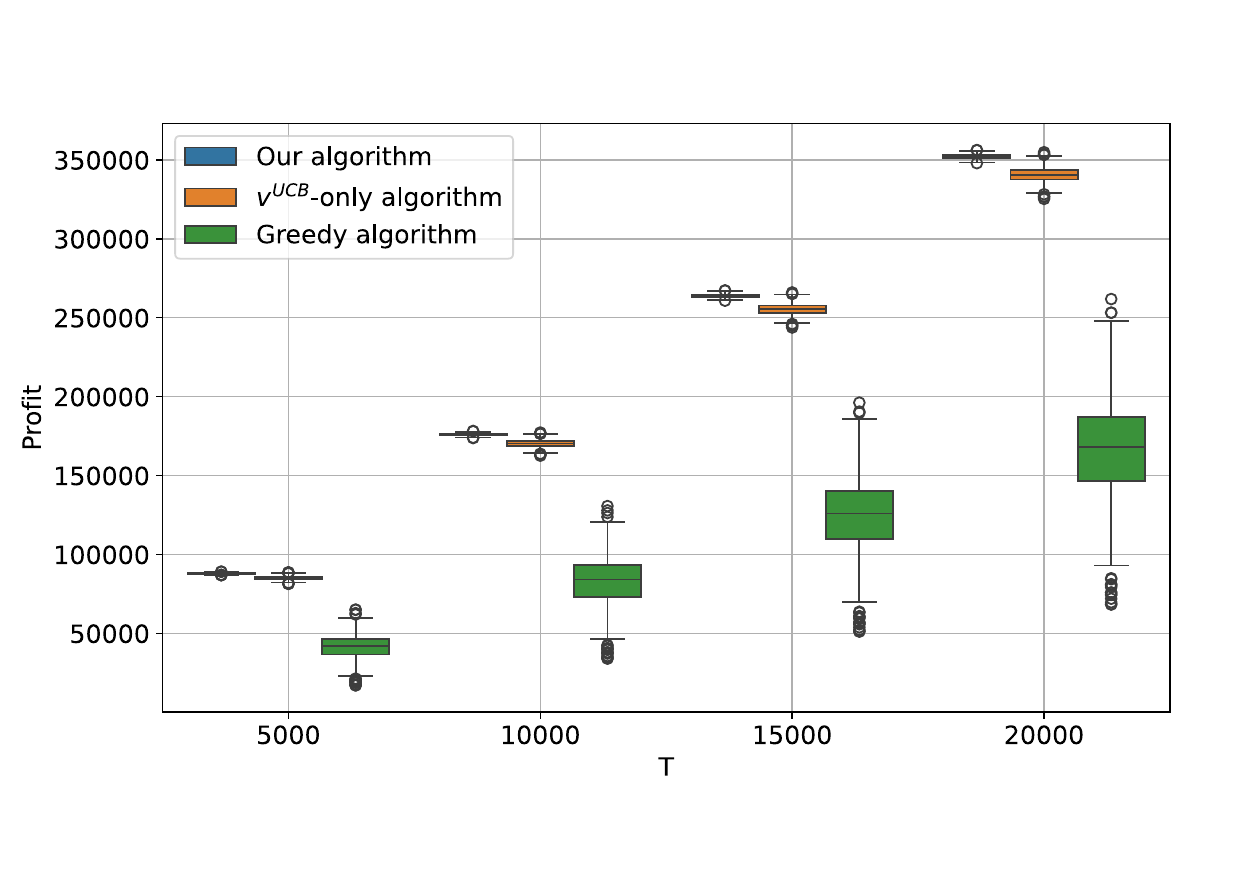}
		\end{minipage}
	}
        \subfigure[Setting $3$]
	{
		\begin{minipage}[b]{.48\linewidth}
			\centering
			\includegraphics[scale=0.4]{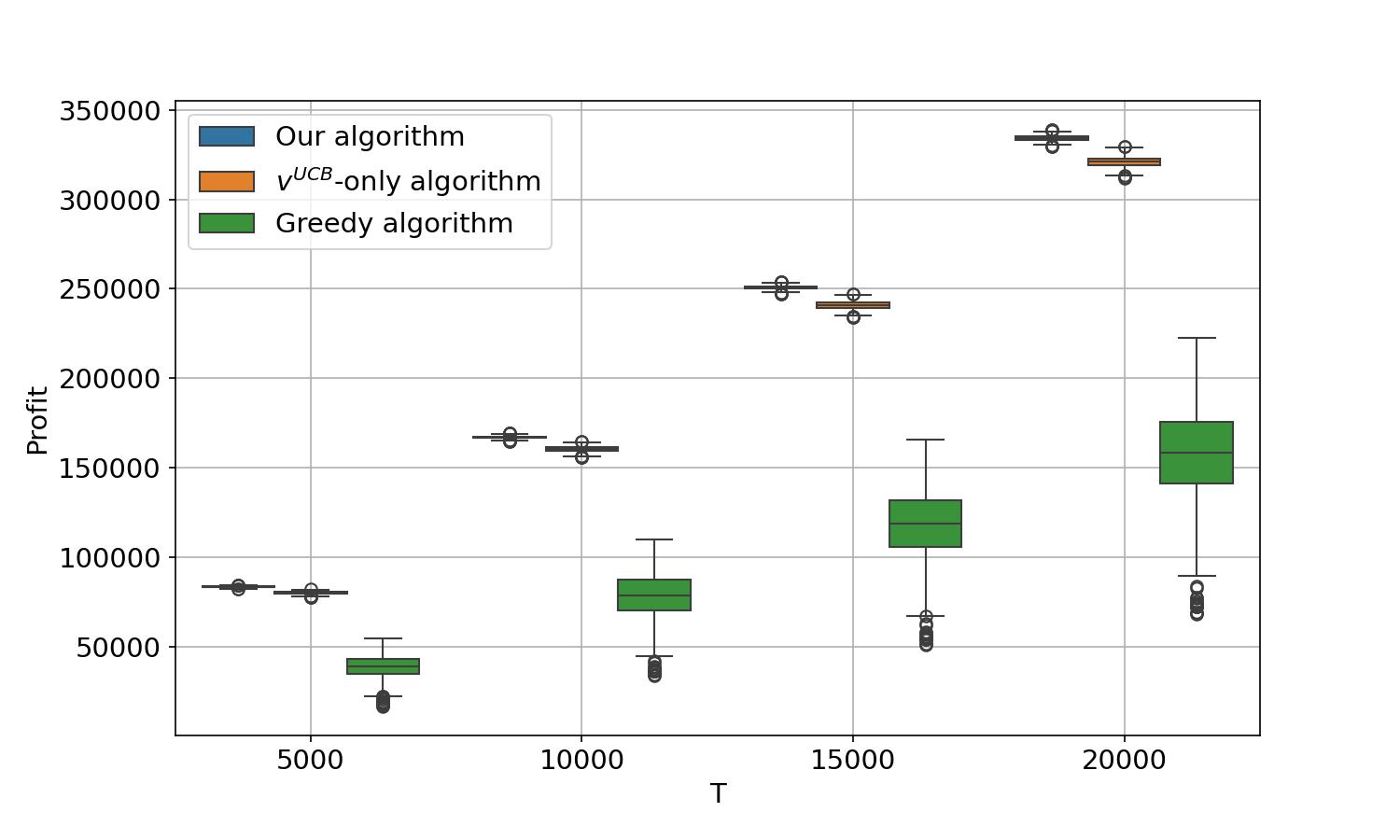}
		\end{minipage}
	}
        \subfigure[Setting $4$]
	{
		\begin{minipage}[b]{.48\linewidth}
			\centering
			\includegraphics[scale=0.4]{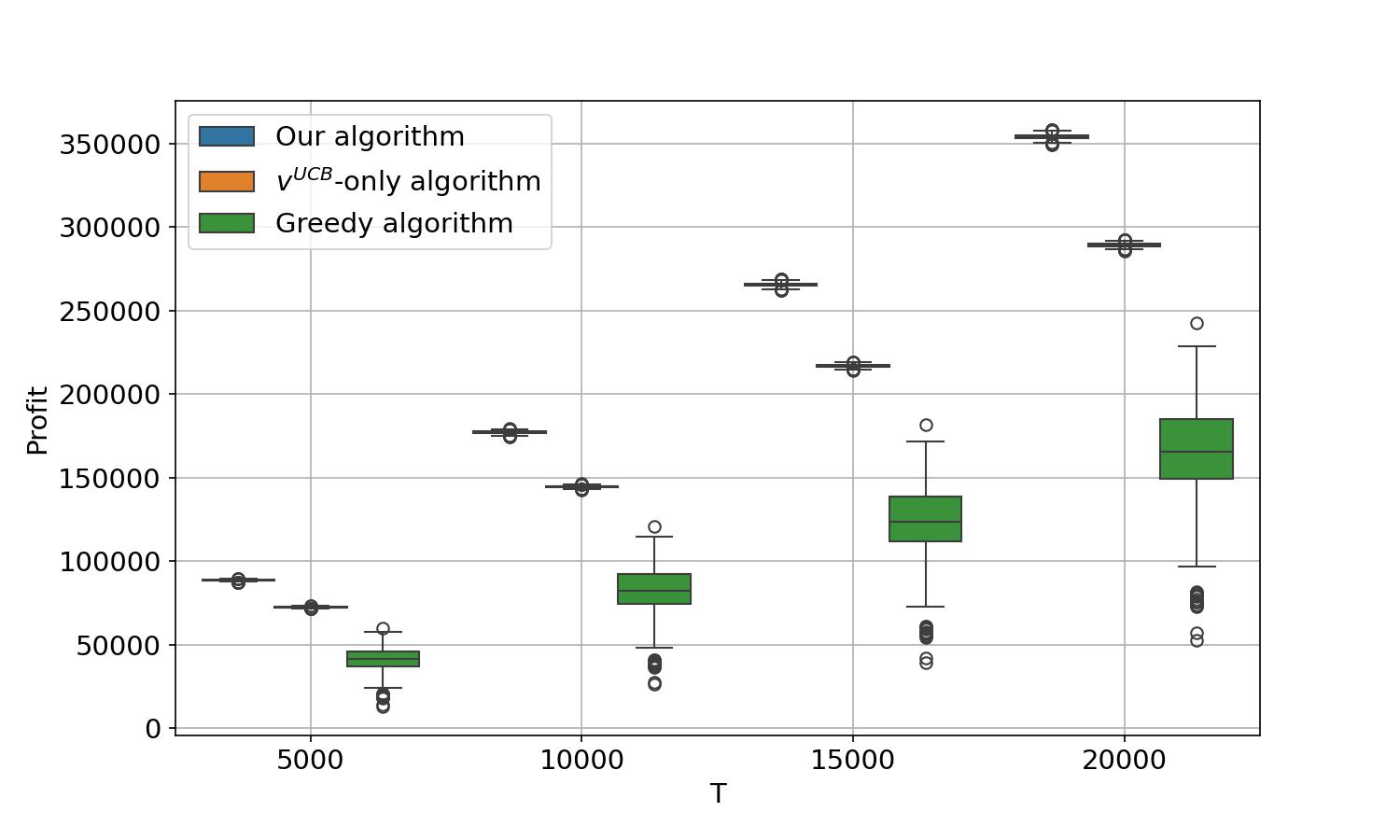}
		\end{minipage}
	}
	\caption{Boxplots of the cumulative profits at four time points, $t \in \{5000, 10000, 15000, 20000\}$. Results are based on $1,000$ independent replications of each experiment.}	\label{fig:large scale experiment}
\end{figure}

\mnote{blue}{New section\\AEQ4}{0mm}\edit{To further assess the robustness of our proposed algorithm, we conduct  additional experiments with randomized parameters.
 These experiments, like those in \Cref{sec: numerical-sushi}, impose no  constraints and utilize the approximate oracle from \cite{liang2021assortment} for all algorithms.
 Unlike the previous experiments where selling prices were randomized and attraction parameters were fixed through calibration, we now fix the selling prices, using the four vectors specified in \Cref{table:r settings}, and randomly generate the attraction parameters from the hypercube $[0.15, 0.2]^N$ for $N = 20$ products. 
 The ordering costs and salvage values for all products remain at $0.1$ and $0$, respectively, and the number of customer arrivals per cycle is now set to $M = 30$. 
We replicate each experiment for $1,000$ times, with   a new vector of attraction parameters for each replication. 
Boxplots in \Cref{fig:large scale experiment} show the cumulative profit  distributions for  different algorithms at time points $T \in \{5000, 10000, 15000, 20000\}$, for each of the four settings in \Cref{table:r settings}. Again, we observe that our proposed algorithm tends to outperform the two benchmarks, which further illustrates the benefit of our proposed exploration-exploitation strategy. 
}

\begin{table}[h]
	\caption{Settings of unit selling prices for the experiments in \Cref{sec: numerical-randomized}.}
    \small
	\label{table:r settings}
	\begin{tabularx}{\textwidth}{p{0.12\textwidth} X}
		\toprule
		 & Unit Selling Prices \\
		\midrule
         \small\bd{Setting 1} & (2.1 , 2.09, 2.09, 2.08, 2.07, 2.04, 2.03, 2.02, 2.02, 1.99, 1.98, 1.98, 1.96, 1.96, 1.96, 1.95, 1.94, 1.92, 1.92, 1.9)\\
		\small \bd{Setting 2}&  (1.5, 1.4, 1.3, 1.2, 1.18, 1.16, 1.14, 1.12, 1.1, 1.1, 1.0, 1.0, 1.0, 1.0, 1.0, 1.0, 1.0, 1.0, 1.0, 1.0) \\
		\small \bd{Setting 3}& (1.2, 1.2, 1.1, 1.1, 1.1, 1.1, 1.1, 1.1, 1.1, 1.1, 1.0, 1.0, 1.0, 1.0, 1.0, 1.0, 1.0, 1.0, 1.0, 1.0) \\
        \small\bd{Setting 4}& (1.19, 1.18, 1.18, 1.18, 1.15, 1.15, 1.15, 1.14, 1.12, 1.12, 1.12, 1.11, 1.1 , 1.1 , 1.09, 1.07, 1.07, 1.06, 1.04, 1.01)\\
		\bottomrule
	\end{tabularx}
\end{table}

\section{Extensions}\label{sec:Extensions}
\subsection{Inventory Carryover} \label{sec:carry over} \mnote{blue}{New section\\AEQ3\\R2Q2}{5mm}

\edit{
Previously, we assumed that the unsold inventory at the end of each cycle was salvaged. We now explore an extension where the unsold inventory can be carried over to the subsequent cycle as its initial inventory. With this modification, decisions made in one cycle have the potential to impact future cycles through the carryover of the initial inventory. Our Algorithm~\ref{alg-main} may no longer be suitable, as it does not consider any initial inventory and hence could prescribe inventory levels that are less than and hence incompatible with the initial inventory carried over. In this section, we extend our algorithm to handle this problem, adopting the ``low switching'' idea proposed in \cite{lyu2024minibatch}. 
}

\edit{
	The ``low switching'' idea is grounded in the observation that the issue of incompatible decisions does not arise when ordering to the same inventory levels across cycles.
	Specifically, for any inventory level at the beginning of a cycle, the remaining inventory at the end of the cycle, after accounting for customer purchases, cannot exceed the beginning level. Therefore, replenishing the inventory to the same  level is inherently compatible. 
	This implies that the problem of incompatible decisions only surfaces when we change inventory ordering levels between cycles. To avoid frequent occurrence of such issues, it is natural to reduce the frequency of switching inventory ordering decisions.
}

\edit{We now provide an overview of our extended algorithm, with the full details presented in \Cref{sec:app carry over}. Our algorithm employs an epoch-based structure, dividing the total $T$ inventory cycles into multiple epochs. Each epoch consists of two distinct stages. During the first stage, the algorithm consistently orders inventory up to the same level, ensuring compatibility as previously discussed. The second stage is dedicated to clearing out inventory, where no replenishment occurs. This clearance stage continues until all remaining inventory is sold, thereby preventing any carryover that could lead to incompatible decisions in the subsequent epoch.  The algorithm then leverages historical data to update the inventory levels for the first stage of the upcoming epoch. It carefully determines the length of the first stage, ensuring on average at most $O(\log T)$ many decision switches (\Cref{lemma:bound L co}). The inventory decisions continue to rely on the same MNL parameter estimator and profit confidence bound as used in Algorithm~\ref{alg-main}, but with updates occurring at a much reduced frequency. \Cref{sec:app carry over} \Cref{thm:main co} analyzes the policy derived from this extended algorithm, utilizing exact static optimization oracles, and provides an upper bound of order $O\left(KMN\sqrt{T\log \left(\sqrt{N}MT\right)}\log T\right)$.}

\edit{
	Although our extended algorithm conveniently manages the inventory carryover scenario with a proven regret guarantee, we recognize that the decreased frequency of decision switching and the inclusion of an inventory clearance stage may lead to diminished profits. Additionally, the regret rate guarantee might not be ideal, with a worse  dependence on parameters such as $K, M, N$ than the regret rate in \Cref{thm-main}. It remains an important question how to develop more sophisticated algorithms that offer improved performance and better theoretical guarantees, which we leave for future research.
}

\subsection{Unknown Customer Arrival Distribution}
\label{sec:mt}
\mnote{blue}{New section\\AEQ2\\R2Q1}{0mm}
\edit{In  previous sections, we assumed that the number of customer arrivals per cycle, \ie, $M_t$ for $t \in [T]$,  followed a known distribution. In this section, we study another extension where the distribution of customer arrivals is not only unknown but also requires estimation.}

\edit{
	We assume that $\{M_t\}_{t=1}^{T}$ are independent and identically distributed (i.i.d) according to an unknown distribution $\Mcal$. This distribution is assumed to belong to a parametric  class $\mathscr{M}$, characterized by certain unknown parameters. For instance, detailed discussions on classes of Poisson and Negative Binomial distributions with unknown parameters can be found in \Cref{app sec:mt detail}. These distribution classes are commonly used  to model count data like customer arrivals \citep{dobson2018introduction}. The unknown parameters in these classes can be  estimated from customer arrival data by standard maximum likelihood estimation.  
}

\edit{Recall that the  one-cycle expected profit function $R(\U; \bv, \bre)$, defined in \Cref{equ-onecycle-revenue}, captures the expected total profit from all customers in one cycle, which inherently  depends on the distribution of the customer arrivals. We now clarify this dependence  by introducing some additional notations. We denote $R(\U; \bv, \bre,m)$ as the one-cycle expected profit when the number of arrivals is fixed at a positive integer $m$. Then the expected profit  under a distribution $\Mcal$ of customer arrivals can be written as follows:
\begin{align*}
    R^{\Mcal}(\U; \bv, \bre) \coloneqq \sum_{m = 1}^\infty \mathbb{P}_{\Mcal}\left({M = m}\right)\Eb{R(\U; \bv, \bre, M)\mid M=m},
\end{align*}
where $\mathbb{P}_{\Mcal}\left({M = m}\right)$ is the probability of $m$  customer arrivals in a cycle under the distribution $\Mcal$.  
}

\edit{
	To adapt our  Algorithm~\ref{alg-main} to handle the unknown distribution $\Mcal$, we need to  estimate this distribution in addition to  our existing computations. 
	Specifically, at the start of each inventory cyle $t$, alongside constructing $\bvUCB{t}$ and $\hat\bre_t$, we also build a distribution estimator $\hat{\Mcal}_t$ using   historical customer arrival data. We then  determine the inventory decision $\U_t$ by solving the optimization problem $\max_{\U\in\setU}R^{\hat\Mcal_t}(\U; \bvUCB{t}, \hat\bre_t)$. We show in appendix that the estimated profit $R^{\hat\Mcal_t}(\U; \bvUCB{t}, \hat\bre_t)$ upper bounds the true expected profit function $R^{\Mcal}(\U; \bv, \bre)$ when the estimated distribution $\hat\Mcal_t$ stochastically dominates the true distribution $\Mcal$. We further discuss the  construction of such distribution estimators for the  Poisson  and Negative Binomial classes in \Cref{sec:poisson,sec:neg-binomial} respectively. Leveraging our previous analyses and bounding the errors in distribution estimation, we derive a regret upper bound of order $O\left(\sqrt{MNT\log\left(\sqrt{MN}T + 1\right)}\right)$ in \Cref{thm:upperbound mt}. This bound is comparable to that in \Cref{thm-main}, indicating that the impact of estimating the customer arrival distribution on regret is relatively minor. 
}

\section{Concluding Remarks}
\label{sec-conclusion}

In this paper, we study the online joint assortment-inventory optimization problem under the widely used multinomial logit (MNL) choice model with {unknown} attraction parameters. 
In this problem, the assortment set can dynamically change due to stochastic product stockout events, so customers' choices display intricate substitution patterns. 
As a result, the joint assortment-inventory optimization  problem is notoriously difficult even with known  attraction parameters. 
In this paper, we tackle the more challenging online optimization problem with unknown attraction parameters. We propose a novel algorithm that can simultaneously learn the attraction parameters and maximize the expected total profit. 
This algorithm builds on a new estimator for MNL attraction parameters, and features novel upper confidence bounds on decisions' expected profits based on adaptively tuning products' unit profits.
The proposed tuning of the known unit profit parameters provides a novel way to incentivize exploration. 
We analyze the regret of our algorithm when applying different optimization oracles to single-cycle static joint assortment-inventory planning problems. In particular, we prove a regret upper bound for our algorithm and a worst-case lower bound for our problem, showing that our algorithm achieves a nearly optimal regret rate when applying an exact optimization oracle for single-cycle problems. At last, we use numerical experiments to demonstrate that our algorithm effectively balances the exploration and exploitation and achieves desirable performance.

To the best of our knowledge, this is the first paper that studies online joint assortment-inventory optimization  under the MNL model. 
There are many exciting  future research directions on this topic. %
For example, in many applications there are a large number of products so learning the demand distribution for each product separately may be inefficient. Exploring possible ways to reduce the dimensionality of products holds potential interest. For instance, it is possible that the products can be effectively described by a small number of structured product features. It would be interesting to study how to leverage such structure to efficiently learn the demands for the products and solve for the optimal assortment-inventory decisions.

\bibliographystyle{informs2014} %
\bibliography{bibfile-new} %

\ECSwitch
\EquationsNumberedBySection

\ECHead{Electronic Companions}

	\section{Notations} \label{notation}
In the paper, boldface letters such as $\U, \bv, \bre$ denote vectors. Arithmetic operations on a pair of vectors of equal length should be understood element-wise.  %
Calligraphic letters such as $\s$ and $\setU$ denote sets. For a set $\s$, $|\s|$ denotes the cardinality of the set. For any positive integer $N$, $[N]$ denotes the set $\{1, 2, \dots, N\}$. The use of $O$ and $\Omega$ is standard, that is, given functions $f, g: \mathbb{N}\rightarrow [0, \infty)$, we say
\begin{equation*}
	f(x) = O(g(x)) \quad\text{if}\quad \limsup_{x\rightarrow \infty}{ \frac{f(x)}{g(x)} < \infty},
\end{equation*}
and
\begin{equation*}
	f(x) = \Omega(g(x)) \quad \text{if}\quad \liminf_{x\rightarrow \infty}{ \frac{f(x)}{g(x)} > 0}.
\end{equation*}

\section{A Detailed Analysis of \Cref{example-counter-example}} \label{app-counter-example}
In this section, we provide a detailed analysis of \Cref{example-counter-example} in \Cref{subsec-naive-extension}. 
Recall that in \Cref{example-counter-example}, the number of products $N$ is 2, the product-wise capacity vector $\cvec$ is $(1, 1)$, the total capacity $\C$ is 2, and the numbers of customer arrivals $\M_1, \dots, M_T$ are all equal to 2. Suppose that the attraction parameters of the no-purchase option and product $1$ are $v_0 = v_1 = 1$, and the unit profit of product $1$ is $r_1 = 1$. The attraction parameter $v_2$ and the unit profit $r_2$ of product $2$ are left undetermined, except that $r_2$ is restricted to satisfy $r_2 < r_1$. 

There are four feasible inventory decisions: $\U^{(0) }= (0, 0)$, $\U^{(1)}= (1, 0)$, $\U^{(2)} = (1, 1)$, and $\U^{(3)} = (1, 0)$. Obviously, the decision $\U^{(0)}$ is not optimal. Below we show the single-cycle expected profits associated with the remaining three decisions. These are calculated by considering all potential selections made by the two arriving customers during an inventory cycle. 
\begin{align*}
	R(\U^{(1)}; \bv, \bre ) 
	=& \left(1 - \frac{1}{(v_0 + v_1)^2}\right) \cdot r_1 = \frac{3}{4},\\
	R(\U^{(2)}; \bv, \bre) 
	=& \frac{v_1}{v_0 + v_1 + v_2}\cdot \frac{v_2}{v_0 + v_2} \cdot (r_1 + r_2) + \frac{v_1}{v_0 + v_1 + v_2}\cdot \frac{v_0}{v_0 + v_2} \cdot r_1\\
	&+ \frac{v_2}{v_0 + v_1 + v_2}\cdot \frac{v_1}{v_0 + v_1} \cdot (r_1 + r_2) + \frac{v_2}{v_0 + v_1 + v_2}\cdot \frac{v_0}{v_0 + v_1} \cdot r_2\\
	& + \frac{v_0}{v_0 + v_1 + v_2}\cdot \frac{v_1}{v_0 + v_1 + v_2} \cdot r_1 + \frac{v_0}{v_0 + v_1 + v_2}\cdot \frac{v_2}{v_0 + v_1 + v_2} \cdot r_2.\\
	=& \frac{1}{2 + v_2} \cdot  \Big[  \frac{v_2\cdot (1 + r_2)}{1 + v_2}  +  \frac{1}{1 + v_2}
	+ \frac{v_2\cdot (1 + r_2) }{2} + \frac{v_2\cdot r_2 }{2}
	+  \frac{1}{2 + v_2} + \frac{v_2\cdot r_2}{2 + v_2}  \Big]\\
	= & \frac{v_2r_2}{1 + v_2}  + \frac{1}{2} + \frac{1 + v_2r_2}{(2 + v_2)^2},\\
	R(\U^{(3)}; \bv, \bre) = & \left(1 - \frac{1}{(v_0 + v_2)^2}\right) \cdot r_2 = \frac{v_2r_2\cdot (2 + v_2)}{(1 + v_2)^2}.	
\end{align*}

Based on the above given closed-form expressions for expected profits, we can make several important observations. 

\subsubsection*{Observation 1: the optimal inventory decision is either $\U^{(1)}$ or $\U^{(2)}$, but not $\U^{(3)}$.}
We can easily verify that the inventory decision $\U^{(3)}$ is dominated by $\U^{(2)}$: 
\begin{align*}
	R(\U^{(2)}; \bv, \bre) - R(\U^{(3)}; \bv, \bre) &=  \frac{1}{2} + \frac{1 + v_2r_2}{(2 + v_2)^2} -  \frac{v_2r_2}{(1 + v_2)^2}\\
	&= \frac{1 + v_2^2 + 2v_2 \cdot (1 - r_2)}{2(1 + v_2)^2} + \frac{1 + v_2r_2}{(2 + v_2)^2} \ >\  0,
\end{align*}
where the inequality follows from the assumption that $r_2 < r_1 = 1$. 
Therefore, the optimal inventory decision is either $\U^{(1)}$ or $\U^{(2)}$, depending on the values of $v_2$ and $r_2$. 

\subsubsection*{Observation 2: the single-cycle expected profit function $R(\U; \bv, \bre)$ may not be monotone in $\bv$.} 
We now show that when $r_2 \in (1/5, 1/4)$, $R(\U^{(2)};\bv, \bre)$ is not a monotone function of $v_2$. 
In particular, we show the existence of two intervals of $v_2$ such that $R(\U^{(2)};\bv, \bre)$ increases with $v_2$ on one interval but decreases with $v_2$ on the other interval. 

Taking the partial derivative of $R(\U^{(2)};\bv, \bre)$ with respect to $v_2$, we have
\begin{equation*}
	\frac{\partial R(\U^{(2)}; \bv, \bre)}{\partial v_2} = \frac{r_2}{(1 + v_2)^2} + \frac{2r_2 - r_2v_2 - 2}{(2 + v_2)^3}.
\end{equation*}
When $r_2 \in (1/5, 1/4)$, we have $\frac{\partial R(\U^{(2)}; \bv, \bre)}{\partial v_2}\vert_{v_2 = 0} = (5r_2 - 1)/4 > 0$ while $\frac{\partial R(\U^{(2)}; \bv, \bre)}{\partial v_2}\vert_{v_2 = 1} = (31r_2 - 8)/108 < 0$. 
Since $\frac{\partial R(\U^{(2)}; \bv, \bre)}{\partial v_2}$ is a continuous function of $v_2$, there exists $v_2', v_2'' \in (0, 1)$ such that $\partial R(\U^{(2)}; \bv, \bre) / \partial v_2 \mid_{v_2} > 0$ for any $v_2 \in (0, v_2')$ and $\partial R(\U^{(2)}; \bv, \bre) / \partial v_2 \mid_{v_2} < 0$ for any $v_2 \in (v_2'', 1)$. 
This verifies our second observation. 

\subsubsection*{Observation 3: the heuristic approach in \Cref{eq: heuristic} may have a linear regret rate. }
If we follow the heuristic approach in \Cref{eq: heuristic}, we would determine the inventory decision $\bre_{t}$ for cycle $t$ by maximizing $R(\U; \bv^{\UCB}_{t}, \bre)$ over feasible inventory decisions $\U \in \setU$, where $\bv^{\UCB}_{t}$ is the upper confidence bound on the attraction parameter $\bv$ calculated from data accumulated in the first $t - 1$ cycles. 
We now show that this heuristic approach may get stuck in a sub-optimal decision due to lack of exploration, and thus it can have a linear regret rate.

\begin{figure}[htbp]
	\begin{center}
		\includegraphics[height=2.3in]{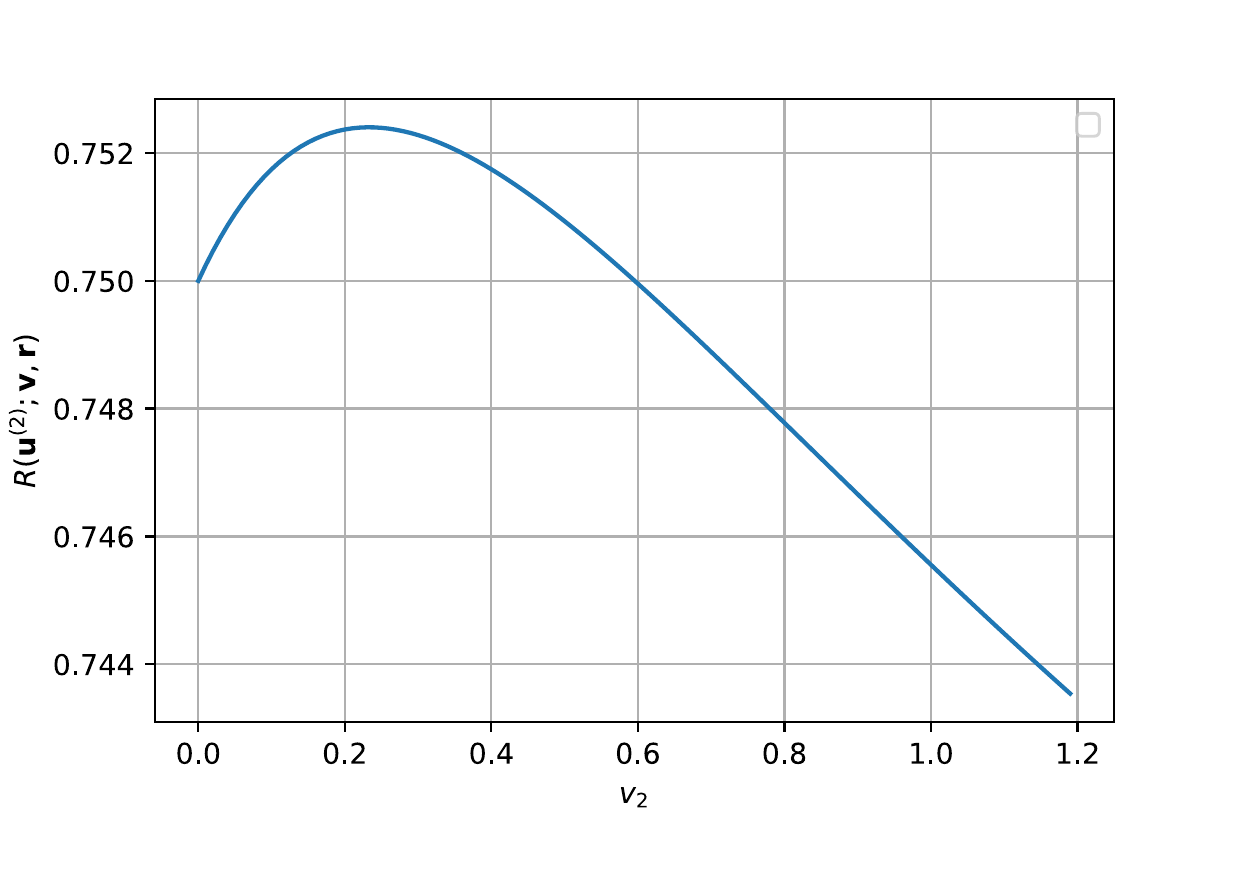}
		\caption{The expected profit of $\U^{(2)}$ as a function of $v_2$. } \label{fig-counter-app}
	\end{center}
\end{figure}

We note that when $r_2 \in (1/5, 1/4)$, there exists a range of values of $v_2$ such that the inventory decision $\U^{(2)}$ is optimal. 
For example, \Cref{fig-counter-app} plots $R(\U^{(2)}; \bv, \bre )$ across different values of $v_2$ when $r_2 = 0.22$. 
We observe that $R(\U^{(2)}; \bv, \bre)$ first increases and then decreases with $v_2$.
There exists $\tilde{v}_2$ around $0.6$ such that when $v_2 \in (0, \tilde{v}_2)$, we have $R(\U^{(2)}; \bv, \bre) > R(\U^{(1)}; \bv, \bre ) = 0.75$, that is, the inventory decision $\U^{(2)} = (1, 1)$ is optimal. 
Therefore, when $r_2 = 0.22$ and $v_2 \in (0, \tilde{v}_2)$, we should assort both the first and second products into our inventory. 

However, when $\U^{(2)} = (1, 1)$ is optimal, the heuristic approach in \Cref{eq: heuristic} may always miss the second product and get stuck in the suboptimal decision $\U^{(1)}$. 
Specifically, the heuristic approach would start with $\bv_{1}^{\UCB} = (v_{\max}, v_{\max}) = (1, 1)$. Since $R(\U^{(2)}; \bvUCB{1}, \bre) = 0.745 < 0.75 = R(\U^{(1)}; \bvUCB{1}, \bre)$, the heuristic approach would implement $\U^{(1)} = (1, 0)$ in the first cycle. 
However, this decision $\U^{(1)}$ does not assort product 2, the upper confidence bound of $v_2$ cannot be updated. Consequently, the algorithm will persistently select $\U^{(1)}$ throughout, resulting in linear total regret in the long run.

\subsubsection*{Observation 4: our proposed approach in \Cref{equ-opt-epoch} works.}
According to our proposed algorithm, we select the decision $\U_t$ for cycle $t$ by maximizing $R(\U; \bv^{\UCB}_{t}, \hat\bre_{t})$ over feasible inventory decisions $\U \in \setU$. 
Consequently, our algorithm would start with $\bv_{1}^{\UCB} = (v_{\max}, v_{\max}) = (1, 1)$ and $\hat \bre_{1} = (r_{\max}, r_{\max}) = (1, 1)$. Since $R(\U^{(2)}; \bvUCB{1}, \hat\bre_1) = 11/9 > 3/4 = R(\U^{(1)}; \bvUCB{1}, \hat\bre_1)$, our algorithm would implement the optimal decision $\U^{(2)} = (1, 1)$ and explore both products in the first cycle. 
After collecting the data in the first cycle, our algorithm updates the upper bounds on attraction parameters and unit profits, and continues to explore decisions with high uncertainty in all subsequent cycles.

\section{A Step-by-Step Explanation to Algorithm~\ref{alg-main}}
\label{app-sec-alg-step}
In this section, we provide a step-by-step explanation of Algorithm~\ref{alg-main} in \Cref{subsec-adjust-revenue}. According to Algorithm~\ref{alg-main}, we first initialize the upper bounds on the attraction parameter $\bv$ and the unit profits $\bre$ to $\bvUCB{1} = (v_{\max}, \dots, v_{\max})$, $\hat\bre_{1} = (r_{\max}, \dots, r_{\max})$, and initialize the inventory decision $\U_1$ for the first cycle. For each cycle $t$, 
\textit{lines~\ref{alg-startWhile}} to \textit{\ref{alg-end-while}} count, for each product $i$, 1) the number of no-purchases $n_i$ between two consecutive purchases of product $i$ and 2) the number of times $k_i$ that product $i$ has been purchased. Specifically, in cycle $t$, for an assorted product $i$, each purchase of the product triggers the algorithm to update the cumulative purchase count $k_i$, record the no-purchase count $\mu_{i, k_i}$, and reset the counting variable $n_i$ to zero. Furthermore, after each purchase of $i$, the algorithm continues to update the no-purchase counting variable $n_i$ for each no-purchase, as long as the no-purchase occurs when  product $i$ is available (\ie, $i\in\mathcal{S}_{t, m}$).

At the end of each cycle, \textit{lines~\ref{alg-all-prod} to \ref{alg-all-prod-2}} update the cycle index $t$, the cumulative purchase count $k_{i, t}$, and compute $\bar \recv_{i,t}$, $\mu_{i, t}^{\LCB}$, $\mu_{i, t}^{\UCB}$, $v_{i, t}^{\LCB}$, $v_{i, t}^{\UCB}$, $\hat r_{i, t}$ for each assorted product $i \in [N]$ of the cycle. Finally, the algorithm needs to determine an inventory decision for the next cycle. In particular, \textit{line~\ref{alg-calculate-u}} solves \eqref{equ-opt-epoch} to obtain the inventory decision $\U_{t}$.

{ 
	\section{Maximum Likelihood Estimation} \label{sec-MLE}\mnote{blue}{New section\\AEQ1\\R1Q1}{-10mm}
	
	In this section, we adapt the maximum likelihood estimation (MLE) to our setting, expanding on the high-level discussions in \Cref{subsec:mle}. We first detail the estimation process and online decision-making, followed by a numerical comparison of our proposed estimator with MLE.    
	
	\subsection{Online Algorithm with MLE}\label{sec-mle-estimator}
	To apply MLE to our problem, we reparameterize our MNL choice model in terms of $\btht^* = (\theta_1^*, \cdots, \theta_N^*)$ where $\theta_i^* = \log v_i$ for $i \in [N]$ and $\theta_0^* = 0$. For any potential value $\btht = (\theta_1, \dots, \theta_N)$ of the unknown $\btht^*$, the choice probability given an assortment set $\s$ can be written as follows:  
	\begin{align*}
		p_{i}(\s \mid \btht) = \begin{cases}
			\frac{\exp(\theta_i)}{1 + \sum_{j\in \s} \exp(\theta_j)}, & \text{ if } i\in\s\cup\{0\} \\
			0, & \text{ otherwise }
		\end{cases}.
	\end{align*}
	At the beginning of each cycle $t$,
	we construct MLE for the unknown parameter $\btht^*$  by  forming the log-likelihood function from the observations of historical assortment sets and customer choices:   
	\begin{align*}
		L_t(\btht) \coloneqq &\  \sum_{s=1}^{t-1}\sum_{m=1}^{M_{s}}\sum_{i\in \s_{s,m}\cup\{0\}}\1{d_{s,m}=i}\log p_{i}(\s_{s,m}\mid \btht),
	\end{align*}
	where $\s_{s, m}$ is the assortment set displayed to the $m$-th  customer in the $s$-th cycle and $d_{s,m} \in \{0\}\cup [N]$ denotes the customer's choice.
	Maximizing the log-likelihood function 
	$L_t(\btht)$  yields the maximum likelihood estimator $\hat\btht_t$. Since the log-likelihood function is convex, the  estimator $\hat\btht_t$ can be also characterized by the first order condition. Hence, we can implement MLE by using off-the-shelf solvers  to either maximize the log-likelihood function or solve the first-order condition equation. 
	\begin{restatable}{lemma}{MLEFOC}
		\label{lemma:MLE_FOC}
		At the beginning of each inventory cycle $t$, the maximum likelihood estimate $\hat\btht_t$ is the solution to the following equation: 
		\begin{equation}
			\nabla L_{t}(\btht) = \sum_{s=1}^{t-1}\sum_{m=1}^{M_{s}}\sum_{i\in \s_{s,m}}
			\left[ \1{d_{s,m}=i} - p_{i}(\s_{s,m}\mid \btht)\right] \eb_i = \boldsymbol{0}, \label{eq:MLE_FOC}
		\end{equation}
		where $\eb_i \in \mathbb{R}^N$ is a vector with its $i$-th entry as $1$ and all other entries as $0$.
	\end{restatable}
	
	For the purpose of online algorithm design, we need to further quantify the uncertainty in MLE. With a slight abuse of notation, we define $n_{i,t} \coloneqq \sum_{s=1}^{t-1}\sum_{m=1}^{M_{s}}\1{i\in \s_{s,m}}$ as the cumulative number of customer arrivals for whom product $i$ is available in the first $t - 1$ cycles. The following lemma provides a probabilistic error bound on $\hat\btht_t$. 
	
	\begin{restatable}{lemma}{thetaconvergence}\label{lemma:theta convergence}
		\edit{Let $\kappa$ be a positive constant such that $\kappa \le \inf_{\norm{\btht-\btht^*}{}\le 1,i\in[N],\s}  p_{i}(\s \mid  \btht)p_{0}(\s \mid \btht)$ and define a constant $\alpha_t$ for any cycle $t$ as follows:
			\begin{align}\label{eq: alphat}
				\alpha_t \coloneqq \frac{1}{2\kappa}\sqrt{\frac{N}{2}\log\left(1 + \frac{2\sum_{s=1}^{t-1} M_s}{N}\right) + \log {t}}.
			\end{align}
			For any cycle $t$ such that $\min_{i\in[N]}n_{i, t} \ge \alpha_t^2$, we have 
			\begin{align}
				\label{eq:theta converge 1}
				\prob{\sum_{i\in [N]}(\hat\theta_{i,t}-\theta_{i}^*)^2 \cdot n_{i,t} \ge \alpha_t^2} \le \frac{1}{t}.
			\end{align}
			In particular, inequality~\eqref{eq:theta converge 1} holds for $\kappa = \frac{v_{\min}}{e\left(1 + Kv_{\max}\right)^2}$.
		}
	\end{restatable}
	\Cref{lemma:theta convergence} motivates natural confidence bounds for each $\theta_i^*$ as follows: the lower confidence bound (LCB) is given by $\theta_{i, t}^{\text{LCB}} = \hat\theta_{i, t} - {\alpha_t}/{\sqrt{n_{i,t}}}$ and  the upper confidence bound (UCB) is $\theta_{i, t}^{\text{UCB}} =  \hat\theta_{i, t} + {\alpha_t}/{\sqrt{n_{i,t}}}$.
	Using these bounds, we can derive confidence intervals for the attraction parameters $v_i$'s:
	\begin{align}\label{eq: CI-v-MLE}
		\vUCB{i,t} \coloneqq \min\left\{v_{\max},\ \exp\left(\hat\tht_{i,t} + {\alpha_t}/{\sqrt{n_{i,t}}}\right)\right\},\quad 
		\vLCB{i,t} \coloneqq \exp\left(\hat\tht_{i,t} - {\alpha_t}/{\sqrt{n_{i,t}}}\right).
	\end{align}
	According to \Cref{lemma:theta convergence}, theses confidence bounds require  specifying  a constant $\kappa$. This constant must  lower bound the minimum eigenvalue of the Hessian matrix of the log-likelihood function, uniformly over all $\btht$ vectors within a unit ball centered at the true parameter vector $\btht^*$. Identifying such a constant can be challenging. However, when $v_{\min}$ is known, \Cref{lemma:theta convergence} provides a theoretically valid choice of $\kappa$: $\kappa = \frac{v_{\min}}{e\left(1 + Kv_{\max}\right)^2}$, which we will evaluate numerically in \Cref{app-sec: mle-experiment}. Moreover, the validity of the confidence bounds above relies on the condition $\min_{i\in[N]}n_{i, t} \ge \alpha_t^2$, \ie, all products have been purchased sufficiently often by cycle $t$, relative to the threshold $\alpha_t$. To ensure that this condition holds, we propose to constantly monitor this condition in our online algorithm. Whenever this condition is violated because $n_{i, t} < \alpha_t^2$ for some under-explored  product $i$, we additionally explore this product until the condition is restored. 
	
	Following \eqref{equ-define-rhat} and with a slight abuse notation, we further tune the unit profit parameter $\bre$  at the beginning of each cyle $t$ according to the following rule: 
	\begin{align}\label{eq: adjust-r-MLE}
		\hat r_{i,t}\coloneqq \min\{1, r_i + \delta_{i, t} \}, \text{ where } \delta_{i, t} \coloneqq \frac{v_{i, t}^{\UCB}}{v_{i, t}^{\LCB}}  - 1.
	\end{align}
	Subsequently, we update the inventory decision again by optimizing the profit UCB, as proposed in \Cref{subsec-adjust-revenue}, setting $\U_{t} \in \argmaxx_{\U \in \setU} \R(\U; \bv^{\UCB}_{t}, \hat \bre_{t})$ according to \Cref{equ-opt-epoch}.  
	
	Algorithm~\ref{alg: main mle} summarizes our revised online decision-making algorithm. This algorithm is similar to Algorithm~\ref{alg-main}, but differs in two aspects. Firstly, it employs the confidence bounds and adjusted profits from \Cref{eq: CI-v-MLE,eq: adjust-r-MLE}, which are based on MLE rather than our  counting estimator in \Cref{sec-algorithm} (\Cref{line:vit-mle,line:rit-mle}).
	Secondly, it incorporates exploratory phases for products that are under-explored, leading to violations of the condition $\min_{i\in[N]}n_{i, t} \ge \alpha_t^2$, thereby ensuring the validity of the MLE confidence bounds per \Cref{lemma:theta convergence} (\Cref{line:exploration-mle}). Specifically, we  define the set of exploratory cycles as 
	\begin{align}\label{eq:explore-phase}
		\setexp \coloneqq \left\{t \mid \exists i \in[N] \text{ such that } n_{i,t} < \alpha_t^2 ,\ 1\le t\le T\right\}.
	\end{align}
	For any exploratory cycle $t$ in the set $\setexp$, our inventory decisions focus exclusively on products that have not been adequately explored. We repeatedly and randomly add product $i \in [N]$  with $n_{i,t} < \alpha_t^2$ to the inventory, until the requirement $\min_{i\in[N]} n_{i,t} \ge \alpha_t^2$ is met.
	Notably, we need to ensure adequate exploration of  all products to validate the confidence bounds for each individual product. 
	In stark contrast,  the confidence bounds derived from our counting estimator rely solely on the adequate exploration of the product in question, as indicated by $Q_{i, t} < 1/48$ in \Cref{lemma: coverage-probability,lemma:convergence-barmu}, regardless of the exploration status of other products. This means that our confidence bounds for different products are decoupled. We leverage this decoupling property to directly bound the regret that arises when the adequate exploration condition $Q_{i, t} < 1/48$ is not met for a particular product, thereby  avoiding additional forced explorations to meet the condition $Q_{i, t} < 1/48$ for every product $i\in[N]$.

	Finally, assume that we solve \eqref{equ-opt-epoch} with an exact oracle, and let $\pi_{\text{mle}}$ denote the policy derived by the  Algorithm~\ref{alg: main mle} with MLE. The next theorem upper bounds the total regret of $\pi_{\text{mle}}$. 
	
	\begin{restatable}{theorem}{mainmle}\label{thm:main mle}
		For any instance of the online joint assortment-inventory optimization problem with $N$ products, attraction parameters $\bv=(v_i)_{i \in [N]}$, unit profits $\bre = (r_i)_{i \in [N]}$, maximum assortment cardinality $K$, and expected number of customers $M$ per cycle, the worst-case total regret of the policy $\pi_{\mathrm{mle}}$ generated by Algorithm~\ref{alg: main mle} with MLE over $T$ inventory cycles satisfies
		\begin{equation*}
			\Reg(\pi_{\mathrm{mle}} ; T) = O(NK^2\sqrt{MT \log (MT) }) \ .
		\end{equation*}
	\end{restatable}
	
	The regret bound in 
	\Cref{thm:main mle} is comparable to that in \Cref{thm-main} for Algorithm~\ref{alg-main} using our proposed counting estimator, at least in terms of the order of $T$ and $M$. However, it appears to have a worse dependence on $N$ and $K$. This may be an artifact of our theoretical analysis rather than a reflection of the MLE-based algorithm’s intrinsic weakness.
	We stress that our Algorithm~\ref{alg: main mle} and the theoretical analysis in \Cref{thm:main mle} represent  preliminary efforts to adapt MLE to the online joint inventory-assortment optimization problem. While these contributions advance the existing literature that applies MLE to online assortment optimization without inventory limits, they may have significant room for improvement. In particular,  it remains an open question whether the forced exploration phases for cycles in \Cref{eq:explore-phase} can be avoided, whether the MLE confidence bounds can be improved,  and whether the regret bound in 
	\Cref{thm:main mle} can be further tightened. However, these questions are beyond the scope of our current paper, as our main focus is on Algorithm~\ref{alg-main} using our proposed counting estimator. Given the challenge in comparing MLE and our proposal theoretically, we further compare them numerically in the next subsection. 
	
	\begin{algorithm}[t]
		\label{alg: main mle}
		\small
		\DontPrintSemicolon
		\KwInit{$\vUCB{i,1} = v_{\max}$,\ $\hat\re_{i,1} = r_{\max}$,\ $n_i = 0$,\ $k_{i} = 0$,\ 
			$\forall i\in [N]$\; 
			$\bv_{1}^{\UCB} = (v_{1,1}^{\UCB}, \dots,  v_{N,1}^{\UCB})$,\ $\hat\bre_1 = (\hat r_{1, 1}, \dots, \hat r_{N, 1})$,\ 
			$\U_{1} \in \arg\max_{\U\in \setU}R(\U; \bvUCB{1}, \hat\bre_{1})$,\  $t = 1$}
		\While{$t < T$}
		{
			Order up to $\U_t$, and observe the choices  $\seq_t = (d_{t, m})_{m \in [M_t]}$ made by $M_t$ customers in  cycle $t$\;
			\For{$m \gets1$ \KwTo $M_t$} {
				\label{alg-startWhile mle}
				\For{$i\in\s_{t,m}$}{$n_i = n_i + 1$}
			}{\label{alg-end-while mle}}
			$t = t + 1$ \;
			\For{$i\gets1$ \KwTo $N$}{
				\label{alg-all-prod mle}
				Update $n_{i,t} = n_i$\;
				Update $\alpha_t$ per \Cref{eq: alphat}\;
				Compute $v_{i,t}^{\LCB}, v_{i,t}^{\UCB}$ per \Cref{eq: CI-v-MLE}\;\label{line:vit-mle}
				Compute $\hat \re_{i, t}$ according to \Cref{eq: adjust-r-MLE}\;\label{line:rit-mle}}
			\label{alg-all-prod-2 mle}
			Set $\bv_{t}^{\UCB} = (v_{1,t}^{\UCB}, \dots,  v_{N,t}^{\UCB})$ and $\hat\bre_t = (\hat r_{1, t}, \dots, \hat r_{N, t})$ \;
			\eIf{$\exists i \in [N]$ such that $n_{i, t} < \alpha_t^2$\label{alg-exp-epoch-1}}
			{
				Randomly choose  $\U_{t} \in \left\{\U \in \setU \setminus \{\mathbf{0}\} \mid u_i \in \{0, 1\} \textrm{ if } n_{i, t} < \alpha_t^2 \textrm{ and } 0 \textrm{ otherwise}, \forall i\in [N] \right\}$ \label{line:exploration-mle}
			}
			{
				Compute $\U_{t} \in \arg\max_{\U\in \setU}R(\U; \bvUCB{t}, \hat\bre_{t})$\;
			}\label{alg:explore cycle mt}
		\label{alg-calculate-u mle}
	}		
	
	\caption{Exploration-Exploitation Algorithm with MLE}
\end{algorithm}

\subsection{Numerical Comparisons of MLE and Our Proposed Method}\label{app-sec: mle-experiment}
In this subsection, we numerically compare our proposed counting estimator (CE) and MLE. The comparison aims to highlight the benefits of our CE in multiple dimensions, including computational efficiency, estimation accuracy, and the tightness of confidence intervals.

In addition to the standard MLE described in \Cref{sec-mle-estimator}, we also evaluate a stochastic gradient descent (SGD) variant. This variant employs online gradient descent updates instead of fully re-optimizing the log-likelihood function repeatedly, which significantly accelerates the computation. Specifically, we consider the SGD method introduced by \cite{Oh_Iyengar_2021}, detailed in their appendix section titled  ``Online Parameter Update.''

This SGD method begins by estimating the MNL parameters through direct maximization of the log-likelihood until each product has been purchased at least 
$K$ times, with 
$K$ being the upper limit on the assortment size. Then the method transitions to online gradient updates for the MNL parameters, performing one update per customer choice.
For the $m$-th customer arrival in the $s$-th cycle, 
let $\hat\btht_{s, m-1} = (\hat\theta_{1, s, m-1}, \dots, \hat\theta_{N, s, m-1})$ denote the MNL parameter estimate prior to observing this customer’s choice. Upon observing the customer's choice $d_{s, m}$
given the assortment set $\s_{s, m}$, the method calculates the log-likelihood gradient for this new data point as $\sum_{i \in \s_{s, m}}\left[ \1{d_{s,m}=i} - p_{i}(\s_{s,m}\mid \btht)\right] \eb_i$. The parameter estimate is then updated along the gradient direction:
\begin{align*}
	\hat \theta_{i, s, m} = \begin{cases}
		\hat \theta_{i, s, m-1} + \frac{2}{\kappa\cdot n_{i, s, m-1}}\mbr{\1{d_{s, m}=i} - p_{i}(\s_{s, m}\mid \hat \btht_{s, m-1})} & \text{ if } i\in \s_{s, m} \\
		\hat \theta_{i, s, m-1} & \text{otherwise}
	\end{cases}, ~~ \forall i \in [N], 
\end{align*}
where the update step size involves the constant $\kappa$ that satisfies the condition in \Cref{lemma:theta convergence} and $n_{i, s, m-1}$ is the total number of updates made for the $i$-th product's estimate before  the $m$-th customer in the $s$-th cycle. The attraction parameter $\bv$ can be estimated by applying an  element-wise exponential transformation to $\hat\btht_{s, m} = (\hat\theta_{1, s, m}, \dots, \hat\theta_{N, s, m})$. By avoiding repeated optimization of the log-likelihood function, this SGD approach is significantly faster than the standard MLE method.

We now detail the simulation setup used to compare the different estimation methods. In our simulations, we set the number of products to $N = 5$, the total horizon to $T = 5000$, the inventory capacity limit to $\bar{c} = 5$, the assortment cardinality limit to $K = 5$, and the average number of customers per cycle to $M=5$ or $M = 10$ respectively.  The attraction parameters 
$\bv$ are randomly generated from a uniform distribution over  $[0, 1]^N$. 
To collect data for the estimators, we employ the same purely exploratory policy, which randomly picks an inventory level from all feasible ones at the beginning of each cycle. We choose this data collection policy because it is completely random and does not favor any particular estimator. In each of the $T$ inventory cycles, we simulate the number of customer arrivals $M_t$ from a Poisson distribution with expectation $M$. Customer choices are then simulated sequentially from the available assortments according to the MNL model. The assortment is updated dynamically in response to stockout events. 
Throughout this process, we apply each of the three estimators—our proposed counting estimator (CE), the standard MLE estimator, and the SGD variant of MLE—to the simulated data. To ensure reliability, we conduct 
$20$ independent replications of the entire experiment for each estimator.

\begin{table}[t]
	\caption{Average running time of the whole simulation (in seconds).}
	\small
	\label{table:estimator running time}
	\begin{tabularx}{\textwidth}{p{0.32\textwidth} X X X}
		\toprule
		\bd{Estimator} & CE & MLE & MLE-SGD \\
		\midrule
		\small \bd{Average Running Time ($M=5$)}& $0.084
		$ & $41636.197$ & $0.877$  \\
		\small \bd{Average Running Time ($M=10$)}& $0.077 $ & $64426.227$ & $1.302$  \\
		\bottomrule
	\end{tabularx}
\end{table}

We first report the average running time for each estimator across the $20$  replications in \Cref{table:estimator running time}, for both $M = 5$ and $M = 10$. According to this table, the standard MLE is very slow because it needs to repeatedly maximize log-likelihood. 
The SGD variant of MLE, termed ``MLE-SGD,'' enhances computational efficiency significantly, yet it still lags behind our proposed counting estimator (CE). This comparison underscores the computational advantage of our proposed method.

We further evaluate the estimation errors of different estimators and report the results in 
\Cref{fig:estimator mle}. 
The estimation error is measured as the Euclidean distance between the estimated parameter values and the actual parameter values. Figure~\ref{fig:estimator mle} depicts the average estimation errors across the $20$ replications for the three estimators at varying time points. It is evident that the estimation errors for all methods decrease as the sample size grows. Notably, our proposed CE and standard MLE achieve comparably low errors, despite their huge difference in computational efficiency.
In contrast, the MLE-SGD variant, although computationally efficient, yields substantially higher and more volatile estimation errors.
Therefore, our proposed CE stands out for its computational efficiency and accuracy, whereas MLE and MLE-SGD fall short in either computational speed or estimation accuracy.

\begin{figure}[t]
	\centering
	\subfigure[$N = 5$, $\C = 5$, $K = 5$, $M = 5$]
	{
		\begin{minipage}[b]{.48\linewidth}
			\centering
			\includegraphics[scale=0.4]{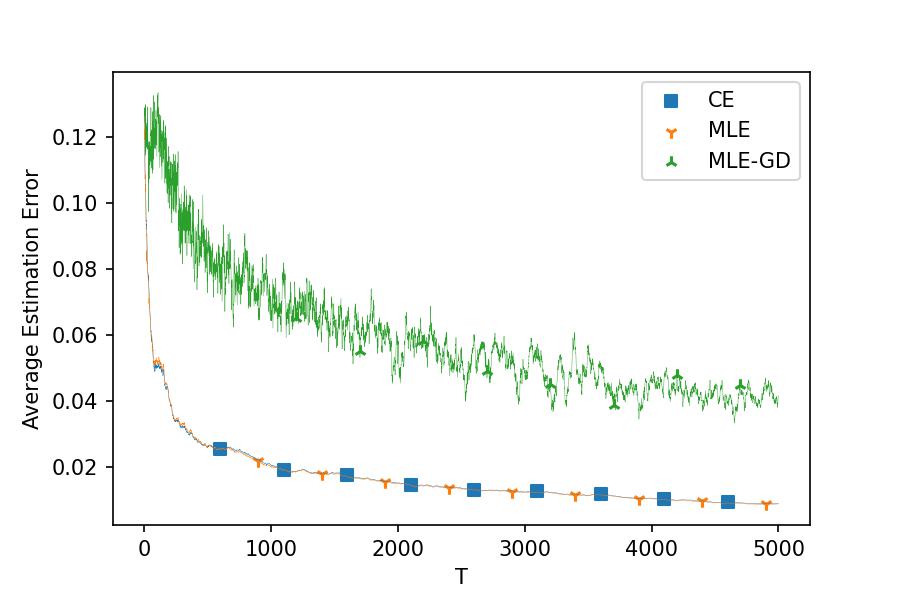}
		\end{minipage}
	}
	\subfigure[$N = 5$, $\C = 5$, $K = 5$, $M = 10$]
	{
		\begin{minipage}[b]{.48\linewidth}
			\centering
			\includegraphics[scale=0.4]{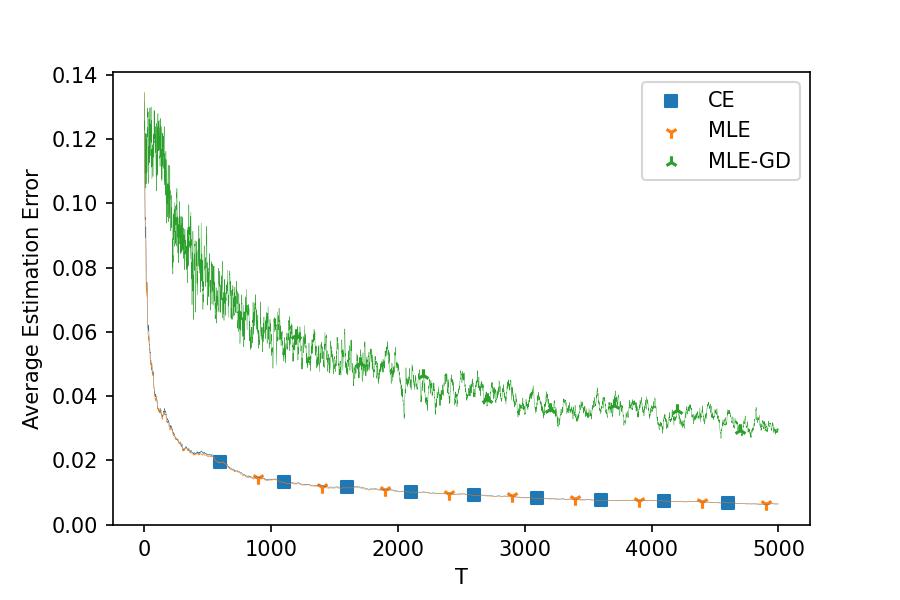}
		\end{minipage}
	}
	\caption{The trajectories of average estimation errors of CE, MLE and the SGD variant of MLE.}
\label{fig:estimator mle}
\end{figure}

Finally, we compare the confidence bounds derived from CE and MLE. The CE confidence bounds are calculated according to \Cref{equ-define-v-LCB}, while the MLE confidence bounds are determined by \Cref{eq: CI-v-MLE}, with $\kappa = \frac{v_{\min}}{e\left(1 + Kv_{\max}\right)^2}$ as specified in \Cref{lemma:theta convergence}. It is important to note that the MLE confidence bounds  require knowing $v_{\min}$ while our CE confidence bounds do not. 
We evaluate the confidence tightness by measuring the 1-norm distance of the confidence upper bounds $\bv^{\UCB}_{t}$ to the truth, namely $\norm{\min\{\bv_{\max},\ \bv^{\UCB}_{t} \} - \bv}{1}$, for both CE and MLE. 
We set a high truncation threshold $v_{\max} = 30$ to more clearly illustrate the behavior of the confidence bounds.  
\Cref{fig:mle_widthcompare} summarizes the average confidence width at various time points. The results indicate that the confidence bounds produced by our CE are significantly narrower than those of the MLE and converge more quickly to the true parameters. This suggests that while both sets of confidence bounds are theoretically valid, the MLE confidence bounds with the  $\kappa$ choice specified in \Cref{lemma:theta convergence} tend to be excessively wide, potentially leading to higher exploration costs in online decision-making processes.

In summary, the numerical experiments presented here  demonstrate the superiority of our proposed CE over the MLE in terms of computational efficiency, estimation accuracy, and the tightness of confidence bounds. These findings, reinforced by our discussion in Section \Cref{subsec:mle},  establish the advantage of our proposed CE method in addressing the learning challenges of our problem.

\begin{figure}[t]
\centering
\subfigure[$N = 5$, $\C = 5$, $K = 5$, $M = 5$]
{
\begin{minipage}[b]{.48\linewidth}
	\centering
	\includegraphics[scale=0.4]{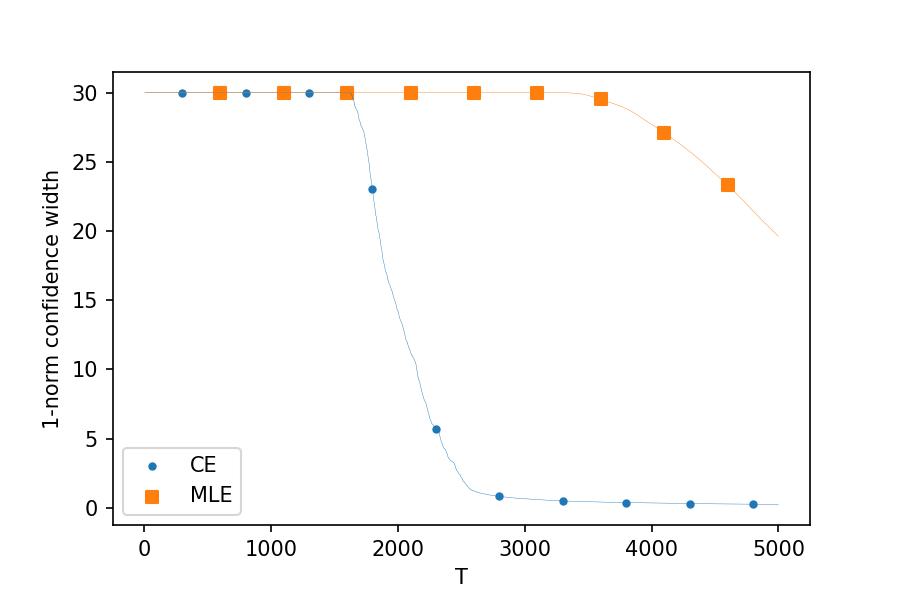}
\end{minipage}
}
\subfigure[$N = 5$, $\C = 5$, $K = 5$, $M = 10$]
{
\begin{minipage}[b]{.48\linewidth}
	\centering
	\includegraphics[scale=0.4]{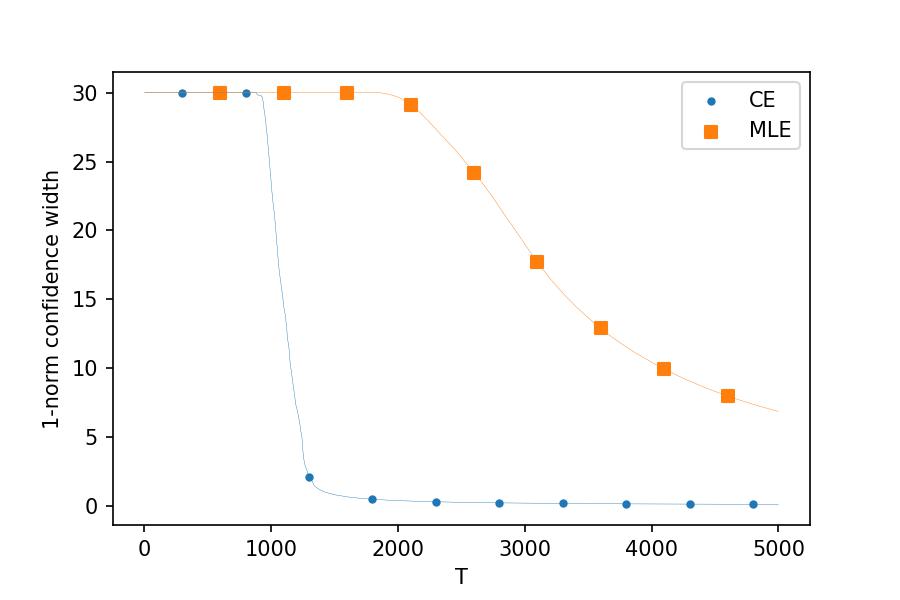}
\end{minipage}
}
\caption{Average distance of upper confidence bounds to truth for CE  and MLE.}
\label{fig:mle_widthcompare}
\end{figure}

}

\section{A Reduction Approach to Deriving Regret Lower Bound}\label{sec: reduction}
In this section, we outline the proof of the regret lower bound in \Cref{thm-lower-bound}. 
This proof is based on a reduction of the MNL-bandit problem in \cite{agrawal2019mnl} to our problem. 

\subsubsection*{MNL-bandit Problem.} We first review the MNL-bandit problem of online assortment optimization studied in \cite{agrawal2019mnl}. 
Consider an MNL-bandit problem with a total horizon of $\check T$ and $N$ products, where the unit profit vector is $\bre = (r_1, \dots, r_N)$. The $\check T$ customers arrive sequentially over time and, for each customer,
the retailer needs to choose an assortment $\check\s$ out of all feasible ones in a set $\setS$. 
In particular, \cite{agrawal2019mnl} impose totally unimodular constraints on $\setS$ and highlight the maximum cardinality constraint as a canonical example. Given the available assortment, each customer makes a choice according to the same MNL choice model described in \Cref{equ-define-MNL-choice-probability}. Again, the MNL attraction parameters $\bv = (v_1, \dots, v_N)$ are unknown \textit{a priori}. 
We let $\check{R}(\check\s; \bv, \bre)$ denote the expected profit when offering the assortment $\check\s$ to a customer given the true attraction parameters $\bv$ and the unit profits $\bre$, and $\check \s^* \in \argmaxx_{\check\s\in \setS} \check R(\check\s; \bv, \bre)$ denote a clairvoyant optimal assortment. 
Consider an MNL-bandit policy $\pi_{\MNL}$  that offers assortments $\check\s^{\pi}_{\check t}$ to the $\check t$-th arriving customer for $\check t \in [\check T]$. The regret of policy $\pi_{\MNL}$ is defined as 
\begin{equation*}
\check{\Reg}(\pi_{\MNL}; \bv, \bre, \check T) \coloneqq \mathbb{E} \left[ \sum_{\check t  = 1}^{\check T}\left(\check{R}(\check\s^*;\bv, \bre) - \check{R}(\check\s^{\pi}_{\check t};\bv, \bre) \right) \right] \ .
\end{equation*}
The goal of the MNL-bandit problem is to find a policy that achieves as small regret as possible. Apparently, the MNL-bandit problem is a special case of our problem with $T = \check T$ inventory cycles and only one customer arrival in each cycle (\ie, $M_1 = \cdots = M_{T} = 1$). 

\cite{agrawal2019mnl} designs an MNL-bandit algorithm that sequentially learns the unknown attraction parameters while maximizing profit.  To establish the (near) optimality of their algorithm, they derive a worst-case lower bound for the MNL-bandit problem when the maximum assortment size is constrained by $K \le N$.  This lower bound is refined by \cite{chen2018note} when $K \le N/4$. We summarize their lower bounds in the following lemma. 

\begin{lemma}\label{lemma: MNL-bandit-lb}
For any admissible policy $\pi_{\mathrm{MNL}}$, there exists an instance of the MNL-bandit problem  with $N$ products, $\check T$ inventory cycles satisfying $\check T \ge N$, maximum assortment size $K$ satisfying $K \le N$, such that the total regret of the policy $\pi_{\mathrm{MNL}}$ on this instance satisfies 
\begin{align*}
\check\Reg(\pi_{\MNL} ; T) = 
\begin{cases}
\Omega\left(\sqrt{N\check T}\right) & \text{ if } K \le N/4  \\
\Omega\left(\sqrt{N\check T/K}\right) & \text{otherwise}
\end{cases}
\quad .
\end{align*}
\end{lemma}

We remark that \Cref{lemma: MNL-bandit-lb} slightly differs from the statement of Theorem 2 in \cite{agrawal2019mnl}. 
\cite{agrawal2019mnl} shows that the regret of any admissible policy averaged over a mixture of a class of MNL-bandit instances has to be at least $\Omega\left(\sqrt{N\check T/K}\right)$. 
Their statement directly implies our statement in \Cref{lemma: MNL-bandit-lb} since the average regret over a mixture of a  class of instances is always a lower bound for the worst-case regret over the same class of instances. 
Thus, their average regret lower bound is also a worst-case lower bound, leading to \Cref{lemma: MNL-bandit-lb}. 

To derive a regret lower bound for our online joint assortment-inventory optimization problem, it suffices to focus on the special case where a deterministic number of customers arrive in each inventory cycle, \ie, $M_1 = \cdots = M_T = M$. 
As discussed above, the special case of our problem with $M = 1$ is exactly an MNL-bandit problem. Thus, a regret lower bound for this special case directly follows from \Cref{lemma: MNL-bandit-lb} with $\check T = T$. 

\subsubsection*{A Reduction Approach. } 
Now we use a reduction approach to derive a more general lower bound for  our problem. 
Specifically, for any MNL-bandit instance $\mathcal{I}_{\text{MNL}}$ with a total horizon $\check T = MT$, we can construct an instance $\mathcal{I}_{\text{MNLI}}$ of our problem with $T$ inventory cycles and $M$ customer arrivals in each cycle (the additional ``I'' in the subscript of $\mathcal{I}_{\text{MNLI}}$ stands for ``inventory'').
Moreover, for any online algorithm $\mathcal{A}_{\MNLI}$ for the instance $\mathcal{I}_{\text{MNLI}}$, we can construct an MNL-bandit algorithm $\mathcal{A}_{\MNL}$ for the instance $\mathcal{I}_{\text{MNL}}$, such that the decisions of $\mathcal{A}_{\MNL}$ are completely determined by the decisions of $\mathcal{A}_{\MNLI}$ on $\inst_{\MNLI}$. Importantly, with our construction, the total regret of $\mathcal{A}_{\MNLI}$ on the instance $\mathcal{I}_{\MNLI}$
is equal to the total regret of $\mathcal{A}_{\MNL}$ on the instance $\mathcal{I}_{\MNL}$.
In this way, we construct a reduction of an MNL-bandit problem to our problem. Accordingly, the regret lower bound in \Cref{lemma: MNL-bandit-lb} with $\check T = MT$ also provides a regret lower bound for our problem with $T$ inventory cycles. 

We now describe the construction of the reduction. 
For an MNL-bandit instance $\mathcal{I}_{\text{MNL}}$ with a total horizon $\check T = MT$, we can divide its arriving customers into $T$ consecutive batches of $M$ customers, and map them to the customers arriving sequentially in an instance $\mathcal{I}_{\MNLI}$. Specifically, the $m$-th customer in the $t$-th inventory cycle of $\mathcal{I}_{\MNLI}$ corresponds to the $[\check t = M(t-1) + m]$-th customer of $\mathcal{I}_{\text{MNL}}$, and these two customers make the same choice. Moreover, these two instances also share the same values for relevant parameters including $\bv, \bre, K, N$. To fully construct the $\mathcal{I}_{\MNLI}$ instance, we further specify the inventory capacity parameters to be any such that $\min\{c_1, \dots, c_N\} \ge M$ and $\bar{c} \ge KM$. Under any of such specifications, the clairvoyant optimal MNLI decision will not lead to any stockout event. Consequently, its expected profit can be shown to be equal to the clairvoyant optimal expected profit for the $\mathcal{I}_{\text{MNL}}$ instance. Algorithm~\ref{alg-reduction} summarizes how to map a given algorithm $\mathcal{A}_{\text{MNLI}}$ for the $\mathcal{I}_{\text{MNLI}}$ instance to an MNL-bandit algorithm $\mathcal{A}_{\MNL}$ for the  $\mathcal{I}_{\text{MNL}}$ instance  and how to map a sample path of the $\mathcal{I}_{\text{MNL}}$ instance to a sample path of a $\mathcal{I}_{\text{MNLI}}$ instance.

\begin{figure}
\centering 
\includegraphics[width=0.9\textwidth]{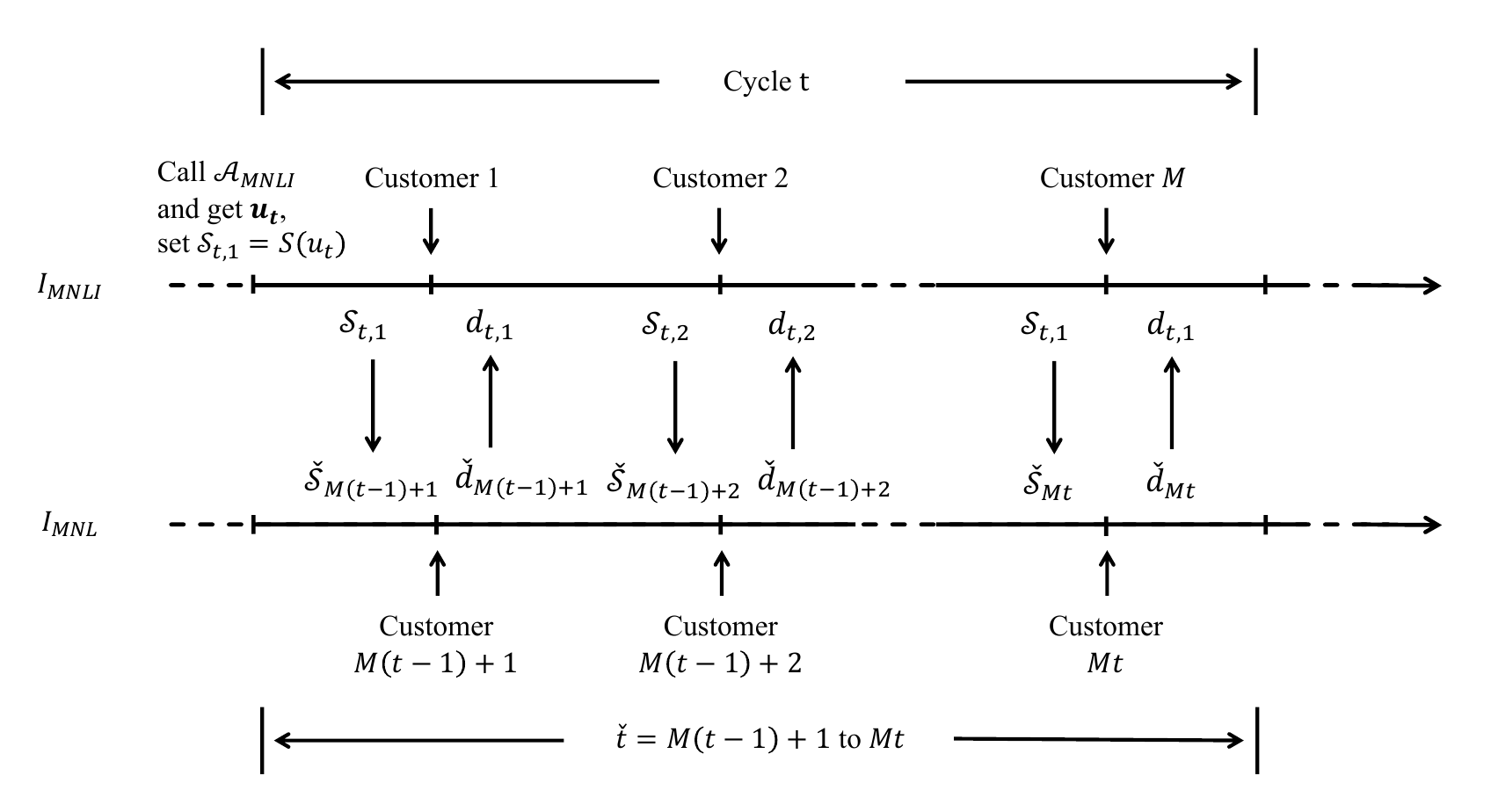}
\caption{An illustration of the reduction of an MNL-bandit instance $\mathcal{I}_{\text{MNL}}$ to an instance $\mathcal{I}_{\text{MNLI}}$ of the online joint assortment-inventory optimization problem. The arrows indicate the relations between a pair of variables. For example, the downward arrow from $\s_{t, 1}$ to $\check\s_{M(t-1)+1}$ means that $\check\s_{M(t-1)+1}$ is set  as $\s_{t, 1}$, while the upward arrow from $\check d_{M(t-1)+1}$ to $d_{t, 1}$ means that $d_{t, 1}$ is set as $\check d_{M(t-1) + 1}$.}\label{fig: reduction}
\end{figure}

To explain Algorithm~\ref{alg-reduction}, consider the beginning of cycle $t$ in $\mathcal{I}_{\text{MNLI}}$, or equivalently, the time right before the $t$-th batch of customers starts arriving in $\mathcal{I}_{\text{MNL}}$. We start with calling algorithm $\mathcal{A}_{\text{MNLI}}$ to determine an inventory decision $\U_t$. We offer the corresponding assortment $S(\U_t)$ to the first customer arriving in the $t$-th batch in $\mathcal{I}_{\text{MNL}}$ (or equivalently, the $[M(t-1)+1]$-th customer in $\mathcal{I}_{\text{MNL}}$) and record this customer's choice $\check d_{M(t-1)+1}$. Then the choice $\check d_{M(t-1)+1}$ is sent as a feedback to $\mathcal{A}_{\text{MNLI}}$, and accordingly, it determines the choice $d_{t, 1}$ of the $1$-st customer arriving in the $t$-th cycle of $\mathcal{I}_{\text{MNLI}}$. After the choice $d_{t, 1}$, the remaining inventory gives an updated assortment $\s_{t, 2}$.  We then offer this assortment to the $[M(t-1)+2]$-th customer in $\mathcal{I}_{\text{MNL}}$, record her choice $\check d_{M(t-1)+2}$, and send $\check d_{M(t-1)+2}$ as a feedback to $\mathcal{A}_{\text{MNLI}}$. 
This new feedback determines the choice $d_{t, 2}$ of 
the $2$-nd customer in the $t$-th cycle of $\mathcal{I}_{\text{MNLI}}$, which in turn updates the available assortment. 
This process repeats over and over again until the end of the horizons of both $\mathcal{I}_{\text{MNLI}}$ and $\mathcal{I}_{\text{MNL}}$.
In this process, the assortment offered to each customer in $\mathcal{I}_{\text{MNL}}$ is  determined by the available assortment to the corresponding customer in $\mathcal{I}_{\text{MNLI}}$ when running algorithm $\mathcal{A}_{\text{MNLI}}$. 
Moreover, the choice of each customer in $\mathcal{I}_{\text{MNLI}}$ is determined by the choice of the corresponding customer in $\mathcal{I}_{\text{MNL}}$.
This process is illustrated in \Cref{fig: reduction}.

From the construction described above, we can prove that the regrets of the two algorithms $\mathcal{A}_{\text{MNL}}$ and $\mathcal{A}_{\text{MNLI}}$ on their respective instances are identical. 
Our proof is based on the observations that the total profits achieved by the two algorithms on the two instances are identical according to construction in Algorithm~\ref{alg-reduction}, and that the clairvoyant optimal expected profits for the two instances are also identical per our discussions above. 

\begin{restatable}{lemma}{sameregret}\label{lemma-same-regret}
For an arbitrary algorithm  $\alg_{\MNLI}$ for the online joint assortment-inventory optimization problem, let $\alg_{\MNL}$ denote the algorithm constructed according to Algorithm~\ref{alg-reduction}. Then for any MNL-bandit instance $\inst_{\MNL}$  and the corresponding  online joint assortment-inventory optimization instance $\inst_{\MNLI}$,
the regret of $\alg_{\MNLI}$ on $\inst_{\MNLI}$ with $T$ inventory cycles and the regret of $\alg_{\MNL}$ on $\inst_{\MNL}$ with horizon $\check T = MT$ are identical, that is,  
\begin{equation*}
\Reg(\alg_{\MNLI} ; T) = \check{\Reg}(\alg_{\MNL}; \check{T}).
\end{equation*} 
\end{restatable}

Based on \Cref{lemma-same-regret}, we can readily show that the lower bound in \Cref{lemma: MNL-bandit-lb} for MNL-bandit problems with horizon $\check T = MT$ is also a lower bound for our problem with $T$ inventory cycles, which proves our lower bound in \Cref{thm-lower-bound}.

\begin{algorithm}[h]\label{alg-reduction}
\DontPrintSemicolon
\small
\KwInit{ $\check t = 1$, $t = 1$, m = 1 \;}
Call $\alg_{\MNLI}$ and receive the initial inventory decision $\U_1$\;
Initialize assortment $\check\s_1 = S(\U_1)$\;
\While{$\check t < \mathbf{\check T}$}
{
\While{$m \le M$}
{
Provide assortment $\check\s_{\check{t}}$ to the $\check{t}$-th customer and record her choice $\check d_{\check t}$ \;
Send \bd{Feedback} to Algorithm \ref{alg-main} such that the $m$-th customer in the $t$-th inventory cycle  makes a purchase decision $d_{t, m} = \check d_{\check t}$ \;
$m = m + 1$,\ 
$\check{t} = \check t + 1$ \;
\If{$m \le M$}{
Call $\alg_{\MNLI}$ and receive the  assortment $\s_{t, m}$ available to the $m$-th customer in the $t$-th inventory cycle \;
Set assortment $\check\s_{\check t} = \s_{t, m}$\;}
}
$m = 1$,\ $t = t + 1$ \;
Call $\alg_{\MNLI}$ and receive the inventory decision $\U_t$ \;
Set assortment $\check\s_{\check t} = S(\U_t)$\;
}		
\caption{An MNL-bandit algorithm based on $\alg_{\MNLI}$}
\end{algorithm}

\section{Approximate Oracles} \label{sec:oracle-survey}
\subsection{Examples of Approximate Oracles}
{\mnote{blue}{R2Q3\\R2Q4}{5mm}In this subsection, we survey the existing examples of approximate oracles for the   single-cycle static joint assortment and inventory optimization problems. We summarize each algorithm and also outline their computational complexities and theoretical guarantees. 
}

\edit{Leveraging a fluid approximation for the static optimization problems, 
\cite{liang2021assortment} propose a linear programming (LP)-based algorithm to compute the inventory decisions, and show that the algorithm is asymptotically optimal when customer arrivals follow a Poisson process and the expected number of customers is large. In particular, the suboptimality gap of their approach in terms of the expected profit can be bounded by $O(\sqrt{NM\log(NM)})$.
\cite{aouad2018greedy} propose a greedy-like algorithm %
for a broad class of customer arrival processes. The algorithm first categorizes the products into multiple classes, according to a price threshold and the products' attraction parameters, and then employs distinct greedy procedures for different classes to construct the corresponding inventory decisions. Finally, it uses the subadditivity of the expected revenue function to combine the solutions obtained for each product class. Their algorithm achieves a $(0.139-\epsilon)$-approximation with a probability at least $(1-\delta)$, for any $\epsilon\in(0, 1/4)$ and $\delta\in(0,1)$. Under the assumption that the expected revenue function can be efficiently evaluated, the running time of their algorithm is polynomial in the input size, $N^{1 / \epsilon}$, and $1 / \delta$. %
\cite{aouad2022stability} devise a polynomial-time approximation scheme (PTAS) for arbitrarily distributed customer arrivals, such that for any $\epsilon \in (0, 1)$ and $\delta\in(0,1)$, their polynomial-time approximation algorithm achieves a $(1-\epsilon)$-approximation with a probability of at least $(1-\delta)$. The algorithm constructs near-optimal inventory vectors by combining together many subvectors, each obtained by a highly non-trivial enumeration method. The running time of the algorithm is $O\left(\operatorname{poly}\left(N, \frac{1}{\epsilon}, \frac{1}{\delta}\right) \cdot \bar{c}^{\tilde{O}\left(\frac{1}{\epsilon^3} \log \frac{v_{\max}}{v_{\min}}\right)}\right)$, where $\bar{c}$ is the inventory capacity. 
The algorithms in \cite{aouad2018greedy} and \cite{aouad2022stability} rely on Monte Carlo sampling and dynamic programming to estimate the expected revenue of a given assortment and inventory plan. Recently, \cite{sun2024unifiedalgorithmicframeworkdynamic} develop algorithms with an improved approximation guarantee and running time. They use a two-step framework, which first finds an approximate inventory solution to maximize the fluid approximation of the expected profit and then utilizes a rounding technique to transform the approximate inventory solution into a feasible decision. The algorithm can achieve $(0.194-\epsilon)$-approximation with a running time of $O\left(N \log _{1+\epsilon}^2\left(\frac{K}{\epsilon}\right)\right)$, where $K$ is the cardinality constraint on the number of assorted products. 
}

\edit{We remark that the  computational bottleneck of our algorithm mainly lies in the static optimization oracle. The parameter estimation and  confidence bound construction in our algorithm involve only simple counting and averaging, whose complexity  grows linearly with the number of products $N$ and total number of inventory cycles $T$. The choice of different  oracles  described above  would have a major impact on the computational complexity of our algorithm. Moreover, the computational complexity of our algorithm also depends on its frequency of calling the static optimization oracle. For example, our main Algorithm~\ref{alg-main} requires calling the oracle at the end of every cycle, hence resulting in $T$ calls of the static optimization oracle. In contrast, the ``low switching'' Algorithm~\ref{alg:carry over} in \Cref{sec:carry over} only requires $O(\log T)$ calls of the static optimization oracle in expectation, so it needs much less computation. }

{ 
\subsection{The Approximate Oracle Adopted in Numerical Studies}
In our large-scale numerical studies in \Cref{sec: numerical-sushi} and \Cref{sec: numerical-randomized}, we use the approximate oracle of \cite{liang2021assortment} to solve single-cycle static optimization problems. 
\cite{liang2021assortment} consider no hard constraints on the inventory and assortment, but instead account for ordering costs. The ordering costs act as ``soft constraints'' on the inventory, preventing excessive inventory ordering. To employ this oracle in our large-scale numerical studies, we relax the inventory and assortment capacity constraints (which is equivalent to considering sufficiently large capacities), and incorporate an ordering cost. Specifically, the approximate oracle for the static optimization problem solves the following linear programming problem $R^{\text{LP}}(\bv,\bre)$
\begin{align}
\R^{\text{LP}}(\bv,\bre) \coloneqq \max_{\U, u_0} & \ \sum_{i\in[N]}u_i (r_i-o_i) \label{eq:app LP}\\
\text{s.t.}\ &\ 0\le u_i\le v_i \cdot u_0 \quad \forall i \in [N] \notag \\
&\ \sum_{i\in[N]\cup\{0\}} u_i = M ,\notag
\end{align}
where $r_i$ and $o_i$ are the unit revenue and unit ordering cost of product $i$, respectively, and $u_0$ is an auxiliary variable representing the number of no-purchases. 
This approximate oracle solves a linear programming problem to get an inventory decision (and its associated assortment), which is computationally efficient. According to \cite{liang2021assortment}, the oracle provides an asymptotic optimality guarantee to the single-cycle static optimization problem when the number of customer arrivals during an inventory cycle approaches infinity. To adopt the oracle in our online optimization problem, in each inventory cycle $t$, our proposed exploration-exploitation algorithm first solves $\R^{\text{LP}}(\bvUCB{t}, \hat \bre_{t})$ to get an optimal solution $\hat \U_t^{\text{LP}}$, and then implements $\floor{\hat \U_t^{\text{LP}}}$ as a feasible inventory decision for cycle $t$. In contrast, the benchmark ``$\bvUCB{}$-only'' algorithm solves $\R^{\text{LP}}(\bvUCB{t}, \bre)$ to get an optimal solution $\tilde \U_t^{\text{LP}}$, and implements the feasible inventory decision $\floor{\tilde \U_t^{\text{LP}}}$.

}

\section{General Ordering Costs and Salvage Values}\label{sec-relax-assumption}
In the main text, we assume in \Cref{assum-order-salvage} that the ordering cost and salvage value of each product are identical. 
Under this simplified assumption, the leftovers at the end of each cycle do not contribute to any profit or cost, and thus the expected profit of each inventory cycle takes a particularly simple form. 
In this section, we relax \Cref{assum-order-salvage} and consider the more general setting where the ordering costs and salvage values may be different.
We extend our algorithm to this setting and derive the regret bounds accordingly.

\subsubsection*{Problem Reformulation.} First, we introduce a few more notations, and rewrite the expected profit in each inventory cycle. Let $o_i'$, $s_i'$ and $r_i'$ denote the ordering cost, the salvage value and the unit selling price of product $i\in[N]$, respectively. To rule out trivial cases, we assume that $s_i' \le o_i' \le r_i'$ for all product $i\in [N]$. Recall that $X_i^{\U, \bv}$ stands for the demand for product $i$ in a single inventory cycle when the inventory decision is $\U$ and the vector of attraction parameters is $\bv$, and by definition we have $X_i^{\U, \bv} \le u_i$ for all $i \in [N]$. 
In the current general setting, the profit resulting from selling product $i$ in a single inventory cycle is 
\begin{equation*}
- o_i'\cdot u_i + r_i' \cdot X^{\U, \bv}_i + s_i'\cdot \max\{u_i - X^{\U, \bv}_i, 0\} = -(o_i' - s_i')\cdot u_i + (r_i' - s_i')\cdot X^{\U, \bv}_i\ .
\end{equation*}
Accordingly, the expected total profit of inventory decision $\U$ under parameters $\bv$, $\bre'$, $\bo'$, and $\bs'$ can be written as 
\begin{equation*}
\profit'(\U; \bv, \bre', \bo', \bs') \coloneqq \Eb{ \sum_{i\in [N]} \left((r'_i - s_i')\cdot X_i^{\U, \bv} -  (o_i' - s_i')\cdot u_i \right) } .
\end{equation*}

Let $\U^*$ denote the optimal inventory decision, that is, $\U^* \coloneqq \argmaxx_{\U \in \setU} \profit'(\U; \bv, \bre', \bo', \bs')$. 
The total regret of  a policy $\pi$ that prescribes decisions $\U_1, \dots, \U_T$ in the $T$ inventory cycles is given by  
\begin{equation*}
\Reg'(\pi; \bv, \bre', T) = \Epi{\sum_{t = 1}^T \left(\profit'(\U^*; \bv, \bre', \bo', \bs') - \profit'(\U_t; \bv, \bre', \bo', \bs')\right)}.
\end{equation*}

It is convenient to further normalize the profit function $\profit'$ and the regret function $\Reg'$. With a slight abuse of notation, define $\re_i \coloneqq (r'_i - s_i') / \const_4$ with $\const_4 \coloneqq \max_{j\in[N]}\{r'_j - s_j'\}$ as the adjusted unit-profit of product $i$, such that $r_{\max} \coloneqq \max_{i\in[N]} \re_i = 1$. Similarly, define $o_i \coloneqq (o'_i - s_i') / \const_4$. Then, we have the following normalized expected profit function $\profit(\U; \bv, \bre, \bo)$: 
\begin{equation*}
\profit(\U; \bv, \bre, \bo) \coloneqq  \Eb{\sum_{i\in [N]} \left\{r_i\cdot  X_i^{\U, \bv} -  o_i \cdot u_i \right\}}=\sum_{i\in [N]} \left\{r_i\cdot  \Eb{X_i^{\U, \bv}}-  o_i\cdot u_i \right\} \ . 
\end{equation*}

Apparently, the original profit function $\profit'$ and normalized profit function $\profit$ differ only in a constant factor $\const_4$. Thus $u^*$ also optimizes the normalized profit function $\profit$. 
We can accordingly define the normalized regret of a policy $\pi$ as follows:  
\begin{equation*}
\Reg(\pi; \bv, \bre, T) = 	\Epi{\sum_{t = 1}^T \left\{\profit(\U^*; \bv, \bre, \bo) - \profit(\U_t; \bv, \bre, \bo) \right\}} = \Reg'(\pi; T)/\const_4\ .
\end{equation*}
We note that the normalized profit function $\profit$ is very similar to the profit function $R$ defined under \Cref{assum-order-salvage} (see \Cref{subsec-clairvoyant-problem} \Cref{equ-onecycle-revenue}). 
In particular, the adjusted unit profits $\bre$ play the same role as the raw unit profits in \Cref{equ-onecycle-revenue}. 
The only difference is that the normalized profit $\Pi$ involves an additional term $-\sum_{i\in[N]}o_iu_i$.

\subsubsection*{Extending Our Algorithm.}
We can straightforwardly tailor our proposed algorithm to the normalized profit function $\Pi$. 
In particular, we define the upper bound on the adjusted unit profits: 
\begin{equation*}
\hat{\re}_{i, t} \coloneqq \min\{1, r_i + \drecv_{i, t} \}, \text{ where } \drecv_{i, t} = v_{i, t}^{\UCB}/ v_{i, t}^{\LCB} - 1, \text{ for all } i \in [N].
\end{equation*}
Again, in each non-exploratory cycle $t \in [T]$, we compute the attraction parameter upper bounds $\bv_{t}^{\UCB} = (v_{1, t}^{\UCB}, \dots, v_{N, t}^{\UCB})$ and unit profit upper bounds  $\hat\bre_{t} = (\hat r_{1, t}, \dots, \hat r_{N, t})$, and then determine the inventory decision according to
\begin{equation}\label{equ-salvagecost-opt-oracle}
\U_t = \argmaxx_{\U \in \setU} 	\profit(\U; \bvUCB{t},\hat\bre_{t}, \bo).
\end{equation}
Therefore, we only need to slightly change the definition of unit profits and the expected profit function to extend our algorithm to the more general setting in this section. 

It is noteworthy that the objective of the single-cycle optimization problem in \Cref{equ-salvagecost-opt-oracle} involves an additional term $-\sum_{i\in[N]}o_iu_i$. Thus, we need an optimization oracle that can handle this additional term. 
For example, the approximate oracles in \cite{liang2021assortment,aouad2018greedy} are applicable. 
In particular, the linear programming based approach in \cite{liang2021assortment} can accommodate an additional holding cost at the end of an inventory cycle. 
We can simply view $o_i$ as the holding cost of product $i$, and apply the approach in \cite{liang2021assortment} to approximately solve \eqref{equ-salvagecost-opt-oracle}. 
\cite{aouad2018greedy} also remark that their approach can be extended to  the setting with  product-specific per-unit costs. Therefore, their algorithm can also be applied to approximately solve problem \eqref{equ-salvagecost-opt-oracle}. %

\subsubsection*{Regret Analysis.}
Since the modified algorithm in this section is nearly identical to Algorithm~\ref{alg-main}, its regret can also be bounded analogously. 
In the following lemma, we first show that the objective function in \Cref{equ-salvagecost-opt-oracle} also provides a valid optimistic estimate for the true normalized profit function. This extends the counterpart \Cref{lemma-policyUCB}.

\begin{restatable}{lemma}{ucbpolicyos}\label{lemma-ucb-os}
For any cycle $t$, if $\bv = (v_i)_{i \in [N]}$ and $\bv^{\LCB}_{t} = (v_{i, t}^{\LCB})_{i \in [N]}$, $\bv^{\UCB}_{t} = (v_{i, t}^{\UCB})_{i \in [N]}$ satisfy $\vLCB{i, t} \le v_i \le \vUCB{i, t}$ for all $i \in [N]$, then
\begin{equation*}
\profit(\U; \bvUCB{t}, \hat \bre_{t}, \bo) \ge \profit(\U; \bv, \bre, \bo), ~~ \textrm{ for any } \U \in \setU\ .
\end{equation*}
\end{restatable}

In a similar vein as the results in Section~\ref{subsec-upperbound}, the optimistic expected profit estimator $\profit(\U; \bvUCB{t}, \hat \bre_{t}, \bo)$ converges to the actual expected profit $\profit(\U; \bv, \bre, \bo)$ from above. The corresponding convergence rate is similar to that prescribed in \Cref{lemma-convergence-rate}. Based on these results, we have the following regret upper bound for the modified algorithm under the general setting in this section. 

\begin{restatable}{theorem}{regretupperboundos}\label{thm-regret-upperbound-os}
For the problem setting where ordering costs and salvage values may be different, let $\pi$ be the policy generated by the corresponding modified Algorithm with an exact optimization oracle to solve \eqref{equ-salvagecost-opt-oracle}. 
The worst-case total regret of policy $\pi$ satisfies
\begin{equation*}
\Reg'(\pi; T) = \const_4 \cdot \Reg(\pi; T) = O\left(\sqrt{NMT\log\left(\sqrt{NM}T\right)}\right).
\end{equation*}
\end{restatable}

Moreover, we note that the regret lower bound in \Cref{thm-lower-bound} is derived under the additional \Cref{assum-order-salvage}. 
Thus it also provides a regret lower bound for the more general setting considered in this section. 
Finally,  
in \Cref{thm-regret-upperbound-os}, we assume an exact optimization oracle to solve \eqref{equ-salvagecost-opt-oracle}. If instead we use an $\epsilon$-$\delta$ oracle as discussed in \Cref{sec-oracles}, then we can easily adapt the analysis in \Cref{sec-oracles} to upper bound the regret of the corresponding algorithm.

{ 
\section{Supplements for \Cref{sec:carry over} (Inventory Carryover)}
\label{sec:app carry over}
In \Cref{sec:carry over}, we study an extension where the leftover inventory at the end of one inventory cycle must be carried over to the next cycle.  This section supplements \Cref{sec:carry over} with additional details.

\subsection{Online Decision-Making Algorithm}
Algorithm~\ref{alg:carry over} outlines our approach for managing inventory carryover. As described in \Cref{sec:carry over}, this algorithm employs an epoch-based structure and adheres to a ``low switching'' principle. Instead of adjusting parameter estimates and inventory decisions in every cycle, it does so only at the start of each epoch. Each epoch consists of two stages of cycles denoted by $\Lambda_{\ell, 1}$ and $\Lambda_{\ell, 2}$, corresponding to different inventory planning  strategies.
During the first stage, $\Lambda_{\ell, 1}$, the algorithm orders up to the same inventory level $\U_{\ell}$
across all cycles  within this stage (\textit{line} \ref{line:stage1}). This stage persists for a number of cycles  
to be specified later. If any inventory remains after the end of the first stage, the algorithm transitions to a clearance stage, $\Lambda_{\ell, 2}$, where no new orders are placed, and the focus is on depleting the remaining inventory (\textit{line} \ref{line:stage2}). Once the second stage is complete, or if the first stage ends without leftover inventory, the algorithm progresses to the next epoch, $\ell + 1$. At this point, it updates parameter estimates and confidence bounds using all historical data and determines the new inventory decisions $\U_{\ell+1}$
(from lines \ref{line:move} to \ref{line:end}).

The methodology for constructing parameter estimates, confidence bounds, and inventory decisions is identical to that described in \Cref{sec-algorithm}. We define $L$ as the total number of epochs, $t_\ell$ as the cumulative number of cycles before epoch $\ell$ (with $t_{L} = T$), $k_{i, \ell}$ as the total purchases of product $i$ before epoch $\ell$, and $\mu_{i, k}$ denote as number of no-purchases between the $(k-1)$-th and $k$th purchases of product $i$ when it is available. We estimate the reciprocal attraction $\mu_i = 1/v_i$ for each product $i$ by averaging the summary statistics $\mu_{i, k}$'s before epoch $\ell$: 
\begin{equation}\label{eq: mu co}
\bar \recv_{i,\ell} \coloneqq \frac{1}{k_{i, \ell}}\sum_{k \in [k_{i, \ell}]} \recv_{i, \ell}\ . 
\end{equation} 

We then construct a confidence interval  for $\mu_i$ based on $\bar \recv_{i,\ell}$, for each product $i \in [N]$: 
\begin{align}
\label{eq:define confidence bound co}
\recv_{i, \ell}^{\LCB} \coloneqq \max\left\{\recv_{\min}, \ \bar \recv_{i,\ell} - \radius_{i,\ell}\right\}, ~~ \recv_{i,\ell}^{\UCB} \coloneqq \bar \recv_{i,\ell} +\radius_{i,\ell} \ ,
\end{align}
where $\radius_{i,\ell} = \max\{\sqrt{\bar \recv_{i,\ell}}, \bar \recv_{i,\ell}\}\sqrt{48Q_{i, \ell}} + 48Q_{i, \ell}$ for $Q_{i, \ell} = {\log\left(\sqrt{N}\sum_{s=1}^{t_{\ell}}M_s + 1\right)}/{k_{i,\ell}}$. 
The corresponding confidence interval for the attraction parameter $v_i$ is given by 
\begin{align}\label{eq:define v LCB co}
v_{i,\ell}^{\LCB} \coloneqq \frac{1}{\recv_{i, \ell}^{\UCB}}\ , ~~ v_{i,\ell}^{\UCB} \coloneqq \frac{1}{\recv_{i, \ell}^{\LCB}}\ . 
\end{align}
Accordingly, we can set the adjusted unit profits as follows:   
\begin{align}
\label{equ-define-rhat-carryover}
\blue{\hat r_{i, \ell}} \coloneqq &\ \begin{cases}
1, &\blue{\text{if }\ Q_{i, \ell} > 1/48},\\
\min\{1, r_i + \delta_{i, \ell} \}, &\ \text{otherwise},
\end{cases} \quad
\text{ where } \delta_{i, \ell} \coloneqq \ \frac{\recv_{i, \ell}^{\UCB}}{\recv_{i, \ell}^{\LCB}}  - 1 . 
\end{align}
The inventory decision for every cycle of the first stage in epoch $\ell$ is then given by 
\begin{equation}\label{equ-define-Uell}
\U_{\ell} \in \argmaxx_{\U \in \setU} \R(\U; \bv^{\UCB}_{\ell}, \hat \bre_{\ell}).
\end{equation}
For the cycles in the second stage of epoch 
$\ell$, the algorithm places no orders. Consequently, the starting inventory $\U_t$
for each cycle is simply the inventory remaining from the previous cycle. 

Finally, it remains to determine the length of the first stage of each epoch $\ell$ for $\ell \in [L]$. Our Algorithm~\ref{alg:carry over} sets the length of the first stage in the $\ell$-th epoch as $\min\{2^{\tau_\ell}, T - t_\ell\}$, where $\tau_{\ell}$ is the smallest integer $\tau$ such that $2^\tau$ is no less than the minimum   cumulative purchases for products that are assorted in epoch $\ell$ (line  
\ref{line:length}): 
\begin{align}\label{eq: tauell}
\tau_\ell = \min  \lbr{\tau: \min_{i\in S(\U_{\ell})}k_{i,\ell} \le 2^{\tau}, \tau = 1, 2, \dots}.
\end{align}

It is important to note that the total number of epochs, $L$, is not fixed and depends on the random lengths of both the first stages $\Lambda_{\ell, 1}$ and the second stages $\Lambda_{\ell, 2}$. Next, we show that, under an additional assumption about the distribution of customer arrivals $M_t$
over the period $t \in [T]$, we can establish an upper bound for the expected value of $L$.

\begin{restatable}{lemma}{lemmaboundL}
\label{lemma:bound L co}
Suppose that there exists a constant $p > 0$ such that $\prob{M_t \ge 1} \ge p$ for any $t\in[T]$. 
Then the  expected number of epochs that 
Algorithm~\ref{alg:carry over} spans over $T$ cycles has the following  upper  bound:  
\begin{align*}
\mathbb{E}\left[{\Abs{L}}\right] \le \frac{1 + Kv_{\max}}{pv_{\min}}\left(1 + 2\bar c\right) N\log T + 1 \ .
\end{align*}
\end{restatable}

Lemma~\ref{lemma:bound L co} verifies that the decision-making process in Algorithm~\ref{alg:carry over} exhibits a low frequency of switches, occurring no more than $O(\log T)$ times on average.

\begin{algorithm}[t] 	\label{alg:carry over}
\small
\SetAlgoLined
\DontPrintSemicolon
\KwInit{$\bvUCB{1} = (v_{\max}, \dots, v_{\max})$, $\hat\bre_{1} = (r_{\max}, \dots, r_{\max})$, $\kb = \{0,\cdots,0\}$, $\nbb = \{0,\cdots,0\}$, $\U_{1} \in \arg\max_{\U\in \setU}R(\U; \bvUCB{1}, \hat\bre_{1})$, $t = 1$, $\ell = 1$, $\iota = 1$, $\tau_1 = 0$}
\While{$t < T$}
{
\eIf{$\iota \le 2^{\tau_{\ell}}$}
{
Order up to $\U_\ell$,\ 
$\Lambda_{\ell,1} = \Lambda_{\ell,1}\cup \{t\}$ \label{line:stage1} \;
}
{
Order $\mathbf{0}$, \ $\Lambda_{\ell,2} = \Lambda_{\ell,2}\cup \{t\}$ \label{line:stage2}\;
}
Observe the choices  $\seq_t = (d_{t, m})_{m \in [M_t]}$ made by the $M_t$ customers in  cycle $t$\;
\For{$m\gets1$ \KwTo $M_t$} {%
\eIf{$d_{t, m} = 0$}
{\For{$i\in\s_{t,m}$}{$n_i = n_i + 1$}
} 
{ $i = d_{t, m}$,\ 
	$k_i = k_i + 1$,\ $\recv_{i, k_i} = n_i$,\ $n_i = 0$\;
}
}
\If{$\iota \ge 2^{\tau_{\ell}}$ and no leftover inventory \label{line:move}}{%
$\ell = \ell + 1$, $\iota = 0$\;
\For{$i\gets1$ \KwTo $N$}{%
	$k_{i,\ell} = k_i$\;
	Compute $\bar\recv_{i,\ell}$, $\recv_{i, \ell}^{\LCB}, \recv_{i, \ell}^{\UCB}, v_{i,\ell}^{\LCB}, v_{i,\ell}^{\UCB}$ per \Cref{eq: mu co,eq:define confidence bound co,eq:define v LCB co}, respectively\;
	Compute $\hat \re_{i, \ell}$ according to \Cref{equ-define-rhat-carryover}\;}%
Set $\bv_{\ell}^{\UCB} = (v_{1,\ell}^{\UCB}, \dots,  v_{N,\ell}^{\UCB})$ and $\hat\bre_\ell = (\hat r_{1, \ell}, \dots, \hat r_{N, \ell})$ %
\;
Compute $\U_{\ell} \in \arg\max_{\U\in \setU}R(\U; \bvUCB{\ell}, \hat\bre_{\ell})$\;
\label{line:end}
Find $\tau_{\ell}$ according to \Cref{eq: tauell}
\;\label{line:length}
}
$t = t + 1$, $\iota = \iota + 1$\;
}		
\caption{Exploration-Exploitation Algorithm for the Inventory Carryover Setting}
\end{algorithm}

\subsection{Regret Analysis}
The following algorithm upper bounds the regret of the decision policy given by Algorithm~\ref{alg:carry over}.
\begin{restatable}{theorem}{mainls}\label{thm:main co}
Suppose the assumption in \Cref{lemma:bound L co} holds. Then for any instance of the online joint assortment-inventory optimization problem with $N$ products, maximum assortment cardinality $K$, and expected number of customers $M$ per cycle, the worst-case total regret of the policy $\pi_{\text{ls}}$ generated by the online algorithm~\ref{alg:carry over} over $T$ inventory cycles satisfies
\begin{equation*}
\Reg(\pi_{\text{ls}} ; T) = O\left(KMN\sqrt{T\log \left(\sqrt{N}MT\right)}\log T\right) \ .
\end{equation*}
\end{restatable}

The proof for \Cref{thm:main co}
parallels the proof for \Cref{thm-main}. It involves high-probability good events analogous to those in \Cref{def-event}. In particular, we can define events $\Acal_\ell \coloneqq \bigcap_{i\in[N]} \lbr{\Acal_{i,\ell}^{1c}\bigcup \Acal_{i,\ell}^2}$, where 
\begin{align*}
\Acal_{i,\ell}^1\coloneqq & \lbr{Q_{i, \ell} < 1/48}, \\
\Acal_{i,\ell}^2\coloneqq & \lbr{\mu_{i,\ell}^{\LCB}\le \recv_{i} \le \mu_{i,\ell}^{\UCB} \text{ and } \mu_{i,\ell}^{\UCB} -  \mu_{i,\ell}^{\LCB} \le 12\sqrt{3} \max\{\sqrt{\recv_{i}}, \recv_{i}\}\sqrt{Q_{i, \ell}} + 192Q_{i, \ell}}.
\end{align*}
Here, the events $\Acal_{\ell}$ represents the scenario where either product 
$i$ is not adequately explored, or when it is, its confidence interval is well-behaved.

We can then decompose the total regret into three parts based on whether the good events hold and whether each cycle belongs to the first or second stage of an epoch. In particular, for every $\bv \in [v_{\min}, v_{\max}]^{[N]}$ and $\bre \in [0,1]^{[N]}$, 
\begin{align}
&\Reg(\pi_{\text{ls}} ; \bv, \bre, T)=
\ \mathbb{E}_{\pi_{\text{ls}}}\left[{\sum_{\ell \in [L]}\sum_{t\in\Lambda_{\ell}} \left( R(\U^*; \bv, \bre) - R(\U_{\ell}; \bv, \bre) \right)}\right]  \nonumber \\
= &\ 
\mathbb{E}_{\pi_{\text{ls}}}\left[\sum_{l\in [L]}\sum_{t\in\Lambda_{\ell,1}}   \left(R(\U^*; \bv, \bre) - R(\U_\ell; \bv, \bre)\right) \1{\Acal_{\ell}}\right] +\mathbb{E}_{\pi_{\text{ls}}}\left[\sum_{l\in [L]}\sum_{t\in\Lambda_{\ell,2}}   \left(R(\U^*; \bv, \bre) - R(\U_t; \bv, \bre)\right) \1{\Acal_{\ell}}\right]\nonumber\\ 
+& \mathbb{E}_{\pi_{\text{ls}}}\left[ {\sum_{\ell \in[L]}  \sum_{t\in \Lambda_{\ell}}
\left(R(\U^*; \bv, \bre) - R(\U_t; \bv, \bre)\right) \1{\Acal_{\ell}^c}}\right]. \label{eq:regret part co}
\end{align}

Upper bounding the first and third terms in Equation~\eqref{eq:regret part co} is similar to bounding the corresponding terms in Equation~\eqref{equ-reg-part} for Theorem~\ref{thm-main}. The first term is upper bounded by the cumulative parameter estimation errors when the good events hold, which is a straightforward extension of Lemma~\ref{lemma-convergence-rate}, and the upper bound on the expected number of epochs given in Lemma~\ref{lemma:bound L co}. We can show that this term is the dominant term in the regret bound of Theorem~\ref{thm:main co}.
The third term in Equation~\eqref{eq:regret part co}, which accounts for the failure of the good events, can be bounded by the low probability of such failures as shown in appendix. We can show that this term contributes to only a negligible logarithmic regret.

To bound the second term in \Cref{eq:regret part co}, which represents the regret incurred during the second stage of each epoch, we need to further characterize the length of these second stages of different epochs. 
In appendix we demonstrate that the expected length of the second stage for any epoch $\ell \in [L]$ is bounded by a constant, specifically $\Eb{\Abs{\Lambda_{\ell,2}}}\le \bar c \frac{1+v_{\min}}{M v_{\min}} + 1$. 
Given that the average number of epochs is $O(\log T)$, as established in Lemma~\ref{lemma:bound L co}, the second term in Equation~\eqref{eq:regret part co} contributes only a negligible logarithmic term to the overall regret. Combining this finding with the bounds for the first and third terms of Equation~\eqref{eq:regret part co}, we can conclude the proof of the desired regret bound as stated in Theorem~\ref{thm:main co}, by additionally noting that the bounds are independent of the values of $\bv$ and $\bre$.

\section{Supplements for \Cref{sec:mt} (Unknown Customer Arrival Distribution)}
\label{app sec:mt detail}

This section supplements \Cref{sec:mt} for the setting where the customer arrival distribution $\Mcal$ is an unknown distribution, which 
belongs to a  distribution class $\mathscr{M}$. As described in \Cref{sec:mt}, we need to additionally estimate the distribution $\Mcal$, and we assume that it has a finite variance upper bounded by $a_M^2$ for a known positive constant $a_M$. 

Algorithm~\ref{alg:main mt} presents our revised algorithm for this setting. This algorithm is almost  identical to Algorithm~\ref{alg-main}, except that it estimates the unknown distribution $\Mcal$ using some estimator $\hat \Mcal_t$ at the start of each cycle $t$ (\Cref{line:update-Mt}),  alongside the estimation of the MNL parameters. Accordingly, it determines the inventory decision for cycle $t$ by optimizing  the upper confidence bound of the expected profit evaluated at the estimated distribution (\Cref{alg-calculate-u-Mt}):
\begin{align}
\label{eq:ut mt}
\U_t\in \argmaxx_{\U\in\setU}R^{\hat\Mcal_t}(\U; \bvUCB{t}, \hat\bre).
\end{align}

Below we show that when the estimated distribution $\hat\Mcal_t$ stochastically dominates the true distribution $\Mcal$, the objective in \Cref{eq:ut mt} gives a valid upper confidence bound on the true expected profit.
\begin{restatable}{lemma}{coroMtUCB}
\label{coro:mt ucb regret}
For any cycle $t$, if $\hat\Mcal_t$ stochastically dominates $\Mcal$ and the event $\Acal_t$ in \Cref{def-event} holds, then for any $\U\in\setU$ we have
\begin{align*}
R^{\hat\Mcal_t}(\U; \bvUCB{t}, \hat\bre_t) \ge R^{\Mcal}(\U; \bv, \bre),
\end{align*}
and the one-cycle  expected regret of $\U_t$ in \Cref{eq:ut mt} can be upper bounded as 
\begin{align*}
R^{\Mcal}(\U^*; \bv, \bre) - R^{\Mcal}(\U_t; \bv, \bre) \le \Eover{\hat\Mcal_t}{M_t} - \Eover{\Mcal}{M_t}  &+ \sum_{i\in S(\U_t)} \left(2\delta_{i,t} + \delta_{i,t}^2\right)\Eb{X_{i,t}^{\U_t, \bv}}\1{Q_{i, t} < 1/48} \\
&+ \sum_{i\in S(\U_t)}V\Eb{X_{i,t}^{\U_t,\bv}}\1{Q_{i, t} \ge 1/48},
\end{align*}
where $\Eover{\hat\Mcal_t}{M_t}$ and $\Eover{\Mcal}{M_t}$ stand for the expected number of customer arrival in cycle $t$ under the distribution $\hat\Mcal_t$ and $\Mcal$ respectively and $V = v_{\max}/v_{\min}$.
\end{restatable} 

\Cref{coro:mt ucb regret} hinges on the good events  in \Cref{def-event}, which can be proven to hold with a high probability according to \Cref{lemma-event-prob}. When this condition holds and the distribution estimator $\hat\Mcal_t$ stochastically dominates $\Mcal$, the objective of \Cref{eq:ut mt} provides a valid upper confidence bound. The expected regret of the resulting decision can be upper bounded by both the MNL parameter estimation errors,  characterized by $\delta_{i, t}$ for $i\in[N]$, and the distribution estimation error, characterized by $\Eover{\hat\Mcal_t}{M} - \Eover{\Mcal}{M}$. Note that this one-cycle expected regret resembles the bound in \Cref{lemma-convergence-rate}  but incorporates the additional error term due to distribution estimation. 

To obtain a regret bound on our algorithm, we further require a stochastic dominance property of the distribution estimator and specify its error bound in the following assumption. 

\begin{assumption}\label{assump:Mt-error}
For any cycle $t \ge a_1\log T+1$, $\hat\Mcal_t$ stochastically dominates $\Mcal$ and $\Eover{\hat\Mcal_t}{M_t} - \Eover{\Mcal}{M_t} \le a_2\sqrt{\log (t-1)/(t-1)}$ with probability at least $1 - a_3/(t-1)$ for some global constants $a_1$, $a_2$ and $a_3$.
\end{assumption}

For distribution estimators that satisfy \Cref{assump:Mt-error}, we can combine \Cref{coro:mt ucb regret} with the proof of \Cref{thm-main} to derive a regret bound for our new algorithm that learns customer arrival distribution alongside with the attraction parameters. 

\begin{restatable}{theorem}{thmmainmt}\label{thm:upperbound mt}
For any instance of the online joint assortment-inventory optimization problem with $N$ products, maximum assortment cardinality $K$, and an unknown arrival distribution $\Mcal \in \mathscr{M}$, if the distribution estimators $\hat\Mcal_t$ for $t \in [T]$ satisfy \Cref{assump:Mt-error}, then the expected worst-case total regret of the policy $\pi_{m}$ generated by Algorithm~\ref{alg:main mt} can be upper bounded as
\begin{align*}
\Reg(\pi_{m};T) = O\left(\sqrt{MNT\log\left(\sqrt{MN}T + 1\right)}\right) \ .
\end{align*}    
\end{restatable}

The regret bound in \Cref{thm:upperbound mt} is comparable to the regret bound in \Cref{thm-main}. This shows that the additional estimation of customer arrival distribution has only minor impact on the decision regret, provided that it satisfies \Cref{assump:Mt-error}. Below we explain how to construct such estimators when $\mathscr{M}$ is the Poisson distribution class and the Negative Binomial distribution class, respectively.

\begin{algorithm}[!ht]
\label{alg:main mt}
\small
\DontPrintSemicolon
\KwInit{$t = 1$, $\vUCB{i,1} = v_{\max}$,\ $\hat\re_{i,1} = r_{\max}$,\ $n_i = 0$,\ $k_{i} = 0$,\ 
$\forall i\in [N]$\; 
$\bv_{1}^{\UCB} = (v_{1,1}^{\UCB}, \dots,  v_{N,1}^{\UCB})$ and $\hat\bre_1 = (\hat r_{1, 1}, \dots, \hat r_{N, 1})$, 
randomly initialize $\U_1 \in \setU$}
\While{$t < {T}$}
{
Order up to $\U_t$, record the number of customer arrivals $M_t$ in cycle $t$ and observe the choices $\seq_t = (d_{t, m})_{m \in [M_t]}$ made by $M_t$ customers \;
\For{$m\gets1$ \KwTo $M_t$} {
\eIf{$d_{t, m} = 0$}
{\For{$i\in\s_{t,m}$}{$n_i = n_i + 1$}
} 
{
	$k_i = k_i + 1$,\ 
	$\recv_{i, k_i} = n_i$,\ 
	$n_i = 0$\;
}
}
$t = t + 1$ \;
Update $\hat\Mcal_t$\; \label{line:update-Mt}
\For{$i\gets1$ \KwTo $N$}{
Update $k_{i,t} = k_i$\;
Compute $\bar \recv_{i,t}$ according to \Cref{eq: mu-est}\;
Compute $\recv_{i, t}^{\LCB}, \recv_{i, t}^{\UCB}, v_{i,t}^{\LCB}, v_{i,t}^{\UCB}$ per \Cref{eq:define-recv-LCB,equ-define-v-LCB}, respectively\;
Compute $\hat \re_{i, t}$ according to \Cref{equ-define-rhat}\;}
Set $\bv_{t}^{\UCB} = (v_{1,t}^{\UCB}, \dots,  v_{N,t}^{\UCB})$ and $\hat\bre_t = (\hat r_{1, t}, \dots, \hat r_{N, t})$ \;
Compute $\U_{t} \in \arg\max_{\U\in \setU}R^{\hat\Mcal_t}(\U; \bvUCB{t}, \hat\bre_{t})$
\label{alg-calculate-u-Mt}
}		
\caption{Exploration-Exploitation Algorithm with Unknown  Arrival Distribution}
\end{algorithm}

\subsection{Poisson Customer Arrival Distribution}\label{sec:poisson}
Suppose that the true customer arrival distribution $\Mcal$ is a Poisson distribution with an unknown parameter $\lambda > 0$, denoted as $\Mcal \sim \text{Poisson}(\lambda)$. This means that for any $t \in [T]$, the probability mass function of $M_t$ is given by  
\begin{align*}
\mathbb{P}_{\Mcal}\left({M_t = m}\right) = \frac{e^{-\lm}\lm^{m}}{m\, !}, ~ m = 0, 1, 2, \dots
\end{align*}
At the start of each each cycle $t \in [T]$, we can  estimate $\lambda$ by its maximum likelihood estimator $\hat\lambda_t = \sum_{s\in[t-1]}M_s/(t-1)$, where $M_1, \dots, M_{t-1}$ are the observed numbers of customer arrivals in the first $t-1$ cycles. However, this estimated parameter may not result in a valid distribution estimator, because the corresponding Poisson distribution may not satisfy the stochastic dominance condition in \Cref{assump:Mt-error}. Instead, we consider as the distribution estimator a Poisson distribution with its parameter being an upper confidence bound on $\lambda$. That is, we set $\hat \Mcal_t$ as follows:
\begin{align}\label{eq:poisson-Mt}
\hat \Mcal_t \sim 
\text{Poisson}(\hat\lambda_t^{\text{UCB}}), \text{ where } \hat\lambda_t^{\text{UCB}} = \sum_{s\in[t-1]}M_s/(t-1) + a_M(e-1)\sqrt{\log(t-1)/(t-1)}. 
\end{align}

In the following lemma, we verify that the distribution estimator $\hat \Mcal_t$ defined above satisfies the requirement in \Cref{assump:Mt-error}. Consequently, the regret upper bound in \Cref{thm:upperbound mt} holds valid for Algorithm~\ref{alg:main mt} when the true customer arrival distribution is a Poisson distribution and
when the distribution estimator prescribed in \Cref{eq:poisson-Mt} is utilized. 

\begin{restatable}{lemma}{mtpoisson}
\label{prop:poisson}
The distribution estimator $\hat\Mcal_t$ in \Cref{eq:poisson-Mt} satisfies \Cref{assump:Mt-error} with the constants set as $a_1 = (e-1)/\lm$, $a_2 = 2a_M(e-1)$ and $a_3 = 2$, respectively. 
\end{restatable}

\subsection{Negative Binomial  Customer Arrival Distribution}\label{sec:neg-binomial}
Now, suppose that the true customer arrival distribution $\Mcal$ is a Negative Binomial distribution with a known parameter $r$ and an unknown parameter $p$, denoted as $\Mcal \sim \text{NB}(r, p)$. 
This means that for any $t \in [T]$, the probability mass function of $M_t$ is given by
\begin{align*}
\mathbb{P}_{\Mcal}\left({M_t=m}\right) = \binom{m+r-1}{m}(1-p)^m p^r, ~ m = 0, 1, \dots 
\end{align*}

The maximum likelihood estimator for $p$ at the start of cycle $t$ is $\hat p_t = \frac{r}{r + \sum_{s\in[t-1]}M_s/(t-1)}$, where $M_1, \dots, M_{t-1}$ are the observed numbers of customer arrivals in the first $t-1$ cycles. Accordingly, we construct a stochastically dominant distribution estimator as follows:
\begin{align}\label{eq:Mt-nb}
& \hat\Mcal_t \sim \text{NB}(r, \hat p_t^{\text{LCB}}),  \text{ where } \hat p_t^{\text{LCB}} \coloneqq \frac{r}{r + \hat M_t^{\UCB}}, \\
\text{ where } & \hat M_t^{\UCB} \coloneqq \sum_{s\in[t-1]}M_s/(t-1) + \max\{a_M,a_M^2/\sqrt{r}\} \sqrt{\frac{24\log (t + 1)}{t}} + \frac{48\log (t + 1)}{t} \ . \notag
\end{align}

Below we once again verify that this distribution estimator satisfies \Cref{assump:Mt-error}. Hence, the regret upper bound in \Cref{thm:upperbound mt} applies to Algorithm~\ref{alg:main mt} when the true customer arrival distribution $\Mcal$ is a Negative Binomial distribution and when the distribution estimator in \Cref{eq:Mt-nb} is utilized. 

\begin{restatable}{lemma}{mtnegabino} \label{prop:negabino}
The distribution estimator $\hat\Mcal_t$ in \Cref{eq:Mt-nb} satisfies \Cref{assump:Mt-error} with the constants set as $a_1 = 48$, $a_2 = 56\sqrt{3}\max\{a_M, a_M^2/\sqrt{r},1\}$, and $a_3 = 4$, respectively. 
\end{restatable}

}

\end{document}